%% file: arxiv.tex
\pdfoutput=1

\documentclass[acmtog]{acmart}

\renewcommand\footnotetextcopyrightpermission[1]{}

\usepackage{booktabs} %

\usepackage{pifont}
\newcommand{\ie}{\textit{i.e.}}
\newcommand{\eg}{\textit{e.g.}}

\usepackage{tikz}
\usetikzlibrary{spy}
\usepackage{multirow}
\usepackage{subcaption}
\usepackage{overpic}
\usepackage{marvosym}

\usepackage{booktabs}
\usepackage{adjustbox}
\usepackage{xcolor}    
\usepackage{tabulary}
\usepackage{colortbl}
\usepackage{makecell}
\usepackage{float}
\usepackage{url}
\usepackage{balance}
\usepackage[utf8]{inputenc} %
\usepackage[T1]{fontenc}    %
\usepackage{wrapfig}
\definecolor{color3}{rgb}{0.95,0.95,0.95}
\definecolor{myred}{rgb}{0.95,0, 0}
\definecolor{myblue}{rgb}{0, 0, 0.95}
\usepackage{pgfplots}
\usepackage{amsmath}
\usepackage{bm}

\newcommand{\method}{SplatDiff}

\usepackage[ruled]{algorithm2e} %

\SetAlFnt{\small}
\SetAlCapFnt{\small}
\SetAlCapNameFnt{\small}
\SetAlCapHSkip{0pt}

\begin{document}
\title{High-Fidelity Novel View Synthesis via Splatting-Guided Diffusion}

\author{Xiang Zhang}
\affiliation{%
  \institution{ETH Zürich}
  \city{Zürich}
  \country{Switzerland}}
\affiliation{%
  \institution{DisneyResearch|Studios}
  \city{Zürich}
  \country{Switzerland}}
\email{xiangz.ethz@gmail.com}

\author{Yang Zhang}
\affiliation{%
  \institution{DisneyResearch|Studios}
  \city{Zürich}
  \country{Switzerland}}
\email{yang.zhang@disneyresearch.com}

\author{Lukas Mehl}
\affiliation{%
  \institution{DisneyResearch|Studios}
  \city{Zürich}
  \country{Switzerland}}
\email{lukas.mehl.-nd@disneyresearch.com}

\author{Markus Gross}
\affiliation{%
  \institution{ETH Zürich}
  \city{Zürich}
  \country{Switzerland}}
\affiliation{%
  \institution{DisneyResearch|Studios}
  \city{Zürich}
  \country{Switzerland}}
\email{grossm@inf.ethz.ch}

\author{Christopher Schroers}
\affiliation{%
  \institution{DisneyResearch|Studios}
  \city{Zürich}
  \country{Switzerland}}
\email{christopher.schroers@disneyresearch.com}

\input{figs/main/fig-teaser}

\begin{abstract}
Despite recent advances in Novel View Synthesis (NVS), generating high-fidelity views from single or sparse observations remains a significant challenge. Existing splatting-based approaches often produce distorted geometry due to splatting errors. While diffusion-based methods leverage rich 3D priors to achieve improved geometry, they often suffer from texture hallucination. In this paper, we introduce \emph{\method}, a pixel-splatting-guided video diffusion model designed to synthesize high-fidelity novel views from a single image. Specifically, we propose an aligned synthesis strategy for precise control of target viewpoints and geometry-consistent view synthesis. To mitigate texture hallucination, we design a texture bridge module that enables high-fidelity texture generation through adaptive feature fusion. In this manner, \method\ leverages the strengths of splatting and diffusion to generate novel views with consistent geometry and high-fidelity details. Extensive experiments verify the state-of-the-art performance of \method\ in single-view NVS. Additionally, without extra training, \method\ shows remarkable zero-shot performance across diverse tasks, including sparse-view NVS and stereo video conversion.
\end{abstract}



\maketitle

\section{Introduction}
Novel view synthesis (NVS) has attracted considerable interest in the fields of computer vision and computer graphics, showing various applications like augmented/virtual reality, 3D generation, and stereo video conversion~\cite{mehl2024stereoconv,yu2024viewcrafter,gao2024cat3d}. Compared with previous optimization-based NVS approaches, \eg, neural radiance field (NeRF)~\cite{mildenhall2020nerf} and 3D Gaussian splatting (3DGS)~\cite{kerbl3Dgs}, that usually require dense input views and per-scene optimization, an emerging trend is to generate novel views from sparse views or even a single image in a feed-forward manner~\cite{szymanowicz2024flash3d,adampi,chen2025mvsplat,yu2024viewcrafter}. 
Due to the limited information from single/sparse views, such an NVS task is highly ill-posed and requires comprehensive scene understanding, including geometry, texture, and occlusion. Early works have proposed several techniques to tackle this challenging task, \eg, GAN-based inpainting for disocclusion regions~\cite{wiles2020synsin}, feed-forward neural scene prediction~\cite{yu2021pixelnerf}, and implicit 3D transformers~\cite{rombach2021geometry}.
In addition, various 3D representations are designed for efficient view synthesis, such as density fields~\cite{wimbauer2023bts}, multi-plane images~\cite{adampi}, and layered depth images~\cite{shih20203dphoto}. Despite significant progress being achieved, previous methods are often confined to specific domains and struggle to generalize to complex in-the-wild scenes. This usually arises from the limited prior knowledge of the 3D world. 

\par
Recent advancements in generative diffusion models have shown promising performance in a variety of 3D vision tasks~\cite{ke2024marigold,fu2025geowizard,zhang2024betterdepth}, including novel view synthesis~\cite{yu2024viewcrafter,sargent2024zeronvs}.
In order to generate high-quality images/videos across a wide array of domains, diffusion models are generally trained over internet-scale datasets, gaining rich prior knowledge of the visual world~\cite{rombach2022stablediffusion,xing2025dynamicrafter}. Benefiting from this, generative diffusion models exhibit outstanding performance in generating geometry-consistent content, naturally fitting the requirements of novel view synthesis. Many works are devoted to repurposing diffusion models for high-quality NVS, like semantic-preserving generative warping~\cite{seo2024genwarp} and point-conditioned video diffusion~\cite{yu2024viewcrafter}. 
However, because of the generative nature, diffusion models often introduce hallucinated contents, such as different textures, when generating novel views. Consequently, existing diffusion-based NVS methods usually struggle with texture hallucination, failing to preserve the original appearance present in the input view (\eg, Fig.~\ref{fig:main-teaser}).

\par 
Another popular trend is to render novel views with splatting-based approaches, \eg, Gaussian splatting~\cite{chen2025mvsplat,xu2024depthsplat}. For example, Flash3D employs a zero-shot depth estimator to predict the 3D Gaussian position and directly estimates the parameters of the 3D Gaussian for novel view rendering~\cite{szymanowicz2024flash3d}.
By enforcing the appearance consistency with pixel-/feature-level constraints, splatting-based NVS methods generally preserve better textures than diffusion-based approaches. However, since single or sparse observations provide only limited cues for the scene geometry, splatting-based methods often suffer from splatting errors, \eg, misalignment due to inaccurate depth, resulting in novel views with distorted geometry (\eg, Fig.~\ref{fig:main-teaser}).

\par
In this paper, we present \textbf{\method}, a video diffusion model guided by pixel splatting, designed to leverage the strengths of splatting and diffusion for high-fidelity novel view synthesis. The motivations behind our designs are as follows: 
(i) \emph{Pixel Splatting:}
Compared with the popular 3D Gaussian splatting techniques, we found that the simple pixel splatting, such as forward warping~\cite{niklaus2020softmaxsplat}, better preserves appearance under single or sparse input views. While 3D Gaussians can theoretically model complex visual effects, \eg, view-dependent appearance, estimating accurate Gaussian parameters (such as opacity) from limited observations remains a significant challenge. In addition, the estimation errors of Gaussian parameters often result in artifacts and cloudy effects (floaters) in novel views, yielding worse visual results than pixel splatting (\eg, see Fig.~\ref{fig:main-sparseview}).
(ii) \emph{Video Diffusion:} 
Our video diffusion model is designed to synthesize consistent and high-fidelity novel views with the guidance of splatted results. When input observations are limited, the splatted views often exhibit disocclusion regions that vary across different viewpoints. By training on large-scale video datasets, video generative diffusion models gain a deep understanding of visual elements such as geometry and texture. Leveraging this video diffusion prior, we synthesize realistic and consistent contents across varying viewpoints.

\par 
To achieve consistent geometry and high-fidelity texture, we incorporate two key components in \method: a novel strategy to create training pairs for \emph{aligned synthesis} and a \emph{texture bridge} to inject texture details into the diffusion decoder.
Specifically, we first fine-tune the pre-trained video diffusion model with Training Pair Alignment (TPA) and Splatting Error Simulation (SES) for aligned synthesis. TPA enforces the geometry and brightness consistency between the splatted and generated views, enabling precise control of target viewpoints. Meanwhile, our video diffusion model learns to eliminate splatting errors (\eg, flying pixels in Fig.~\ref{fig:main-alignsyn-result}) with SES, generating aligned novel views with consistent geometry. To tackle texture hallucination, we propose a texture bridge that aggregates the features from the splatted views for texture preservation. Additionally, a texture degradation strategy is introduced to facilitate the adaptive fusion of splatted views and diffusion outputs for high-quality synthesis. In summary, our main contributions are:
\begin{itemize}
    \item We introduce \method, a pixel-splatting-guided video diffusion model for synthesizing novel views with consistent geometry and high-fidelity texture from a single image.
    \item An aligned synthesis method to enable precise control of novel views while maintaining consistent geometry. In addition, we design a texture bridge module to achieve high-fidelity synthesis through adaptive feature fusion.
    \item \method\ excels in single-view NVS, sparse-view NVS, and stereo video conversion, demonstrating remarkable cross-domain and cross-task performance with training only on the single-view NVS task, as illustrated in Fig.~\ref{fig:main-teaser}. 
\end{itemize}

\input{figs/main/fig-overview}
\section{Related Work}
\subsection{Feed-Forward Novel View Synthesis}
A significant number of attempts are devoted to synthesizing novel views from single/sparse observations in a feed-forward manner. Due to the limited input information, early works usually adopt depth estimation methods to model scene geometry and then utilize inpainting approaches for content synthesis~\cite{wiles2020synsin,rombach2021geometry,rockwell2021pixelsynth}. To generate realistic novel views, several techniques are developed, including GAN-based inpainting~\cite{wiles2020synsin}, VQ-VAE outpainter~\cite{rockwell2021pixelsynth}, and implicit 3D transformer~\cite{rombach2021geometry}. Recently, novel scene representations are proposed to achieve high-quality view synthesis. For instance, pixelNeRF combines convolutional networks with NeRF representation to render novel views from two images~\cite{yu2021pixelnerf}. Meanwhile, layer-based representations, \eg, multi-plane images (MPI)~\cite{li2021mine,tucker2020svmpi,adampi,khan2023tiled} and layered depth images (LDI)~\cite{shih20203dphoto,jiang2023diffuse3d}, are exploited for efficient rendering.
However, since previous feed-forward NVS methods are mainly designed for specific domains, they often suffer from performance drops in complex scenes due to the limited model capability.

\subsection{Diffusion-Based Novel View Synthesis} 
Diffusion models have demonstrated exceptional performance in generating realistic images and videos~\cite{rombach2022stablediffusion,xing2025dynamicrafter}, reflecting a profound understanding of the 3D world.
To utilize the diffusion prior for novel view synthesis, previous attempts develop conditional diffusion frameworks, \eg, 3D feature-conditioned diffusion~\cite{chan2023gennvs} and viewpoint-conditioned diffusion~\cite{liu2023zero123}, to generate novel views for simple inputs like 3D objects~\cite{zheng2024free3d}. Considering complex real-world scenes, multi-view diffusion models are often employed to synthesize high-quality novel views, which are then used to generate 3D scenes (\eg, 3D Gaussians) for novel view rendering~\cite{liu2024reconx,wu2024reconfusion}. Based on this, ZeroNVS combines diverse training datasets to acquire zero-shot NVS performance~\cite{sargent2024zeronvs}, and Cat3D designs an efficient parallel sampling strategy for fast generation of 3D-consistent images~\cite{gao2024cat3d}. In addition, GenWarp exploits the diffusion prior to achieve semantic-preserving warping~\cite{seo2024genwarp}. Recent works also explore the potential of video diffusion models for novel view synthesis. For instance, ViewCrafter constructs a point-conditioned video diffusion model to iteratively complete the point cloud for consistent view rendering~\cite{yu2024viewcrafter}, and StereoCrafter proposes a tiled processing strategy to generate stereoscopic videos with video diffusion models~\cite{zhao2024stereocrafter}. While diffusion-based NVS approaches excel at synthesizing realistic novel views, the generative nature of diffusion models often introduces hallucinated content (\eg, Fig.~\ref{fig:main-teaser}), leading to inconsistent texture across different viewpoints.

\subsection{Splatting-Based Novel View Synthesis}
Splatting-based NVS approaches are typically trained in a regression manner with pixel-level or feature-level constraints~\cite{zhang2018lpips}. As a result, they often preserve better textures compared to diffusion-based methods. Previous study employs depth-based warping to achieve real-time novel view rendering~\cite{cao2022fwd}. With the rapid advancement of 3DGS techniques~\cite{kerbl3Dgs}, a considerable amount of attention has been drawn to feed-forward Gaussian splatting methods. The pioneer work pixelSplat estimates Gaussian parameters from neural networks and dense probability distributions, achieving efficient novel view synthesis with a pair of images~\cite{charatan2024pixelsplat}. Following this, several techniques are developed for improved performance and efficiency, including cost volume encoding~\cite{chen2025mvsplat} and depth-aware transformer~\cite{zhang2024transplat}. Recent method DepthSplat integrates monocular features from depth models and achieves better geometry in the estimated 3D Gaussians~\cite{xu2024depthsplat}. Instead of utilizing multi-view cues, another line of work focuses on predicting Gaussian parameters from a single image. Splatter Image obtains 3D Gaussian parameters from pure image features~\cite{szymanowicz2024splatterimage}, and Flash3D employs zero-shot depth models for generalizable single-view NVS~\cite{szymanowicz2024flash3d}. 
However, due to the challenges of estimating accurate geometry from limited observations, existing splatting-based methods often suffer from splatting errors, resulting in novel views with distorted geometry (\eg, Fig.~\ref{fig:main-teaser}). By contrast, our \method\ leverages the geometric priors of diffusion models to correct splatting errors, achieving geometry-consistent and high-fidelity novel view synthesis.

\input{figs/main/fig-diff}

\input{figs/main/fig-alignsyn}

\section{Splatting-Guided Diffusion}
As depicted in Fig.~\ref{fig:main-overview}, \method\ synthesizes novel views from a single image in a feed-forward manner. Given the input image, we predict the depth using an off-the-shelf depth estimator and then perform pixel splatting according to the camera poses. Since the splatted views often contain splatting errors and unknown regions, \eg, disocclusion, we leverage the video diffusion prior and propose splatting-guided diffusion to refine the splatted results.
Specifically, we propose a fine-tuning strategy to synthesize geometry-aligned contents while correcting the distortions caused by splatting errors. To tackle texture hallucination, we design a texture bridge to aggregate encoder features for high-fidelity novel view synthesis. In the following sections, we first provide a brief introduction to pixel splatting and video diffusion models in Sec.~\ref{sec:preliminary}. The designs of the aligned synthesis strategy and the texture bridge module are then presented in Secs.~\ref{sec:alignsyn} and \ref{sec:texturebridge}, respectively.

\subsection{Preliminaries}\label{sec:preliminary}
\noindent \textbf{Pixel Splatting.} Given a transformation map, \eg, optical flow, pixel splatting projects pixels from the input image to the target image, which has been widely used in computer vision tasks like frame interpolation~\cite{niklaus2020softmaxsplat}. To generate novel views, we compute the view transformation map with camera poses and the estimated depth. Similar to prior works~\cite{niklaus2020softmaxsplat,mehl2024stereoconv}, we employ softmax splatting to resolve splatting collisions and assign pixel importance based on the depth information. This approach gives higher blending weights to foreground objects, preserving the geometric layout of the input view.

\par 
\noindent \textbf{Video Diffusion.} Diffusion models typically consist of a forward process and a reverse process~\cite{ho2020DDPM,song2020DDIM}. The forward process $q(\mathbf{x}_t|\mathbf{x}_0,t)$ converts data $\mathbf{x}_0 \sim p_{\text{data}}(\mathbf{x})$ into Gaussian noise $\mathbf{x}_T \sim \mathcal{N}(\mathbf{0},\mathbf{I})$ by gradually adding noise at each step $t\in\{1,\dots,T\}$, and the learned reverse process $p_\theta(\mathbf{x}_{t-1}|\mathbf{x}_t,t)$ transforms random Gaussian noise to a new sample with a denoising network $\epsilon_\theta$. At each denoising step, the network $\epsilon_\theta$ is supervised by
\begin{equation}
    \min_\theta \mathbb{E}_{\mathbf{x}\sim p_{\text{data}}, {\epsilon} \sim \mathcal{N}(\bm{0}, \mathbf{I}),t \sim \mathcal{U}(T)} \left\| \epsilon - \epsilon_\theta(\mathbf{x}_t, t) \right\|^2_2,
\end{equation}
where $\epsilon$ is the Gaussian noise, and $\mathbf{x}_t$ denotes the noisy sample at step $t$. By learning to estimate the added noise with $\epsilon_\theta$, diffusion models are able to generate new data samples via iterative denoising. 
\par 
We build \method\ upon latent video diffusion models to balance performance and computational complexity. As shown in Fig.~\ref{fig:main-overview}, we first generate the latent representations $\mathbf{z}\in \mathbb{R}^{C\times h \times w}$ for each image $\mathbf{x}\in \mathbb{R}^{3\times H\times W}$ with a latent encoder. Following that, we concatenate the latent codes $\mathbf{Z}=\operatorname{Cat}(\{\mathbf{z}\})$ as conditioning and perform the forward and reverse processes in the latent space, \ie, $q(\mathbf{Z}_t|\mathbf{Z}_0,t)$ and $p_\theta(\mathbf{Z}_{t-1}|\mathbf{Z}_t,t)$. Finally, the novel views are generated by decoding the denoised latent codes $\mathbf{Z}_{0}$ to the image space. Based on this pipeline, we propose the aligned synthesis strategy and the texture bridge module to achieve high-fidelity novel view synthesis.

\subsection{Aligned Synthesis}\label{sec:alignsyn}
Previous diffusion-based NVS approaches often generate textures and geometry that are misaligned with the conditioning (Fig.~\ref{fig:main-diff}), making precise control of target viewpoints challenging. This misalignment typically stems from the use of unaligned pairs during diffusion training. Given the depth $\mathbf{d}$ and the camera poses $\mathbf{p}$, naive training methods often construct training pairs $\{\mathbf{v}_{\text{tgt}}, \mathbf{x}_{\text{tgt}}\}$ with
\begin{equation}
    \mathbf{v}_{\text{tgt}} = \operatorname{Render}(\mathbf{x}_{\text{src}},\mathbf{d},\mathbf{p}),
\end{equation}
where $\operatorname{Render}(\cdot)$ denotes a novel view renderer, \eg, point cloud renderer~\cite{yu2024viewcrafter}. $\mathbf{x}_{\text{src}},\ \mathbf{x}_{\text{tgt}},\ \text{and}\ \mathbf{v}_{\text{tgt}}$ correspond to the source input view, the target view, and the rendered view, respectively. Then, the diffusion model is trained to predict $\mathbf{x}_{\text{tgt}}$ conditioned on $\mathbf{v}_{\text{tgt}}$. However, the conditioning $\mathbf{v}_{\text{tgt}}$ often shows different texture and geometry with the target view $\mathbf{x}_{\text{tgt}}$ (\eg, Fig.~\ref{fig:main-alignsyn-pipeline}) due to several factors, \eg, different lighting and depth estimation error. This renders the diffusion model to produce misaligned novel views with the conditioned view, as shown in Fig.~\ref{fig:main-alignsyn-result}. To this end, we propose the Training Pair Alignment (TPA) strategy for aligned synthesis.

\par 
\noindent \textbf{Training Pair Alignment.} 
Instead of generating the diffusion conditioning from the input view $\mathbf{x}_{\text{src}}$, TPA utilizes the target view $\mathbf{x}_{\text{tgt}}$ to construct aligned training pairs. As illustrated in Fig.~\ref{fig:main-alignsyn-pipeline}, we first estimate the view transformation map from $\mathbf{x}_{\text{src}}$ and $\mathbf{x}_{\text{tgt}}$ with an optical flow estimator~\cite{xu2023unimatch}. With the estimated flow, we then generate a splatting mask $\textbf{m}_{\text{splat}}$ via pixel splatting, where $\textbf{m}_{\text{splat}}$ indicates the valid splatting regions. Finally, we construct the aligned training pairs $\{\tilde{\mathbf{v}}_{\text{tgt}}, \mathbf{x}_{\text{tgt}}\}$ by masking the target view, \ie,
\begin{equation}
    \tilde{\mathbf{v}}_{\text{tgt}} = \mathbf{x}_{\text{tgt}} \odot \textbf{m}_{\text{splat}}.
\end{equation}
Since $\tilde{\mathbf{v}}_{\text{tgt}}$ aligns with $ \mathbf{x}_{\text{tgt}}$ in both geometry and texture, the diffusion model trained with TPA learns to adhere to the conditioned view, producing consistent novel views (\eg, Fig.~\ref{fig:main-alignsyn-result}). However, due to the existence of splatting errors in real splatted views, the model trained solely with TPA tends to be misled by the splatting errors, resulting in artifacts as shown in the green box in Fig.~\ref{fig:main-alignsyn-result}. To address this, we propose the Splatting Error Simulation (SES) method to enhance the model's robustness against splatting errors.

\par 
\noindent \textbf{Splatting Error Simulation.}
Splatting errors in pixel splatting often appear as flying pixels, \eg, green box in Fig.~\ref{fig:main-alignsyn-result}, which typically arise from inaccuracies in the transformation map, \eg, blurred depth discontinuities~\cite{shih20203dphoto}. As a result, pixels around object boundaries might be projected to incorrect positions in the target view, leading to distorted geometry with flying pixels. To resolve this, we propose to simulate splatting errors in the training pairs (Fig.~\ref{fig:main-alignsyn-pipeline}). Since the flying pixels often stem from the object boundaries, we first generate an edge mask $\mathbf{m}_{\text{edge}}$ from the transformation map using a Sobel operator. Next, we extract the edge regions $\mathbf{e}_{\text{src}}$ from the input view $\mathbf{x}_{\text{src}}$ by $\mathbf{e}_{\text{src}} = \mathbf{x}_{\text{src}} \odot \textbf{m}_{\text{edge}}.$ The extracted edge regions $\mathbf{e}_{\text{src}}$ are then splatted to the target view using the optical flow to generate the splatted edge $\mathbf{e}_{\text{tgt}}$. Finally, we simulate the splatting errors in $\tilde{\mathbf{v}}_{\text{tgt}}$ by random perturbation:   
\begin{equation}
    \hat{\mathbf{v}}_{\text{tgt}} = \mathbf{e}_{\text{tgt}} \odot \mathbf{m}_{\text{error}} + \tilde{\mathbf{v}}_{\text{tgt}}\odot (1-\mathbf{m}_{\text{error}}),
\end{equation}
where $\hat{\mathbf{v}}_{\text{tgt}}$ denotes the rendered view with simulated splatting errors, and $\mathbf{m}_{\text{error}}$ is a randomly generated mask indicating where to introduce splatting errors. By using $\{\hat{\mathbf{v}}_{\text{tgt}}, \mathbf{x}_{\text{tgt}}\}$ as training pairs, our diffusion model learns to correct splatting errors by utilizing its rich geometric prior, while maintaining aligned synthesis (Fig.~\ref{fig:main-alignsyn-result}).

\subsection{Texture Bridge}\label{sec:texturebridge}
\input{figs/main/fig-texturebridge}

Although the diffusion model can generate geometry-consistent novel views with our aligned synthesis strategy, its outputs often contain hallucinated textures that are inconsistent with the input view (Fig.~\ref{fig:main-teaser}). One potential solution is to preserve the texture of the input image by utilizing the splatted view. However, directly blending the splatted view and the diffusion output faces several challenges: (i) the splatted view usually contains aliasing artifacts, such as jagged edges, and may exhibit blurred details due to the resampling process, which can degrade the quality of the novel view, and (ii) detecting and removing splatting errors in the splatted view is crucial for achieving high-quality NVS. Thus, we propose the texture bridge module to adaptively fuse the splatted view with the diffusion output for high-fidelity synthesis. 
As depicted in Fig.~\ref{fig:main-texturebridge}, the texture bridge consists of a fusion block at each feature scale. 
Let $\mathbf{f}^i_{\text{enc}}$ denote the $i$-th scale encoder feature extracted from the splatted view, and let $\mathbf{f}^i_{\text{dec}}$ represent the $i$-th scale decoder feature. We first fuse the features at each scale with the texture bridge, \ie,
\begin{equation}
    \mathbf{f}^i_{\text{fuse}} = \operatorname{Fuse}_i(\mathbf{f}^i_{\text{enc}}\
    ,\ \mathbf{f}^i_{\text{dec}}),
\end{equation}
and then restore the novel view by passing the fused features $\{\mathbf{f}^i_{\text{fuse}}\}$ through the latent decoder.
Each fusion block $\operatorname{Fuse}_i(\cdot)$ is implemented using two residual blocks~\cite{he2016resnet}, though more advanced architecture, \eg, self-attention, could be employed for better performance. With multi-scale fusion, the texture bridge effectively aggregates fine-grained features for high-quality synthesis.

\input{tabs/tab-main-singleview}
\par 
\noindent \textbf{Training with Texture Degradation.}
Our texture bridge is designed to adaptively select optimal features from the splatted view and the diffusion output for view synthesis. For example, the texture bridge should mainly utilize the diffusion output when encountering splatting errors, while relying more on the splatted view in the regions with texture hallucination.
To achieve this, one possible training approach is to feed the texture bridge with the splatted view and the corresponding diffusion output, then supervise the decoded results using the target view. However, the diffusion outputs often differ significantly from the target view in the unknown regions (\eg, disocclusion), resulting in sub-optimal training performance. To overcome this, we propose a texture degradation strategy to facilitate model training. Specifically, we employ the diffusion model to degrade the texture of the target view and train the texture bridge using the degraded view (Fig.~\ref{fig:main-texturebridge}). As a result, the degraded view shares similar contents to the target view while containing hallucinated textures for training. For supervision, we employ the $\ell_1$ loss $\mathcal{L}_1$ and the perceptual loss $\mathcal{L}_{\text{LPIPS}}$~\cite{zhang2018lpips}, \ie,
\begin{equation}
    \mathcal{L}=\mathcal{L}_1 + \alpha \mathcal{L}_{\text{LPIPS}},
\end{equation}
where $\alpha=0.1$. With the texture bridge, our \method\ addresses texture hallucination through the adaptive fusion of the splatted view and the diffusion output. Meanwhile, the texture bridge also learns to refine the input features, \eg, aliasing artifacts and blurry details in the splatted view, for high-fidelity novel view synthesis.

\section{Experiments and Analysis}
\subsection{Experimental Settings}
\par 
\textbf{Implementation Details.}
We implement \method\ based on the open-source video diffusion model DynamiCrafter~\cite{xing2025dynamicrafter} with ViewCrafter weight initialization~\cite{yu2024viewcrafter}. 
We employ the AdamW optimizer~\cite{loshchilov2018adamw} to train \method\ under $320\times512$ patches and batch size 16. For aligned synthesis, we only fine-tune the latent U-Net for 1.5K steps with a learning rate 1$\times$10$^{-5}$. Afterward, we train the texture bridge for 10K steps with a learning rate 1$\times$10$^{-4}$. The total training takes around 2.5 days on a single NVIDIA RTX A6000 GPU. 
At inference, we apply the DDIM scheduler~\cite{song2020DDIM} with 50-step sampling. 

\par 
\noindent \textbf{Datasets and Evaluation.}
We use two datasets \textbf{RealEstate10K} \cite{re10k} and \textbf{DL3DV-10K}~\cite{ling2024dl3dv} for training and evaluation. For each dataset, two evaluation sets (easy and hard sets) are created with different baseline ranges. In RealEstate10K, we follow the setting in Flash3D~\cite{szymanowicz2024flash3d} to skip 5 and random $\pm$30 frames for the easy and hard sets. Since DL3DV-10K features faster camera motion and more complex scenes, we skip 3 and 6 frames for the easy and hard sets. For the methods requiring depth as inputs, we use the same depth models for fair comparisons (UniDepth~\cite{piccinelli2024unidepth} in RealEstate10K and DepthSplat~\cite{xu2024depthsplat} in DL3DV-10K). Quantitative evaluation is conducted with pixel-level metrics (PSNR and SSIM), feature-level metrics (LPIPS~\cite{zhang2018lpips} and DISTS~\cite{dists}), and distribution-level metric FID~\cite{heusel2017fid}. Since the generated novel views are often not perfectly aligned with the ground-truth due to depth estimation errors, the importance of these metrics follows the hierarchy: distribution-level > feature-level > pixel-level. In-the-wild samples are also collected for qualitative evaluation.

\input{figs/main/fig-singleview}

\input{figs/main/fig-singleview-wild}

\subsection{Benchmarking}
Tab.~\ref{tab:main-singleview} shows the single-view NVS results of \method\ compared with prior arts. Previous diffusion-based NVS approaches, \eg, ViewCrafter, struggle to achieve good metrics due to the hallucinated geometry and textures as depicted in Figs.~\ref{fig:main-singleview} and \ref{fig:main-singleview-wild}. Although splatting-based methods, \eg, Flash3D, better preserve texture from the input view, the generated novel views often suffer from geometry distortion due to splatting errors (Fig.~\ref{fig:main-singleview-re10k}). Compared with them, our \method\ produces the best novel views with consistent geometry and high-fidelity textures as shown in Fig.~\ref{fig:main-singleview}. In addition, 
when better depth maps are available (\eg, on DL3DV-10K, where the depth is estimated from two views), \method\ shows significantly better performance than both diffusion-based and splatting-based approaches (Tab.~\ref{tab:main-singleview}). With our aligned synthesis strategy and texture bridge module, \method\ even outperforms the two-view method DepthSplat and generates better visual results with fine-grained details (Fig.~\ref{fig:main-singleview-dl3dv}).

\input{tabs/tab-main-ablation}

\subsection{Ablation Study}
In Tab.~\ref{tab:main-ablation}, we study the effectiveness of each design: training pair alignment (TPA), splatting error simulation (SES), texture bridge (TB), and texture degradation (TD). (i) \textbf{Aligned Synthesis}: Compared with the baseline model (\#1), the model with TPA (\#2) enforces the consistency between the conditioned view and the generated view, boosting the novel view synthesis performance (\eg, 1.69 dB PSNR gain). 
With SES, the diffusion model further learns to correct splatting errors by leveraging its rich geometric prior. Consequently, the model with TPA and SES achieves aligned synthesis while preserving geometric consistency (Fig.~\ref{fig:main-alignsyn-result}). (ii) \textbf{Texture Bridge}: To handle texture hallucination, TB aggregates the multi-scale features from the splatted view and the diffusion output, leading to significant improvements across all metrics (\#4 in Tab.~\ref{tab:main-ablation}). Furthermore, training with TD improves feature fusion and refines the input features for better synthesis. As a result, the model incorporating all designs achieves the best NVS performance (\#5 in Tab.~\ref{tab:main-ablation}).

\input{tabs/tab-main-sparseview}

\input{tabs/tab-main-stereoconv}
\subsection{Applications}
We directly apply the \method\ pre-trained on single-view NVS to different tasks, \ie, sparse-view NVS and stereo video conversion. By simply modifying the model inputs, \method\ demonstrates remarkable cross-domain performance as shown in Tabs.~\ref{tab:main-sparseview} and \ref{tab:main-stereoconv}.

\input{figs/main/fig-sparseview}

\input{figs/main/fig-stereconv}

\par 
\noindent \textbf{Sparse-View Novel View Synthesis.}
We evaluate the sparse-view NVS performance with two input views. Given the two splatted views, we simply select the one with fewer unknown regions as the primary input and use the other one to fill the unknown regions with a blurred blending mask. We follow DepthSplat to conduct the in-domain evaluation on the DL3DV-10K dataset. Under limited observations, it is challenging to estimate accurate parameters for 3D Gaussians, which often leads to cloudy results in novel views (Fig.~\ref{fig:main-sparseview-dl3dv}). With pixel splatting, \method\ achieves state-of-the-art performance as shown in Tab.~\ref{tab:main-sparseview}. We also follow MVSplat to conduct the cross-domain evaluation on the DTU dataset~\cite{jensen2014dtu}. Due to the domain gap, the estimated depth is less accurate, leading to misalignment between the splatted view and the target view. In such cases, Gaussian splatting approaches, \eg, DepthSplat, often blur details to handle the misalignment, resulting in slightly higher pixel-level metrics in Tab.~\ref{tab:main-sparseview}. By contrast, our \method\ achieves significantly better visual quality with fine-grained details (Fig.~\ref{fig:main-sparseview-dtu}). To mitigate the misalignment in \method, we additionally average the novel views generated from the two splatted views, which outperform DepthSplat across all metrics (\method$^\dagger$ in Tab.~\ref{tab:main-sparseview}).

\par 
\noindent \textbf{Stereo Video Conversion.}
Stereo video conversion is widely used in movie production, and thus we employ the Spring dataset~\cite{mehl2023spring} for evaluation, which features high-resolution stereo videos from the Blender movie "Spring". We generate the splatted right-eye views for \method\ from the input left-eye videos, and the ground-truth disparity is used in all methods for fair comparisons. As shown in Fig.~\ref{fig:main-stereoconv}, it is challenging for ViewCrafter to handle dynamic scenes, resulting in significantly different contents in the novel views. Although StereoCrafter generates more consistent novel views with the inputs, the synthesized views often suffer from texture hallucination and blurry details. Benefiting from the aligned synthesis strategy, our \method\ can be directly applied to dynamic videos without additional training. Meanwhile, the proposed texture bridge preserves high-fidelity details from inputs, achieving the best stereo video conversion performance (Tab.~\ref{tab:main-stereoconv} and Fig.~\ref{fig:main-stereoconv}).

\section{Conclusion}
We propose \method, a pixel-splatting-guided video diffusion model designed for geometry-consistent and high-fidelity novel view synthesis. With the aligned synthesis strategy, \method\ achieves precise control of target viewpoints while effectively correcting geometry distortion caused by splatting errors. In addition, the proposed texture bridge recovers high-fidelity texture via adaptive feature fusion. Extensive experiments across single-view novel view synthesis, sparse-view novel view synthesis, and stereo video conversion verify the versatility and state-of-the-art performance of \method.

\bibliographystyle{ACM-Reference-Format}
\bibliography{sample-bibliography}

\appendix

\input{figs/supp/fig-texture_degradation}

\input{figs/supp/fig-ablation-vis}

\section{More Experiments}\label{sec:supp-exp}
\subsection{Texture Degradation}
The goal of our texture bridge is to adaptively fuse the splatted view and the diffusion output for high-quality view synthesis. As depicted in Fig.~\ref{fig:supp-texture-degradation}, the splatted view shows better textures (green box) but contains unknown regions and splatting errors. Although the diffusion output generates reasonable contents for the unknown regions (red box), they often suffer from texture hallucination (green box). Ideally, the texture bridge should learn to detect the problematic regions, \eg, splatting errors and hallucinated textures, and adaptively utilize better features for synthesis. However, directly using the diffusion output often leads to sub-optimal training performance mainly due to the inconsistent contents in the unknown regions (red box in Fig.~\ref{fig:supp-texture-degradation}). To address this, we propose the texture degradation strategy and generate the degraded views for training. As shown in Fig.~\ref{fig:supp-texture-degradation}, the degraded view shares similar contents with the target view while imitating the texture hallucination effects in real diffusion outputs. By substituting the diffusion output with the degraded view for training, our texture bridge better learns to rely more on the generated contents for the unknown regions while maintaining the ability to detect hallucinated textures.

\subsection{Visual Comparisons in Ablation Study}
Fig.~\ref{fig:supp-ablation-vis} provides visual results for the ablation study (Tab. 2 in the main paper). We also compute the absolute difference map between the splatted views and the corresponding novel views to visualize the difference regions. It is obvious that the baseline model synthesizes misaligned novel views and produces hallucinated textures, leading to significant difference regions across the entire image. With the aligned synthesis strategy, the model (\#3 in Tab. 2 of the main paper) generates geometry-consistent results but still suffers from texture hallucination as shown in the green box of Fig.~\ref{fig:supp-ablation-vis}. By contrast, our \method\ utilizes the information from the splatted view with the proposed texture bridge, achieving high-fidelity novel view synthesis.

\input{figs/supp/fig-singleview-wild}

\input{figs/supp/fig-stereoconv}

\section{More Visual Results}\label{sec:supp-vis}
\subsection{Zero-Shot Performance}
To verify the zero-shot performance of \method, we provide additional visual comparisons on diverse in-the-wild samples, including landscapes, buildings, animals, and paintings (Fig.~\ref{fig:supp-singleview-wild}). Previous approaches, \eg, ViewCrafter~\cite{yu2024viewcrafter}, often synthesize contents that differ from the inputs, like the hallucinated lighting and the rooftop texture in the first-row example of Fig.~\ref{fig:supp-singleview-wild}. In contrast, our method preserves the geometric layout while recovering fine-grained details as shown in the green box of Fig.~\ref{fig:supp-singleview-wild}. In addition, benefiting from the aligned synthesis strategy, our method corrects the splatting errors and fills reasonable contents for the unknown regions, \eg, the green boxes in the bird and the axe examples depicted in Fig.~\ref{fig:supp-singleview-wild}.

\subsection{Video Results in Stereo Conversion}
In Fig.~\ref{fig:supp-stereoconv}, we provide more stereo video conversion results on the Spring dataset~\cite{mehl2023spring}. Since ViewCrafter is trained only on static scenes, it is challenging for it to handle dynamic inputs, resulting in severe content hallucination as shown in Fig.~\ref{fig:supp-stereoconv}. Although our method is trained on the same static datasets as ViewCrafter, \method\ exhibits promising zero-shot performance in dynamic scenes. This is because our texture bridge module leverages the splatted views to maintain the same motion as the input video. Meanwhile, with our aligned synthesis strategy, the diffusion model performs similarly to inpainting models and fills reasonable contents for the unknown regions, \eg, disocclusions in the last two rows of Fig.~\ref{fig:supp-stereoconv-detail}, achieving high-quality synthesis of the stereo video.

\par
Compared with StereoCrafter~\cite{zhao2024stereocrafter} that is designed specifically for stereo video conversion, our method still shows superior performance in synthesis quality and consistency. As depicted in Fig.~\ref{fig:supp-stereoconv-detail}, StereoCrafter tends to produce novel views with smoothed details, \eg, the rocks on the road, and fills blurred contents for the disocclusion regions. For consistency, while StereoCrafter generates better novel views than ViewCrafter, inconsistent details are often observed in the outputs of StereoCrafter, such as the girl's eyes in the green box of Fig.~\ref{fig:supp-stereoconv-consistency}. Compared with previous approaches, our method produces the best stereo conversion results with consistent geometry and realistic details, verifying the effectiveness and versatility of \method.

\section{Limitations and Future Works}\label{sec:supp-limitation}
While our \method\ achieves remarkable performance on many novel view synthesis tasks, limitations still exist: 
(i) \textbf{View-Dependent Effects}: Compared with the popular Gaussian splatting techniques, \method\ has demonstrated the effectiveness of the pixel splatting method under limited input views. However, since current pixel splatting methods generally assume the brightness constancy across different viewpoints, how to handle the view-dependent effects, \eg, reflective surfaces, remains an open problem. Although Gaussians are capable of modeling view-dependent effects, estimating accurate Gaussian parameters from limited observations is challenging. One potential solution is to leverage the rich prior from foundation models to facilitate the modeling of view-dependent effects, and we leave it for future works.
(ii) \textbf{Long-Range Consistency}:
Most existing video diffusion models are designed to output a pre-defined number of images, and thus multiple inferences are required to handle long-range inputs, \eg, long videos in stereo conversion. However, due to the generative nature of diffusion models, the synthesized contents are usually different in multiple inferences. Although our approach utilizes the texture bridge to maintain the consistency on the splatted regions, the contents on the unknown regions, \eg, disocclusions, might differ. To achieve long-range consistency, one could draw inspiration from the recent video-based methods, \eg, rolling inference~\cite{ke2024rollingdepth}, for improved consistency of diffusion outputs, and combine the techniques in our \method\ for high-fidelity view synthesis.

\end{document}

%% file: figs/main/fig-teaser.tex
\def\imgWidth{0.238\textwidth} %
\def\scc{(-1.9,-1.4)}
\def\rebigone{(2.1, -0.8)} %
\def\reone{(-1.45,0.42)} %
\def\scc{(-2,-1.2)}

\def\ssmag{3}

\begin{teaserfigure}
    \centering
    \tikzstyle{img} = [rectangle, minimum width=\imgWidth, draw=black]
\centering
\begin{subfigure}{\textwidth}
\centering
    \includegraphics[width=\textwidth, trim=0 0 0 0, clip]{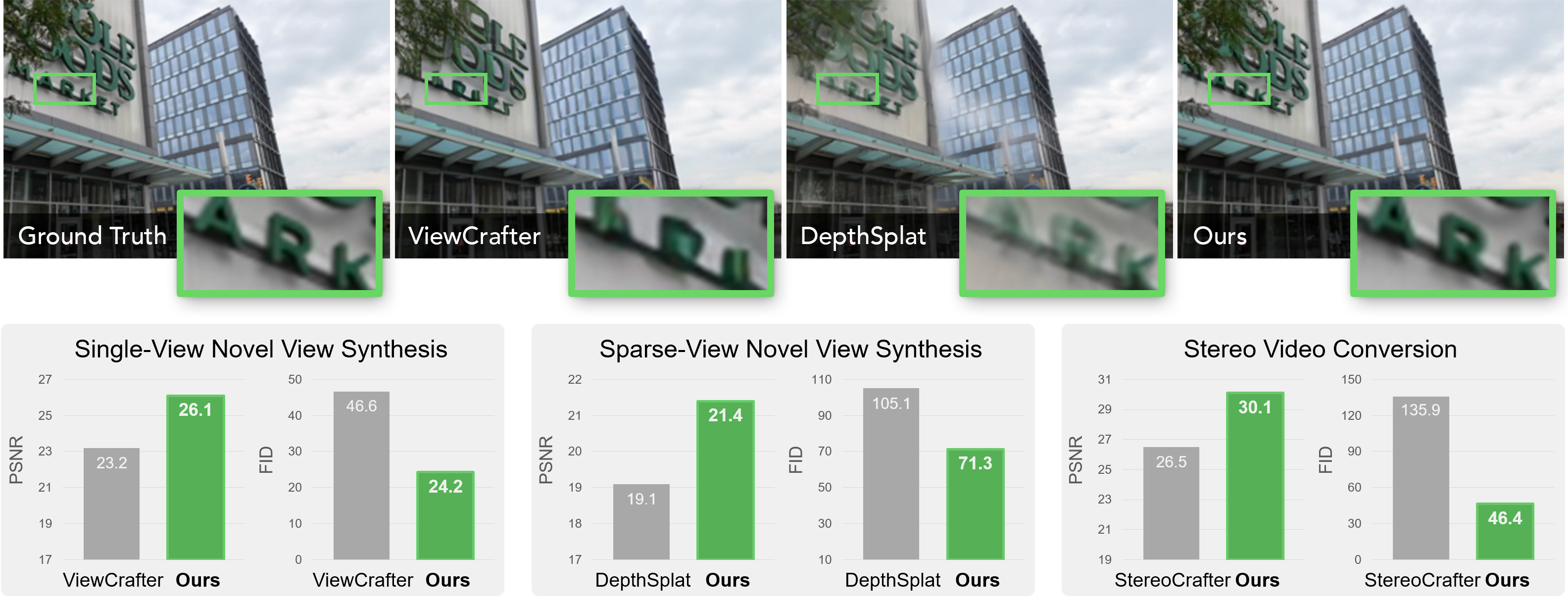}
    \end{subfigure}
    \caption{\textbf{Performance comparison.} Diffusion-based methods, \eg, ViewCrafter, usually hallucinate contents that are inconsistent with the input view. Splatting-based approaches, \eg, DepthSplat, often suffer from distorted geometry due to splatting errors. By contrast, our method produces novel views with consistent geometry and high-fidelity texture, achieving significantly better performance than previous arts on different tasks. Note that our model is trained only on the single-view novel view synthesis and is directly applied to the other tasks, showing promising cross-domain and cross-task performance.}
    \label{fig:main-teaser}
\end{teaserfigure}

%% file: figs/main/fig-overview.tex
\begin{figure*}[!t]
    \centering
    \includegraphics[width=0.8\linewidth]{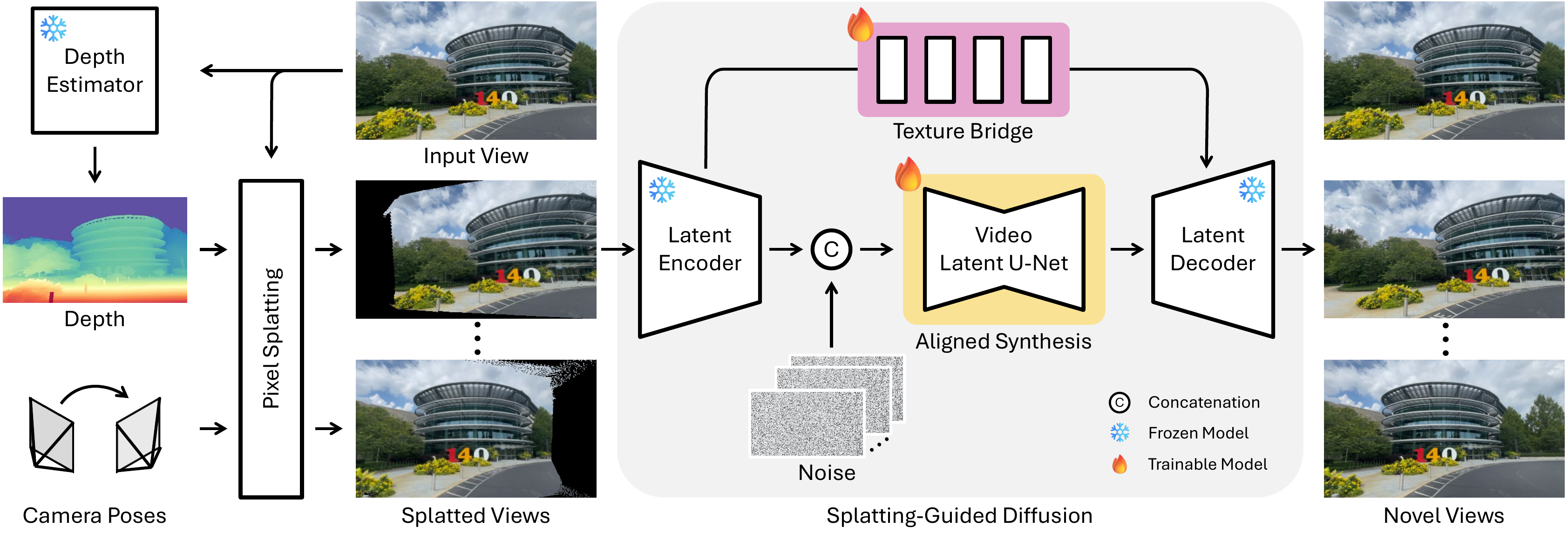}
    \caption{\textbf{Overview of \method.} Given the input view, we first estimate the depth information from a depth estimator and then perform pixel splatting to generate splatted views as diffusion conditioning. In our splatting-guided diffusion, we fine-tune the latent U-Net for aligned synthesis, producing consistent novel views while correcting splatting errors. Meanwhile, a texture bridge module is designed to aggregate the encoder features for high-fidelity synthesis.}
    \label{fig:main-overview}
\end{figure*}

%% file: figs/main/fig-diff.tex
\def\imgWidth{0.32\linewidth} %
\def\scc{(-1.9,-1.4)}
\def\rebigone{(-0.54, 0.5)} %
\def\refour{(0.08,0.23)} %

\def\ssizz{0.8cm} %
\def\ssmag{3}

\begin{figure}[t]
\centering
\tikzstyle{img} = [rectangle, minimum width=\imgWidth, draw=black]
    \centering
    \begin{subfigure}{\imgWidth}
        \begin{tikzpicture}[spy using outlines={green,magnification=\ssmag,size=\ssizz},inner sep=0]
            \node [align=center, img] {\includegraphics[width=\textwidth]{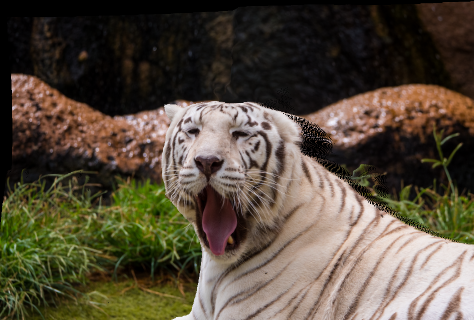}};
            \spy on \refour in node [left] at \rebigone;
    	\end{tikzpicture}
     \caption*{Splatted View}
    \end{subfigure}
    \begin{subfigure}{\imgWidth}
		\begin{tikzpicture}[spy using outlines={green,magnification=\ssmag,size=\ssizz},inner sep=0]
            \node [align=center, img] {\includegraphics[width=\textwidth]{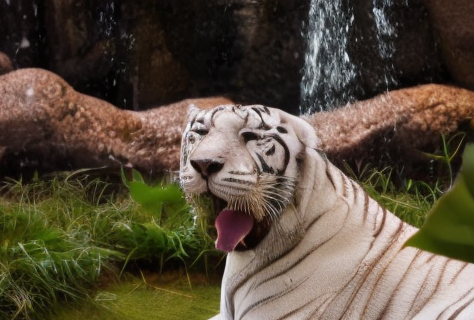}};
            \spy on \refour in node [left] at \rebigone;
    	\end{tikzpicture}
     \caption*{ViewCrafter}
    \end{subfigure}
    \begin{subfigure}{\imgWidth}
        \begin{tikzpicture}[spy using outlines={green,magnification=\ssmag,size=\ssizz},inner sep=0]
            \node [align=center, img] {\includegraphics[width=\textwidth]{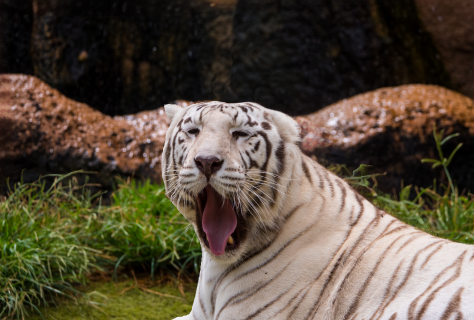}};
            \spy on \refour in node [left] at \rebigone;
    	\end{tikzpicture}
      \caption*{Ours}
      \end{subfigure}
      \\ %
      \begin{subfigure}{\imgWidth}
        \begin{tikzpicture}[spy using outlines={green,magnification=\ssmag,size=\ssizz},inner sep=0]
            \node [align=center, img] {\includegraphics[width=\textwidth]{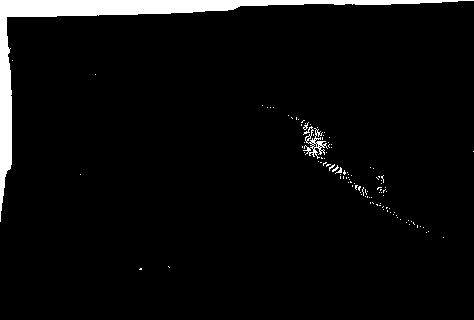}};
    	\end{tikzpicture}
     \caption*{Unknown Region}
    \end{subfigure}
    \begin{subfigure}{\imgWidth}
		\begin{tikzpicture}[spy using outlines={green,magnification=\ssmag,size=\ssizz},inner sep=0]
            \node [align=center, img] {\includegraphics[width=\textwidth]{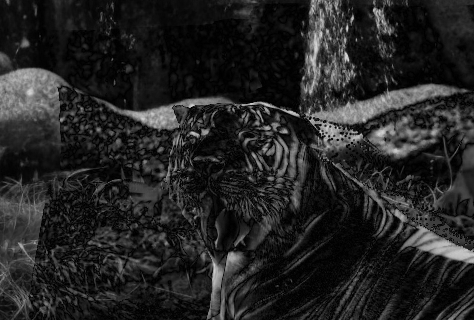}};
    	\end{tikzpicture}
     \caption*{Diff. Map (ViewCrafter)}
    \end{subfigure}
    \begin{subfigure}{\imgWidth}
        \begin{tikzpicture}[spy using outlines={green,magnification=\ssmag,size=\ssizz},inner sep=0]
            \node [align=center, img] {\includegraphics[width=\textwidth]{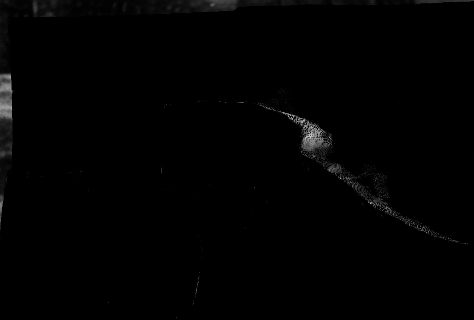}};
    	\end{tikzpicture}
      \caption*{Diff. Map (Ours)}
      \end{subfigure}
    \caption{\textbf{Misalignment.} The difference map shows the absolute difference between the splatted view and the generated view. Diffusion-based methods, \eg, ViewCrafter, often generate misaligned contents in novel views, resulting in significant differences across the image. In contrast, our \method\ is faithful to inputs and shows differences mainly around the unknown region.}
    \label{fig:main-diff}
\end{figure}

%% file: figs/main/fig-alignsyn.tex
\def\imgWidth{0.195\textwidth} %
\def\scc{(-1.9,-1.4)}
\def\rebigone{(-0.43, 0.5)} %
\def\refour{(1.45,-0.66)} %

\def\ssizz{1.3cm} %
\def\ssmag{3}

\begin{figure*}[t]
    \centering
    \tikzstyle{img} = [rectangle, minimum width=\imgWidth, draw=black]
\centering
    \begin{subfigure}[b]{\linewidth}
    \centering
    \includegraphics[width=\linewidth]{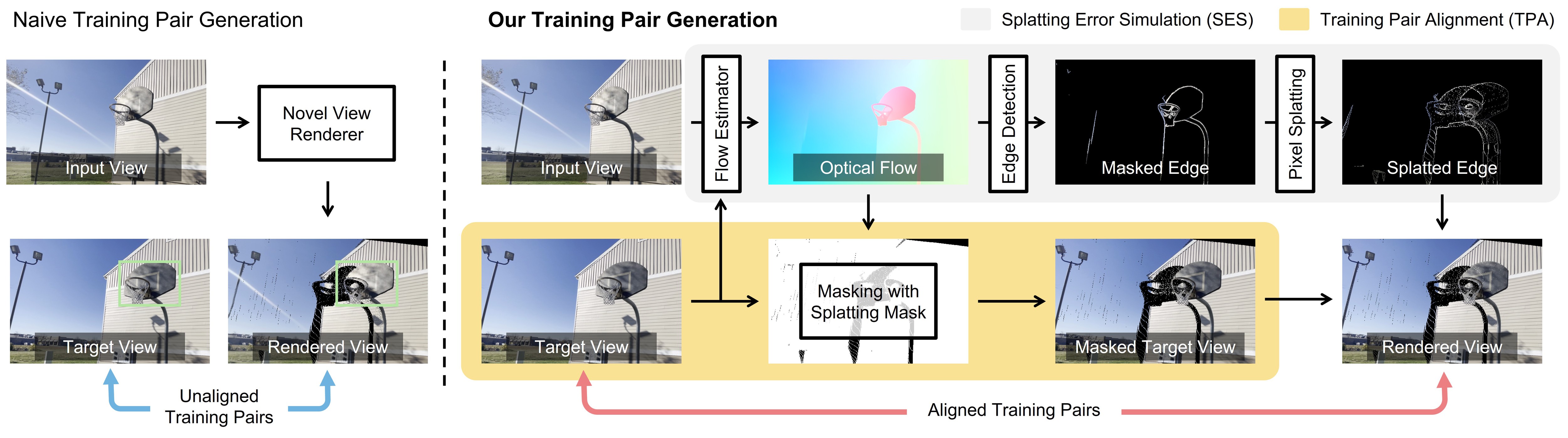}
    \caption{Comparison between naive and our training pair generation}
    \label{fig:main-alignsyn-pipeline}
    \end{subfigure}
    \begin{subfigure}[b]{\linewidth}
    \centering
    \begin{subfigure}{\imgWidth}
        \begin{tikzpicture}[spy using outlines={green,magnification=\ssmag,size=\ssizz},inner sep=0]
            \node [align=center, img] {\includegraphics[width=\textwidth]{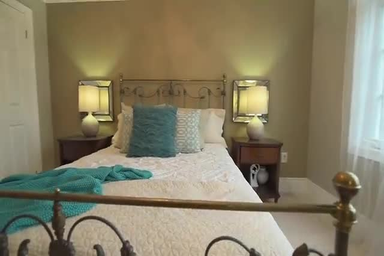}};
    	\end{tikzpicture}
     \caption*{Input View}
    \end{subfigure}
    \begin{subfigure}{\imgWidth}
		\begin{tikzpicture}[spy using outlines={green,magnification=\ssmag,size=\ssizz},inner sep=0]
            \node [align=center, img] {\includegraphics[width=\textwidth]{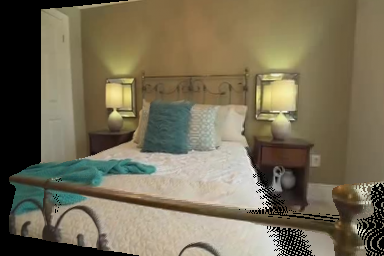}};
            \spy on \refour in node [left] at \rebigone;
    	\end{tikzpicture}
     \caption*{Splatted View}
    \end{subfigure}
    \begin{subfigure}{\imgWidth}
		\begin{tikzpicture}[spy using outlines={green,magnification=\ssmag,size=\ssizz},inner sep=0]
            \node [align=center, img] {\includegraphics[width=\textwidth]{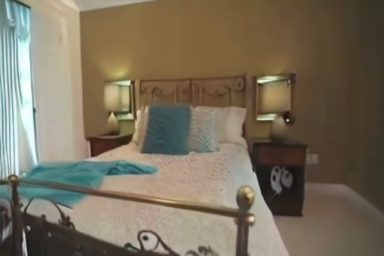}};
            \spy on \refour in node [left] at \rebigone;
    	\end{tikzpicture}
     \caption*{Naive Training}
    \end{subfigure}
    \begin{subfigure}{\imgWidth}
		\begin{tikzpicture}[spy using outlines={green,magnification=\ssmag,size=\ssizz},inner sep=0]
            \node [align=center, img] {\includegraphics[width=\textwidth]{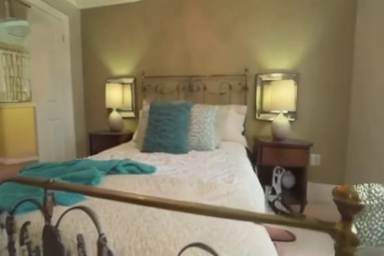}};
            \spy on \refour in node [left] at \rebigone;
    	\end{tikzpicture}
     \caption*{Training w/ TPA}
    \end{subfigure}
    \begin{subfigure}{\imgWidth}
        \begin{tikzpicture}[spy using outlines={green,magnification=\ssmag,size=\ssizz},inner sep=0]
            \node [align=center, img] {\includegraphics[width=\textwidth]{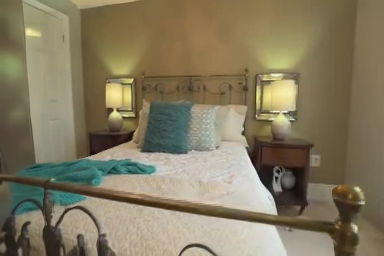}};
            \spy on \refour in node [left] at \rebigone;
    	\end{tikzpicture}
      \caption*{Training w/ TPA and SES}
      \end{subfigure}
    \caption{Results of the proposed training pair alignment (TPA) and splatting error simulation (SES)}
    \label{fig:main-alignsyn-result}
    \end{subfigure}
    \caption{\textbf{Aligned Synthesis.} Naive training pair generation often produces pairs that are locally unaligned in geometry and texture (green box in (a)), which results in unaligned novel view synthesis (naive training in (b)). By masking the target view, the proposed training pair alignment (TPA) enables aligned synthesis, but the generated contents tend to be misled by the splatting errors in the splatted view (training with TPA in (b)). Combining TPA with splatting error simulation (SES) in training, our \method\ learns to handle splatting errors and generates geometry- and texture-aligned novel views.}
    \label{fig:main-alignsyn}
\end{figure*}

%% file: figs/main/fig-texturebridge.tex
\begin{figure}
    \centering
    \includegraphics[width=\linewidth]{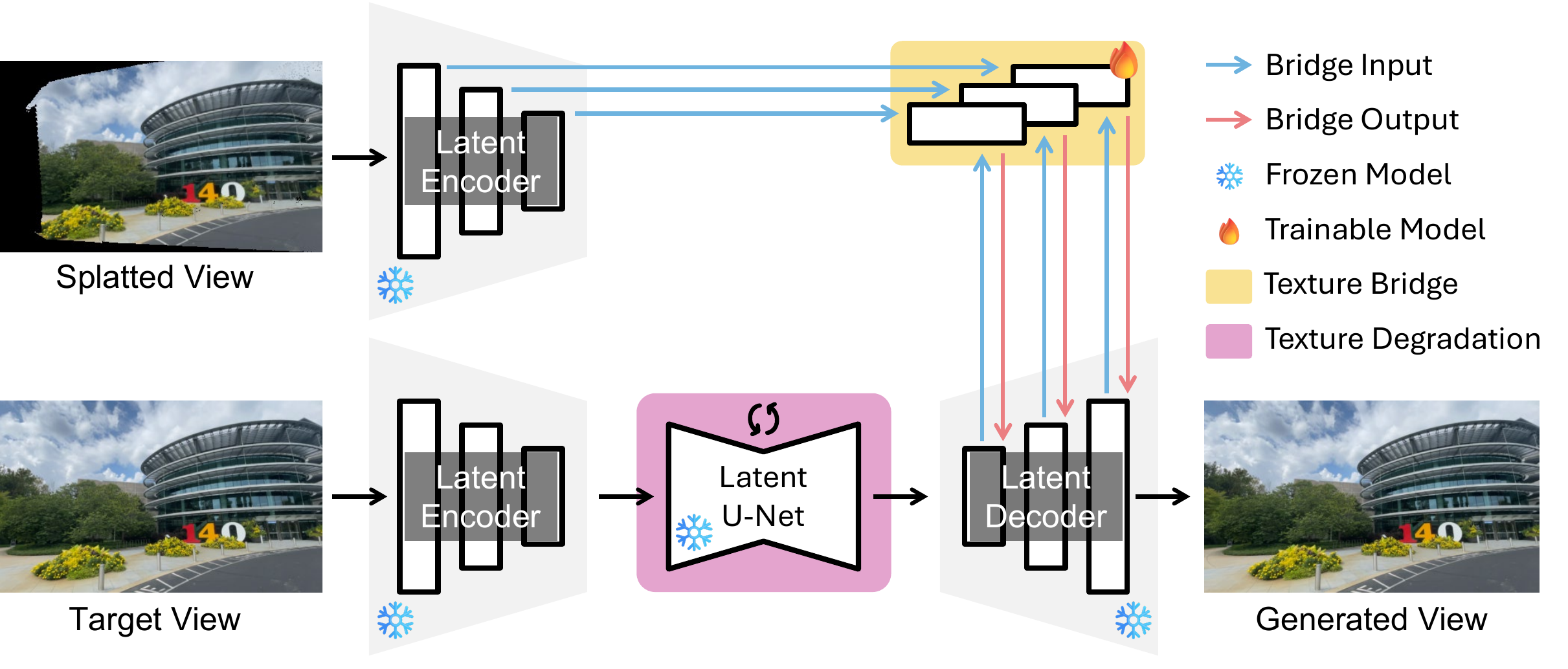}
    \caption{\textbf{Texture bridge.} The texture bridge is designed to complement multi-scale features from the encoder to the decoder, enabling high-fidelity synthesis. For training, we intentionally degrade the texture of the target view with the diffusion model, simulating a degraded view with hallucinated textures. This facilitates the texture bridge to learn how to adaptively fuse information from the splatted view and the diffusion outputs.}
    \label{fig:main-texturebridge}
\end{figure}

%% file: tabs/tab-main-singleview.tex
\begin{table*}[t]
\caption{\textbf{Quantitative evaluation of single-view novel view synthesis.} * denotes methods with two input views. \textcolor{red}{Best} and \textcolor{blue}{second-best} results are marked.}
\label{tab:main-singleview}
\setlength\tabcolsep{4pt}
\begin{tabular}{lcccccccccccc}
\toprule[0.15em]
                         & \multicolumn{1}{l}{} & \multicolumn{11}{c}{\textbf{RealEstate10K Dataset}}                                                                                                                                                                                                                                                                             \\ \cline{3-13} 
                         &                      & \multicolumn{5}{c}{Easy Set}                                                                                                                             &  & \multicolumn{5}{c}{Hard Set}                                                                                                                             \\ \cline{3-7} \cline{9-13} 
\multirow{-3}{*}{Method} &                      & PSNR $\uparrow$                         & SSIM $\uparrow$                         & LPIPS $\downarrow$                        & DISTS $\downarrow$                        & FID $\downarrow$                          &  & PSNR $\uparrow$                         & SSIM $\uparrow$                         & LPIPS $\downarrow$                        & DISTS $\downarrow$                        & FID $\downarrow$                          \\ \midrule[0.15em]
Syn-Sin                  &                      & -                            & -                            & -                            & -                            & -                            &  & 22.30                        & 0.740                        & -                            & -                            & -                            \\
SV-MPI                   &                      & 27.10                        & 0.870                        & -                            & -                            & -                            &  & 23.52                        & 0.785                        & -                            & -                            & -                            \\
BTS                      &                      & -                            & -                            & -                            & -                            & -                            &  & 24.00                        & 0.755                        & 0.194                        & -                            & -                            \\
Splatter Image           &                      & 28.15                        & 0.894                        & 0.110                        & -                            & -                            &  & 24.15                        & 0.810                        & 0.177                        & -                            & -                            \\
MINE                     &                      & 28.45                        & \textcolor{blue}{0.897}                        & 0.111                        & -                            & -                            &  & 24.75                        & \textcolor{blue}{0.820}                        & 0.179                        & -                            & -                            \\
AdaMPI                   &                      & 28.03                        & 0.892                        & 0.104                        & -                            & -                            &  & 23.54                        & 0.809                        & 0.184                        & -                            & -                            \\
GenWarp                  &                      & 16.94                        & 0.519                        & 0.318                        & 0.137                        & 12.27                        &  & 16.05                        & 0.488                        & 0.356                        & 0.149                        & 12.66                        \\
ViewCrafter              &                      & 20.61                        & 0.705                        & 0.242                        & 0.139                        & 13.71                        &  & 16.64                        & 0.588                        & 0.347                        & 0.185                        & 18.30                        \\
Flash3D                  &                      & {\color{blue} 28.46} & {\color{red} 0.899} & {\color{blue} 0.100} & {\color{blue} 0.062} & {\color{blue} 4.55}  &  & {\color{blue} 24.93} & {\color{red} 0.833} & {\color{blue} 0.160} & {\color{blue} 0.098} & {\color{blue} 8.42}  \\
\rowcolor{color3} \method\ (Ours)                     &                      & {\color{red} 28.53} & {\color{black} 0.895} & {\color{red} 0.096} & {\color{red} 0.059} & {\color{red} 3.96}  &  & {\color{red} 25.17} & {\color{blue} 0.820} & {\color{red} 0.154} & {\color{red} 0.088} & {\color{red} 6.04}  \\\midrule[0.15em]
                         &                      & \multicolumn{11}{c}{\textbf{DL3DV-10K Dataset}}                                                                                                                                                                                                                                                                                 \\ \cline{3-13} 
                         &                      & \multicolumn{5}{c}{Easy Set}                                                                                                                             &  & \multicolumn{5}{c}{Hard Set}                                                                                                                             \\ \cline{3-7} \cline{9-13} 
\multirow{-3}{*}{Method} &                      & PSNR $\uparrow$                         & SSIM $\uparrow$                         & LPIPS $\downarrow$                        & DISTS $\downarrow$                        & FID $\downarrow$                          &  & PSNR $\uparrow$                         & SSIM $\uparrow$                         & LPIPS $\downarrow$                        & DISTS $\downarrow$                        & FID $\downarrow$                          \\ \midrule[0.15em]
GenWarp                  &                      & 17.72                        & 0.463                        & 0.309                        & 0.357                        & 77.31                        &  & 16.74                        & 0.409                        & 0.350                        & 0.352                        & 81.58                        \\
ViewCrafter              &                      & 23.21                        & 0.694                        & 0.182                        & 0.349                        & \textcolor{blue}{46.60}                        &  & 19.77                        & 0.565                        & 0.249                        & 0.349                        & \textcolor{blue}{59.56}                        \\
DepthSplat*              &                      & {\color{blue} 24.37} & {\color{blue} 0.790} & {\color{blue} 0.168} & {\color{blue} 0.109} & {\color{black} 52.01} &  & {\color{blue} 21.79} & {\color{blue} 0.709} & {\color{blue} 0.221} & {\color{blue} 0.124} & {\color{black} 61.37} \\
\rowcolor{color3} \method\ (Ours)                     &                      & {\color{red} 26.14} & {\color{red} 0.826} & {\color{red} 0.113} & {\color{red} 0.068} & {\color{red} 24.23} &  & {\color{red} 22.48} & {\color{red} 0.732} & {\color{red} 0.181} & {\color{red} 0.092} & {\color{red} 41.22} \\ \bottomrule[0.15em]
\end{tabular}
\end{table*}

%% file: figs/main/fig-singleview.tex
\def\imgWidth{0.195\textwidth} %
\def\scc{(-1.9,-1.4)}
\def\rebigone{(-0.7, 0.62)} %
\def\reone{(1.05,0.4)} %
\def\retwo{(0.2,-0.22)} %
\def\rethree{(0.2,-0.6)} %
\def\refour{(-1.1,-0.4)} %

\def\dlbigone{(1.69, -0.45)} %
\def\dlone{(-0.82,0.38)} %
\def\dltwo{(0.4,0.52)} %
\def\dlthree{(0.02,0.45)} %
\def\dlfour{(0.15,0.65)} %
\def\ssizz{1cm} %
\def\ssmag{3}

\begin{figure*}[!t] 
\centering
\tikzstyle{img} = [rectangle, minimum width=\imgWidth, draw=black]
\begin{subfigure}{\linewidth}
    \begin{subfigure}{\imgWidth}
        \begin{tikzpicture}[spy using outlines={green,magnification=\ssmag,size=\ssizz},inner sep=0]
            \node [align=center, img] {\includegraphics[width=\textwidth]{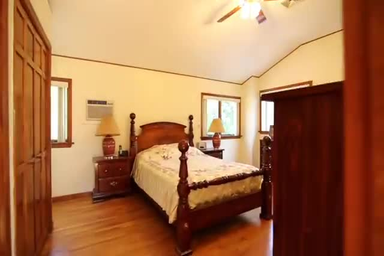}};
    	\end{tikzpicture}
    \end{subfigure}
    \begin{subfigure}{\imgWidth}
        \begin{tikzpicture}[spy using outlines={green,magnification=\ssmag,size=\ssizz},inner sep=0]
            \node [align=center, img] {\includegraphics[width=\textwidth]{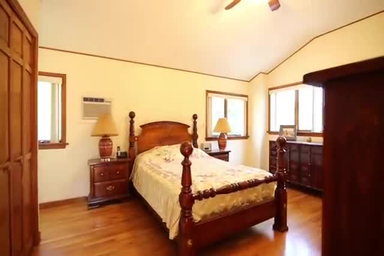}};
            \spy on \reone in node [left] at \rebigone;
    	\end{tikzpicture}
    \end{subfigure}
    \begin{subfigure}{\imgWidth}
		\begin{tikzpicture}[spy using outlines={green,magnification=\ssmag,size=\ssizz},inner sep=0]
            \node [align=center, img] {\includegraphics[width=\textwidth]{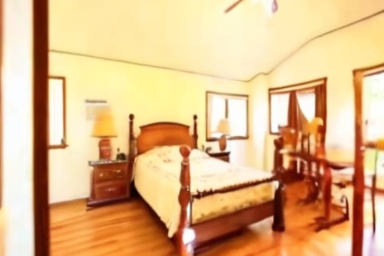}};
            \spy on \reone in node [left] at \rebigone;
    	\end{tikzpicture}
    \end{subfigure}
        \begin{subfigure}{\imgWidth}
        \begin{tikzpicture}[spy using outlines={green,magnification=\ssmag,size=\ssizz},inner sep=0]
            \node [align=center, img] {\includegraphics[width=\textwidth]{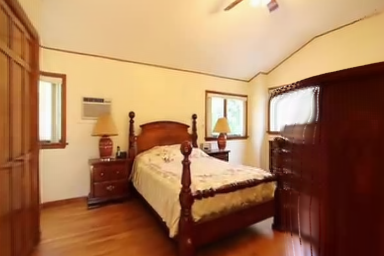}};
            \spy on \reone in node [left] at \rebigone;
    	\end{tikzpicture}
    \end{subfigure}
    \begin{subfigure}{\imgWidth}
        \begin{tikzpicture}[spy using outlines={green,magnification=\ssmag,size=\ssizz},inner sep=0]
            \node [align=center, img] {\includegraphics[width=\textwidth]{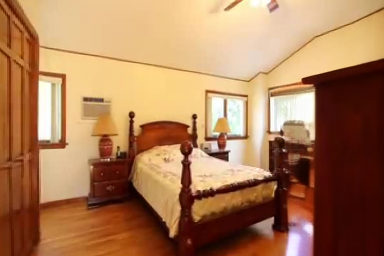}};
            \spy on \reone in node [left] at \rebigone;
    	\end{tikzpicture}
    \end{subfigure}
      \\ %
    \begin{subfigure}{\imgWidth}
        \begin{tikzpicture}[spy using outlines={green,magnification=\ssmag,size=\ssizz},inner sep=0]
            \node [align=center, img] {\includegraphics[width=\textwidth]{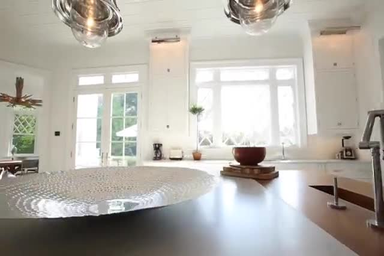}};
    	\end{tikzpicture}
    \end{subfigure}
    \begin{subfigure}{\imgWidth}
        \begin{tikzpicture}[spy using outlines={green,magnification=\ssmag,size=\ssizz},inner sep=0]
            \node [align=center, img] {\includegraphics[width=\textwidth]{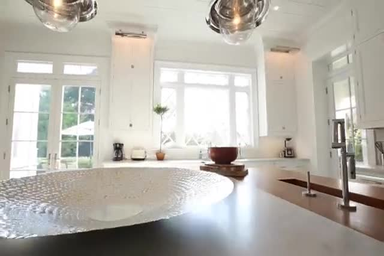}};
            \spy on \retwo in node [left] at \rebigone;
    	\end{tikzpicture}
    \end{subfigure}
    \begin{subfigure}{\imgWidth}
		\begin{tikzpicture}[spy using outlines={green,magnification=\ssmag,size=\ssizz},inner sep=0]
            \node [align=center, img] {\includegraphics[width=\textwidth]{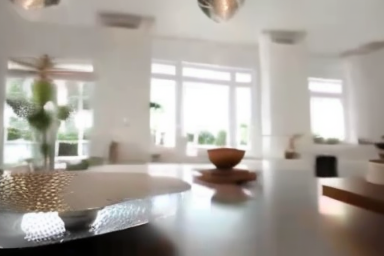}};
            \spy on \retwo in node [left] at \rebigone;
    	\end{tikzpicture}
    \end{subfigure}
        \begin{subfigure}{\imgWidth}
        \begin{tikzpicture}[spy using outlines={green,magnification=\ssmag,size=\ssizz},inner sep=0]
            \node [align=center, img] {\includegraphics[width=\textwidth]{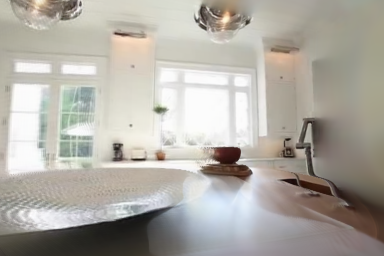}};
            \spy on \retwo in node [left] at \rebigone;
    	\end{tikzpicture}
    \end{subfigure}
    \begin{subfigure}{\imgWidth}
        \begin{tikzpicture}[spy using outlines={green,magnification=\ssmag,size=\ssizz},inner sep=0]
            \node [align=center, img] {\includegraphics[width=\textwidth]{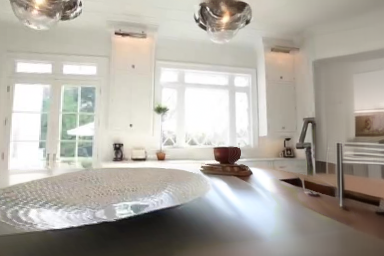}};
            \spy on \retwo in node [left] at \rebigone;
    	\end{tikzpicture}
    \end{subfigure}
      \\ %
    \begin{subfigure}{\imgWidth}
        \begin{tikzpicture}[spy using outlines={green,magnification=\ssmag,size=\ssizz},inner sep=0]
            \node [align=center, img] {\includegraphics[width=\textwidth]{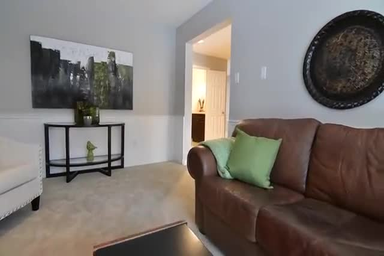}};
    	\end{tikzpicture}
    \end{subfigure}
    \begin{subfigure}{\imgWidth}
        \begin{tikzpicture}[spy using outlines={green,magnification=\ssmag,size=\ssizz},inner sep=0]
            \node [align=center, img] {\includegraphics[width=\textwidth]{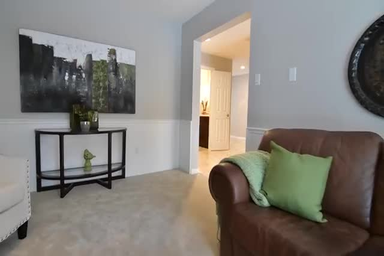}};
            \spy on \rethree in node [left] at \rebigone;
    	\end{tikzpicture}
    \end{subfigure}
    \begin{subfigure}{\imgWidth}
		\begin{tikzpicture}[spy using outlines={green,magnification=\ssmag,size=\ssizz},inner sep=0]
            \node [align=center, img] {\includegraphics[width=\textwidth]{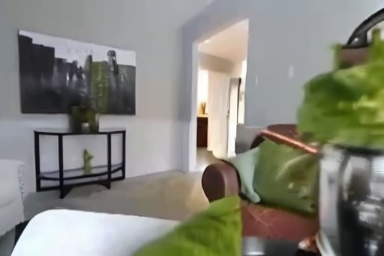}};
            \spy on \rethree in node [left] at \rebigone;
    	\end{tikzpicture}
    \end{subfigure}
        \begin{subfigure}{\imgWidth}
        \begin{tikzpicture}[spy using outlines={green,magnification=\ssmag,size=\ssizz},inner sep=0]
            \node [align=center, img] {\includegraphics[width=\textwidth]{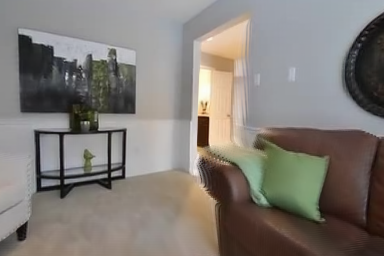}};
            \spy on \rethree in node [left] at \rebigone;
    	\end{tikzpicture}
    \end{subfigure}
    \begin{subfigure}{\imgWidth}
        \begin{tikzpicture}[spy using outlines={green,magnification=\ssmag,size=\ssizz},inner sep=0]
            \node [align=center, img] {\includegraphics[width=\textwidth]{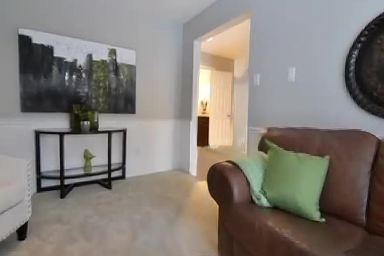}};
            \spy on \rethree in node [left] at \rebigone;
    	\end{tikzpicture}
    \end{subfigure}
      \\ %
          \begin{subfigure}{\imgWidth}
        \begin{tikzpicture}[spy using outlines={green,magnification=\ssmag,size=\ssizz},inner sep=0]
            \node [align=center, img] {\includegraphics[width=\textwidth]{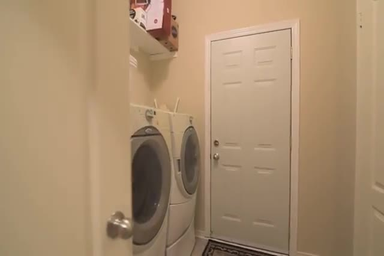}};
    	\end{tikzpicture}
     \caption*{Input View}
    \end{subfigure}
    \begin{subfigure}{\imgWidth}
        \begin{tikzpicture}[spy using outlines={green,magnification=\ssmag,size=\ssizz},inner sep=0]
            \node [align=center, img] {\includegraphics[width=\textwidth]{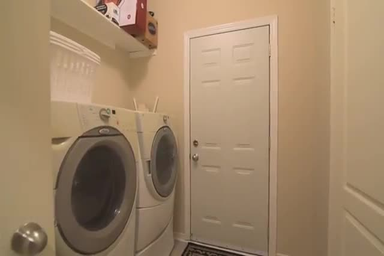}};
            \spy on \refour in node [left] at \rebigone;
    	\end{tikzpicture}
     \caption*{Target View}
    \end{subfigure}
    \begin{subfigure}{\imgWidth}
		\begin{tikzpicture}[spy using outlines={green,magnification=\ssmag,size=\ssizz},inner sep=0]
            \node [align=center, img] {\includegraphics[width=\textwidth]{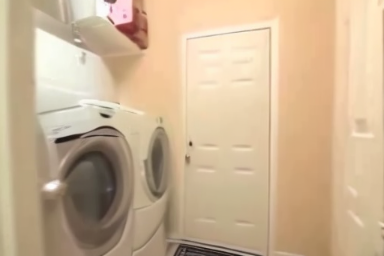}};
            \spy on \refour in node [left] at \rebigone;
    	\end{tikzpicture}
     \caption*{ViewCrafter}
    \end{subfigure}
        \begin{subfigure}{\imgWidth}
        \begin{tikzpicture}[spy using outlines={green,magnification=\ssmag,size=\ssizz},inner sep=0]
            \node [align=center, img] {\includegraphics[width=\textwidth]{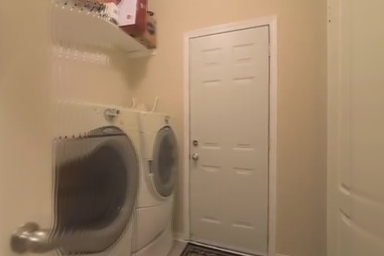}};
            \spy on \refour in node [left] at \rebigone;
    	\end{tikzpicture}
      \caption*{Flash3D}
    \end{subfigure}
    \begin{subfigure}{\imgWidth}
        \begin{tikzpicture}[spy using outlines={green,magnification=\ssmag,size=\ssizz},inner sep=0]
            \node [align=center, img] {\includegraphics[width=\textwidth]{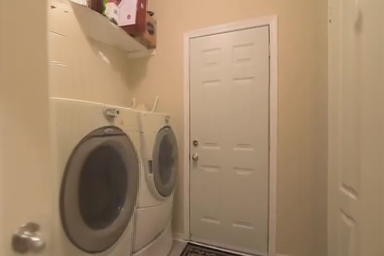}};
            \spy on \refour in node [left] at \rebigone;
    	\end{tikzpicture}
      \caption*{\textbf{\method\ (Ours)}}
    \end{subfigure}
	\caption{Visual comparisons on the RealEstate10K dataset}
 \label{fig:main-singleview-re10k}
 \end{subfigure}
\begin{subfigure}{\linewidth}
    \begin{subfigure}{\imgWidth}
        \begin{tikzpicture}[spy using outlines={green,magnification=\ssmag,size=\ssizz},inner sep=0]
            \node [align=center, img] {\includegraphics[width=\textwidth]{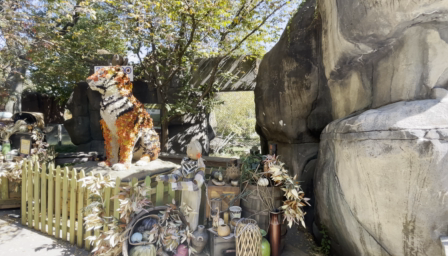}};
    	\end{tikzpicture}
    \end{subfigure}
    \begin{subfigure}{\imgWidth}
        \begin{tikzpicture}[spy using outlines={green,magnification=\ssmag,size=\ssizz},inner sep=0]
            \node [align=center, img] {\includegraphics[width=\textwidth]{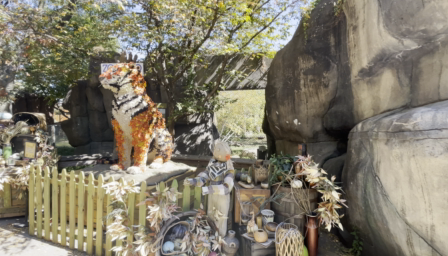}};
            \spy on \dlone in node [left] at \dlbigone;
    	\end{tikzpicture}
    \end{subfigure}
    \begin{subfigure}{\imgWidth}
		\begin{tikzpicture}[spy using outlines={green,magnification=\ssmag,size=\ssizz},inner sep=0]
            \node [align=center, img] {\includegraphics[width=\textwidth]{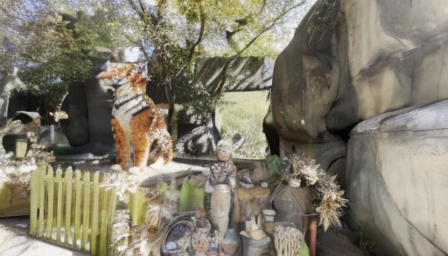}};
            \spy on \dlone in node [left] at \dlbigone;
    	\end{tikzpicture}
    \end{subfigure}
        \begin{subfigure}{\imgWidth}
        \begin{tikzpicture}[spy using outlines={green,magnification=\ssmag,size=\ssizz},inner sep=0]
            \node [align=center, img] {\includegraphics[width=\textwidth]{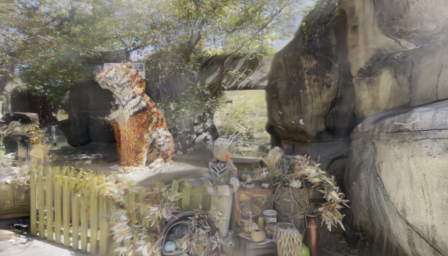}};
            \spy on \dlone in node [left] at \dlbigone;
    	\end{tikzpicture}
    \end{subfigure}
    \begin{subfigure}{\imgWidth}
        \begin{tikzpicture}[spy using outlines={green,magnification=\ssmag,size=\ssizz},inner sep=0]
            \node [align=center, img] {\includegraphics[width=\textwidth]{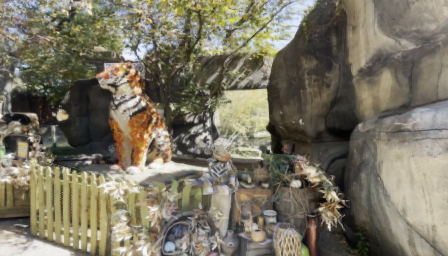}};
            \spy on \dlone in node [left] at \dlbigone;
    	\end{tikzpicture}
    \end{subfigure}
      \\ %
    \begin{subfigure}{\imgWidth}
        \begin{tikzpicture}[spy using outlines={green,magnification=\ssmag,size=\ssizz},inner sep=0]
            \node [align=center, img] {\includegraphics[width=\textwidth]{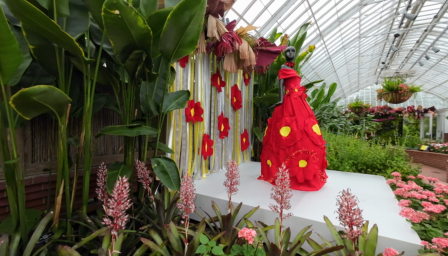}};
    	\end{tikzpicture}
    \end{subfigure}
    \begin{subfigure}{\imgWidth}
        \begin{tikzpicture}[spy using outlines={green,magnification=\ssmag,size=\ssizz},inner sep=0]
            \node [align=center, img] {\includegraphics[width=\textwidth]{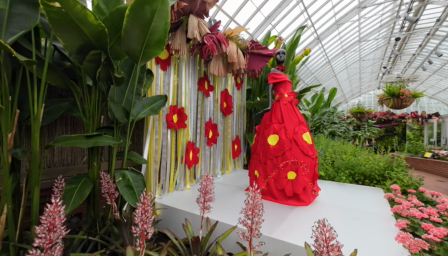}};
            \spy on \dltwo in node [left] at \dlbigone;
    	\end{tikzpicture}
    \end{subfigure}
    \begin{subfigure}{\imgWidth}
		\begin{tikzpicture}[spy using outlines={green,magnification=\ssmag,size=\ssizz},inner sep=0]
            \node [align=center, img] {\includegraphics[width=\textwidth]{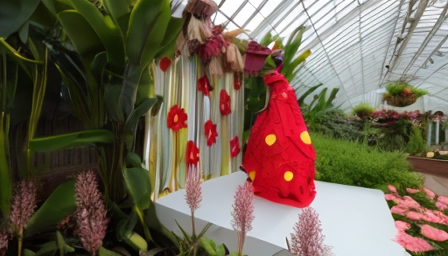}};
            \spy on \dltwo in node [left] at \dlbigone;
    	\end{tikzpicture}
    \end{subfigure}
        \begin{subfigure}{\imgWidth}
        \begin{tikzpicture}[spy using outlines={green,magnification=\ssmag,size=\ssizz},inner sep=0]
            \node [align=center, img] {\includegraphics[width=\textwidth]{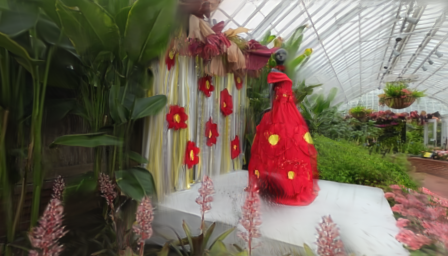}};
            \spy on \dltwo in node [left] at \dlbigone;
    	\end{tikzpicture}
    \end{subfigure}
    \begin{subfigure}{\imgWidth}
        \begin{tikzpicture}[spy using outlines={green,magnification=\ssmag,size=\ssizz},inner sep=0]
            \node [align=center, img] {\includegraphics[width=\textwidth]{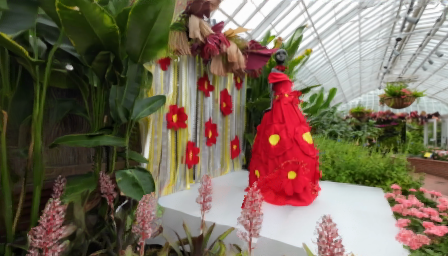}};
            \spy on \dltwo in node [left] at \dlbigone;
    	\end{tikzpicture}
    \end{subfigure}
      \\ %
    \begin{subfigure}{\imgWidth}
        \begin{tikzpicture}[spy using outlines={green,magnification=\ssmag,size=\ssizz},inner sep=0]
            \node [align=center, img] {\includegraphics[width=\textwidth]{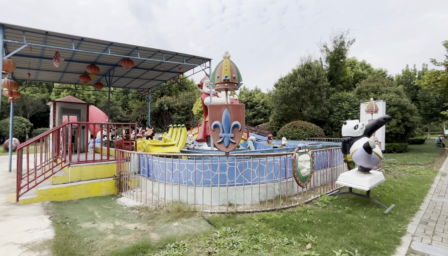}};
    	\end{tikzpicture}
    \end{subfigure}
    \begin{subfigure}{\imgWidth}
        \begin{tikzpicture}[spy using outlines={green,magnification=\ssmag,size=\ssizz},inner sep=0]
            \node [align=center, img] {\includegraphics[width=\textwidth]{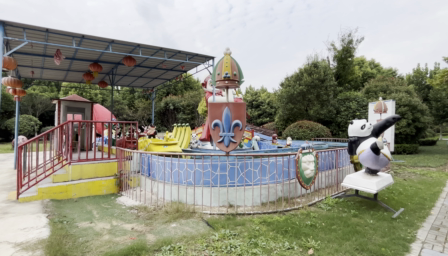}};
            \spy on \dlthree in node [left] at \dlbigone;
    	\end{tikzpicture}
    \end{subfigure}
    \begin{subfigure}{\imgWidth}
		\begin{tikzpicture}[spy using outlines={green,magnification=\ssmag,size=\ssizz},inner sep=0]
            \node [align=center, img] {\includegraphics[width=\textwidth]{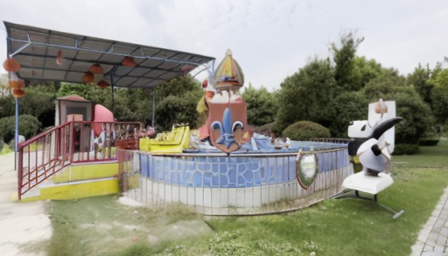}};
            \spy on \dlthree in node [left] at \dlbigone;
    	\end{tikzpicture}
    \end{subfigure}
        \begin{subfigure}{\imgWidth}
        \begin{tikzpicture}[spy using outlines={green,magnification=\ssmag,size=\ssizz},inner sep=0]
            \node [align=center, img] {\includegraphics[width=\textwidth]{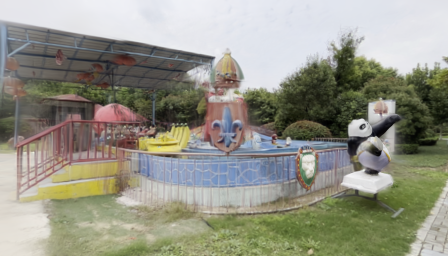}};
            \spy on \dlthree in node [left] at \dlbigone;
    	\end{tikzpicture}
    \end{subfigure}
    \begin{subfigure}{\imgWidth}
        \begin{tikzpicture}[spy using outlines={green,magnification=\ssmag,size=\ssizz},inner sep=0]
            \node [align=center, img] {\includegraphics[width=\textwidth]{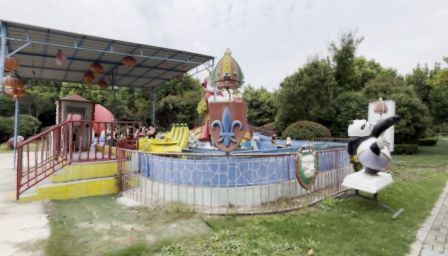}};
            \spy on \dlthree in node [left] at \dlbigone;
    	\end{tikzpicture}
    \end{subfigure}
      \\ %
          \begin{subfigure}{\imgWidth}
        \begin{tikzpicture}[spy using outlines={green,magnification=\ssmag,size=\ssizz},inner sep=0]
            \node [align=center, img] {\includegraphics[width=\textwidth]{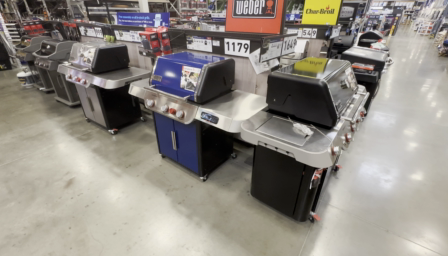}};
    	\end{tikzpicture}
     \caption*{Input View}
    \end{subfigure}
    \begin{subfigure}{\imgWidth}
        \begin{tikzpicture}[spy using outlines={green,magnification=\ssmag,size=\ssizz},inner sep=0]
            \node [align=center, img] {\includegraphics[width=\textwidth]{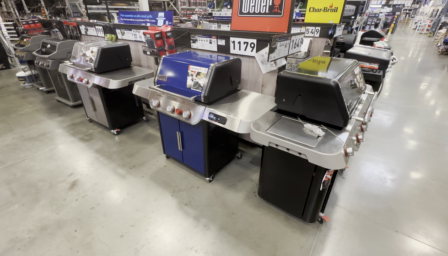}};
            \spy on \dlfour in node [left] at \dlbigone;
    	\end{tikzpicture}
     \caption*{Target View}
    \end{subfigure}
    \begin{subfigure}{\imgWidth}
		\begin{tikzpicture}[spy using outlines={green,magnification=\ssmag,size=\ssizz},inner sep=0]
            \node [align=center, img] {\includegraphics[width=\textwidth]{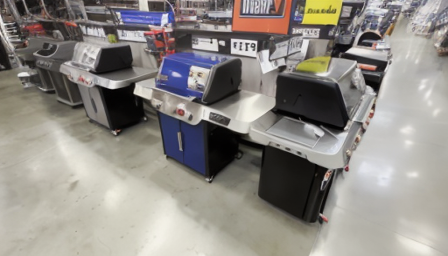}};
            \spy on \dlfour in node [left] at \dlbigone;
    	\end{tikzpicture}
     \caption*{ViewCrafter}
    \end{subfigure}
        \begin{subfigure}{\imgWidth}
        \begin{tikzpicture}[spy using outlines={green,magnification=\ssmag,size=\ssizz},inner sep=0]
            \node [align=center, img] {\includegraphics[width=\textwidth]{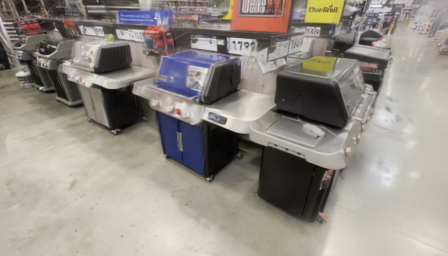}};
            \spy on \dlfour in node [left] at \dlbigone;
    	\end{tikzpicture}
      \caption*{DepthSplat*}
    \end{subfigure}
    \begin{subfigure}{\imgWidth}
        \begin{tikzpicture}[spy using outlines={green,magnification=\ssmag,size=\ssizz},inner sep=0]
            \node [align=center, img] {\includegraphics[width=\textwidth]{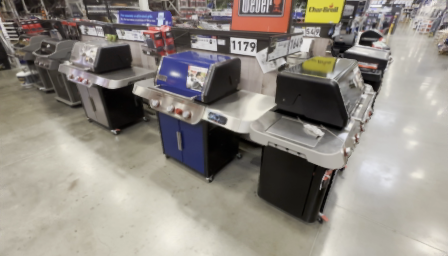}};
            \spy on \dlfour in node [left] at \dlbigone;
    	\end{tikzpicture}
      \caption*{\textbf{\method\ (Ours)}}
    \end{subfigure}
	\caption{Visual comparisons on the DL3DV-10K dataset}
 \label{fig:main-singleview-dl3dv}
 \end{subfigure}
 \caption{\textbf{Qualitative comparisons on the single-view novel view synthesis task.} * indicates that the method uses two views as input. }
 \label{fig:main-singleview}
\end{figure*}

%% file: figs/main/fig-singleview-wild.tex
\def\imgWidth{0.238\textwidth} %
\def\scc{(-1.9,-1.4)}
\def\rebigone{(2.1, -0.65)} %
\def\reone{(0.5,0.1)} %
\def\retwo{(-0.1,0.05)} %
\def\rethree{(0.6,0)} %
\def\refour{(1.4,0.75)} %

\def\ssizz{1cm} %
\def\ssmag{3}

\begin{figure*}[!t] 
\centering
\tikzstyle{img} = [rectangle, minimum width=\imgWidth, draw=black]
\centering
\begin{subfigure}{\textwidth}
\centering
    \begin{subfigure}{\imgWidth}
        \begin{tikzpicture}[spy using outlines={green,magnification=\ssmag,size=\ssizz},inner sep=0]
            \node [align=center, img] {\includegraphics[width=\textwidth]{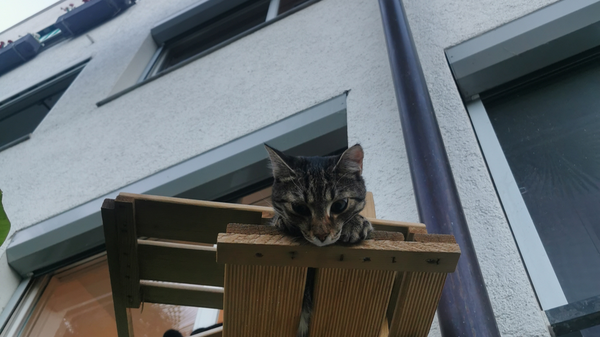}};
    	\end{tikzpicture}
    \end{subfigure}
    \begin{subfigure}{\imgWidth}
        \begin{tikzpicture}[spy using outlines={green,magnification=\ssmag,size=\ssizz},inner sep=0]
            \node [align=center, img] {\includegraphics[width=\textwidth]{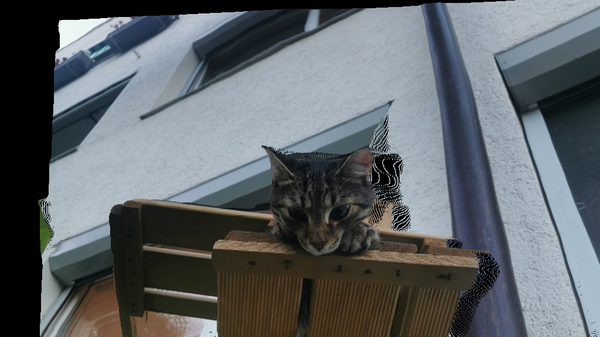}};
            \spy on \reone in node [left] at \rebigone;
    	\end{tikzpicture}
    \end{subfigure}
    \begin{subfigure}{\imgWidth}
		\begin{tikzpicture}[spy using outlines={green,magnification=\ssmag,size=\ssizz},inner sep=0]
            \node [align=center, img] {\includegraphics[width=\textwidth]{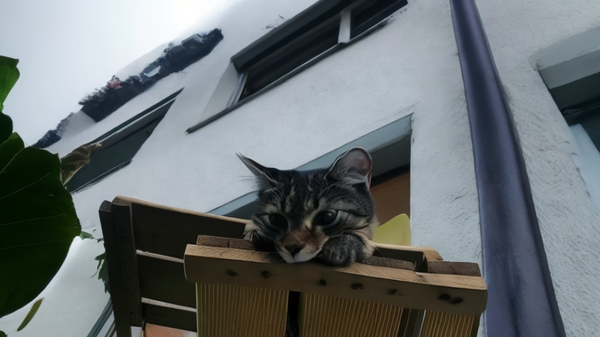}};
            \spy on \reone in node [left] at \rebigone;
    	\end{tikzpicture}
    \end{subfigure}
        \begin{subfigure}{\imgWidth}
        \begin{tikzpicture}[spy using outlines={green,magnification=\ssmag,size=\ssizz},inner sep=0]
            \node [align=center, img] {\includegraphics[width=\textwidth]{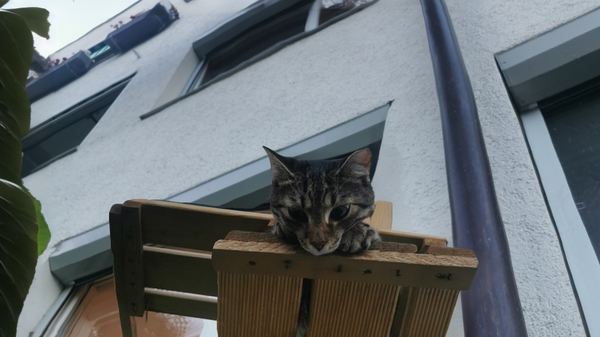}};
            \spy on \reone in node [left] at \rebigone;
    	\end{tikzpicture}
    \end{subfigure}
      \\ %
    \begin{subfigure}{\imgWidth}
        \begin{tikzpicture}[spy using outlines={green,magnification=\ssmag,size=\ssizz},inner sep=0]
            \node [align=center, img] {\includegraphics[width=\textwidth]{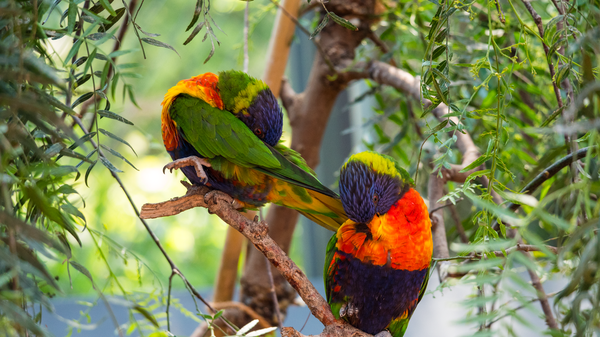}};
    	\end{tikzpicture}
    \end{subfigure}
    \begin{subfigure}{\imgWidth}
        \begin{tikzpicture}[spy using outlines={green,magnification=\ssmag,size=\ssizz},inner sep=0]
            \node [align=center, img] {\includegraphics[width=\textwidth]{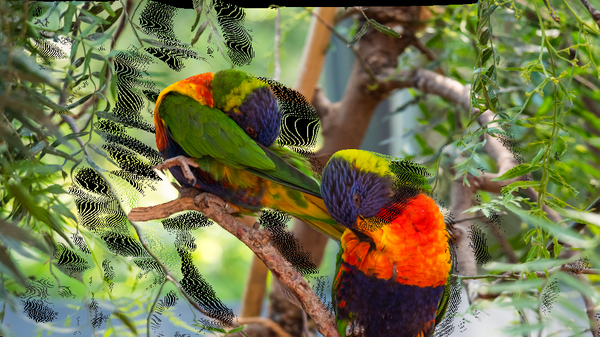}};
            \spy on \rethree in node [left] at \rebigone;
    	\end{tikzpicture}
    \end{subfigure}
    \begin{subfigure}{\imgWidth}
		\begin{tikzpicture}[spy using outlines={green,magnification=\ssmag,size=\ssizz},inner sep=0]
            \node [align=center, img] {\includegraphics[width=\textwidth]{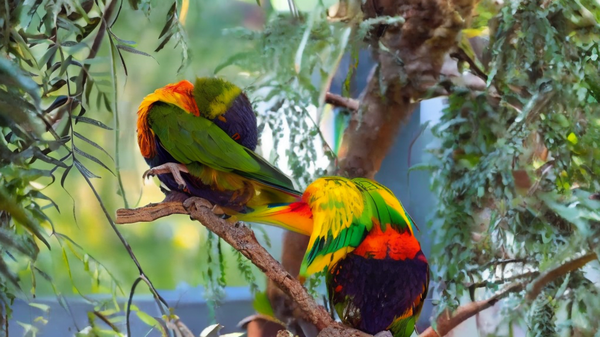}};
            \spy on \rethree in node [left] at \rebigone;
    	\end{tikzpicture}
    \end{subfigure}
        \begin{subfigure}{\imgWidth}
        \begin{tikzpicture}[spy using outlines={green,magnification=\ssmag,size=\ssizz},inner sep=0]
            \node [align=center, img] {\includegraphics[width=\textwidth]{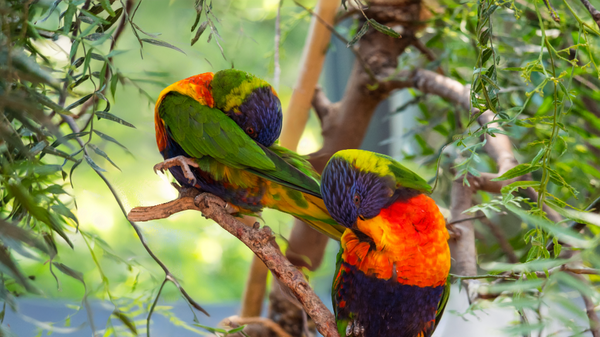}};
            \spy on \rethree in node [left] at \rebigone;
    	\end{tikzpicture}
    \end{subfigure}
      \\ %
          \begin{subfigure}{\imgWidth}
        \begin{tikzpicture}[spy using outlines={green,magnification=\ssmag,size=\ssizz},inner sep=0]
            \node [align=center, img] {\includegraphics[width=\textwidth]{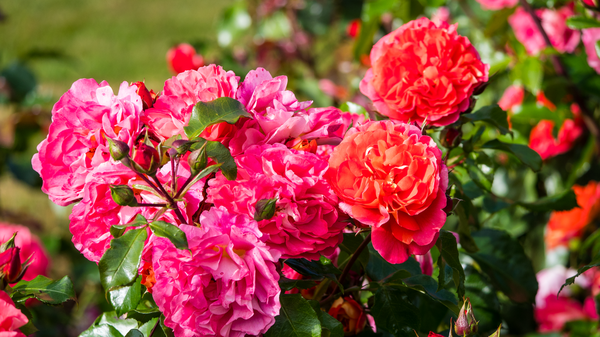}};
    	\end{tikzpicture}
     \caption*{Input View}
    \end{subfigure}
    \begin{subfigure}{\imgWidth}
        \begin{tikzpicture}[spy using outlines={green,magnification=\ssmag,size=\ssizz},inner sep=0]
            \node [align=center, img] {\includegraphics[width=\textwidth]{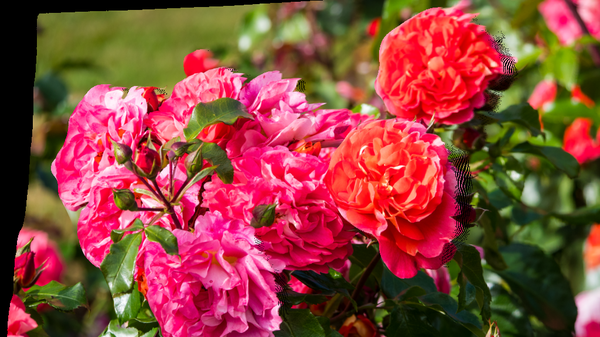}};
            \spy on \refour in node [left] at \rebigone;
    	\end{tikzpicture}
     \caption*{Splatted View}
    \end{subfigure}
    \begin{subfigure}{\imgWidth}
		\begin{tikzpicture}[spy using outlines={green,magnification=\ssmag,size=\ssizz},inner sep=0]
            \node [align=center, img] {\includegraphics[width=\textwidth]{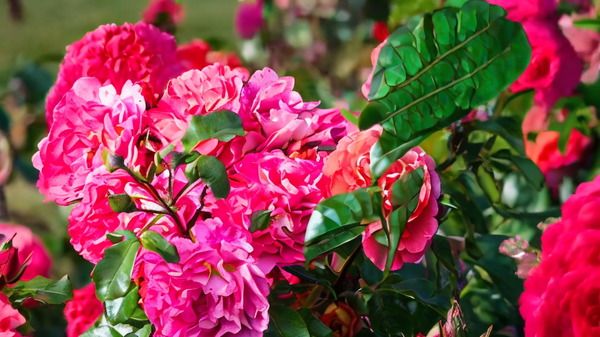}};
            \spy on \refour in node [left] at \rebigone;
    	\end{tikzpicture}
     \caption*{ViewCrafter}
    \end{subfigure}
    \begin{subfigure}{\imgWidth}
        \begin{tikzpicture}[spy using outlines={green,magnification=\ssmag,size=\ssizz},inner sep=0]
            \node [align=center, img] {\includegraphics[width=\textwidth]{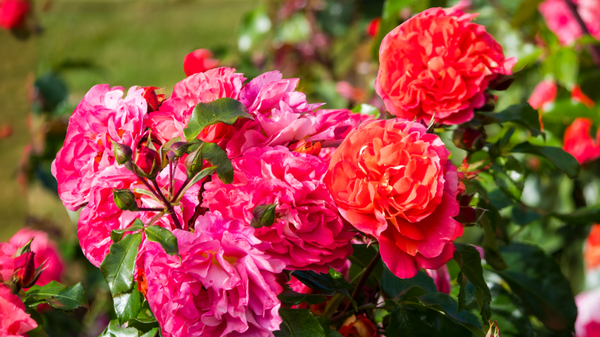}};
            \spy on \refour in node [left] at \rebigone;
    	\end{tikzpicture}
      \caption*{\textbf{\method\ (Ours)}}
    \end{subfigure}
    \end{subfigure}
    \caption{\textbf{Qualitative comparisons on in-the-wild high-resolution (576$\times$1024) samples.} }
 \label{fig:main-singleview-wild}
\end{figure*}

%% file: tabs/tab-main-ablation.tex
\begin{table}[t]
\caption{\textbf{Ablation on single-view novel view synthesis.} TPA, SES, TB, and TD represent training pair alignment, splatting error simulation, texture bridge, and texture degradation. \textcolor{red}{Best} and \textcolor{blue}{second-best} results are marked.}
\label{tab:main-ablation}
\setlength\tabcolsep{2.1pt}
\begin{tabular}{cccccccccccc}
\toprule[0.15em]
                        &  &  &     &  &    &  & \multicolumn{5}{c}{\textbf{DL3DV-10K Dataset}}                                                                                                                                 \\ \cline{8-12} 
\multirow{-2}{*}{ID} &  & \multirow{-2}{*}{TPA}     & \multirow{-2}{*}{SES}   & \multirow{-2}{*}{TB} & \multirow{-2}{*}{TD} &  & PSNR $\uparrow$                         & SSIM $\uparrow$                          & LPIPS $\downarrow$                        & DISTS $\downarrow$                        & FID $\downarrow$                         \\ \midrule[0.15em]
\#1                     &                &                  &               &            &             &  & 23.21                        & 0.694                        & 0.182                        & 0.349                        & 46.60                        \\
\#2                     &  & $\checkmark$                          &              &            &             &  & 24.90                        & 0.749                        & 0.178                        & 0.100                        & 43.97                        \\
\#3                     &  & $\checkmark$         & $\checkmark$             &              &                   &  & 25.00                        & 0.749                        & 0.178                        & 0.098                        & 43.59                        \\
\#4                     &  & $\checkmark$         & $\checkmark$             & $\checkmark$            &             &  & {\color{blue} 26.05} & {\color{blue} 0.825} & {\color{blue} 0.118} & {\color{blue} 0.070} & {\color{blue} 25.49} \\
\#5                     &  & $\checkmark$         & $\checkmark$             & $\checkmark$        & $\checkmark$        &  & {\color{red} 26.14} & {\color{red} 0.826} & {\color{red} 0.113} & {\color{red} 0.068} & {\color{red} 24.23} \\ \bottomrule[0.15em]
\end{tabular}
\vspace{-0.5em}
\end{table}

%% file: tabs/tab-main-sparseview.tex
\begin{table}[t]%
\caption{\textbf{Quantitative evaluation of sparse-view novel view synthesis.} $^\dagger$ indicates the averaged results of the two generated novel views. The \textcolor{red}{best} and \textcolor{blue}{second-best} results are marked.}
\label{tab:main-sparseview}
\setlength\tabcolsep{3.2pt}
\begin{tabular}{lcccccc}
\toprule[0.15em]
                         &                      & \multicolumn{5}{c}{\textbf{DL3DV-10K Dataset (In-domain)}}                                                                                                \\ \cline{3-7} 
\multirow{-2}{*}{Method} &                      & PSNR $\uparrow$                         & SSIM $\uparrow$                         & LPIPS $\downarrow$                        & DISTS $\downarrow$                        & FID $\downarrow$                           \\ \midrule[0.15em]
MVSplat                  &                      & 17.54                        & 0.529                        & 0.402                        & -                            & -                             \\
DepthSplat               &                      & {\color{blue} 19.05} & {\color{blue} 0.610} & {\color{blue} 0.313} & {\color{blue} 0.163} & {\color{blue} 105.09} \\
\rowcolor{color3} \method\ (Ours)                     &                      & {\color{red} 21.42} & {\color{red} 0.619} & {\color{red} 0.294} & {\color{red} 0.130} & {\color{red} 71.26}  \\ \midrule[0.15em]
                         &                      & \multicolumn{5}{c}{\textbf{DTU Dataset (Cross-domain)}}                                                                                                   \\ \cline{3-7} 
\multirow{-2}{*}{Method} &                      & PSNR $\uparrow$                         & SSIM $\uparrow$                         & LPIPS $\downarrow$                        & DISTS $\downarrow$                        & FID $\downarrow$                           \\ \midrule[0.15em]
pixelSplat               &                      & 12.89                        & 0.382                        & 0.560                        & -                            & -                             \\
MVSplat                  &                      & 13.94                        & 0.473                        & 0.385                        & -                            & -                             \\
TranSplat                & \multicolumn{1}{l}{} & 14.93                        & 0.531                        & 0.326                        & -                            & -                             \\
DepthSplat               &                      & {\color{blue} 16.01} & {\color{blue} 0.612} & 0.334                        & 0.201                        & 130.75                        \\
\rowcolor{color3} \method\ (Ours)                     & \multicolumn{1}{l}{} & 15.96                        & 0.590                        & {\color{red} 0.264} & {\color{red} 0.147} & {\color{red} 82.45}  \\
\rowcolor{color3} \method$^\dagger$ (Ours)                    &                      & {\color{red} 16.33} & {\color{red} 0.616} & {\color{blue} 0.285} & {\color{blue} 0.164} & {\color{blue} 98.98}  \\ \bottomrule[0.15em]
\end{tabular}
\vspace{-0.5em}
\end{table}

%% file: tabs/tab-main-stereoconv.tex
\begin{table}[t]%
\caption{\textbf{Quantitative evaluation of stereo video conversion.} The \textcolor{red}{best} and \textcolor{blue}{second-best} results are marked.}
\label{tab:main-stereoconv}
\setlength\tabcolsep{3.5pt}
\begin{tabular}{lcccccc}
\toprule[0.15em]
                         &  & \multicolumn{5}{c}{\textbf{Spring Dataset}}                                                                                                                        \\ \cline{3-7} 
\multirow{-2}{*}{Method} &  & PSNR $\uparrow$                        & SSIM $\uparrow$                        & LPIPS $\downarrow$                       & DISTS $\downarrow$                       & FID $\downarrow$                          \\ \midrule[0.15em]
ViewCrafter              &  & 18.41                        & 0.526                        & 0.266                        & 0.187                        & 143.31                        \\
StereoCrafter            &  & {\color{blue} 26.46} & {\color{blue} 0.765} & {\color{blue} 0.192} & {\color{blue} 0.175} & {\color{blue} 135.86} \\
\rowcolor{color3} \method\ (Ours)                     &  & {\color{red} 30.12} & {\color{red} 0.908} & {\color{red} 0.073} & {\color{red} 0.070} & {\color{red} 46.37}  \\ \bottomrule[0.15em]
\end{tabular}
\vspace{-0.5em}
\end{table}

%% file: figs/main/fig-sparseview.tex
\def\imgWidth{0.279\textwidth} %
\def\scc{(-1.9,-1.4)}
\def\rebigone{(-0.92, -0.55)} %
\def\reone{(1.25,0.48)} %
\def\retwo{(-0.82,0.55)} %
\def\rethree{(1,0.38)} %
\def\refour{(-1.1,-0.4)} %

\def\dlbigone{(0.85, -0.55)} %
\def\dlone{(0.5,0)} %
\def\dltwo{(0.78,0.01)} %
\def\dlthree{(0.75,0.02)} %
\def\dlfour{(0.15,0.65)} %
\def\ssizz{0.6cm} %
\def\ssmag{3}

\begin{figure*}[!t] 
\centering
\tikzstyle{img} = [rectangle, minimum width=\imgWidth, draw=black]
\begin{subfigure}{0.62\linewidth}
\centering
    \begin{subfigure}{0.1355\textwidth}
        \begin{tikzpicture}[spy using outlines={green,magnification=\ssmag,size=\ssizz},inner sep=0]
            \node [align=center, img] {\includegraphics[width=\textwidth]{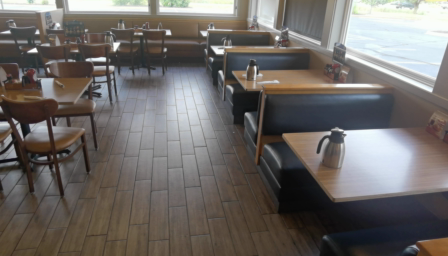}};
    	\end{tikzpicture}
        \\
        \begin{tikzpicture}[spy using outlines={green,magnification=\ssmag,size=\ssizz},inner sep=0]
            \node [align=center, img] {\includegraphics[width=\textwidth]{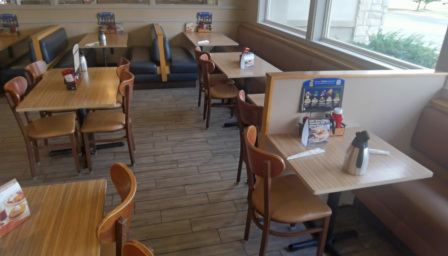}};
    	\end{tikzpicture}
    \end{subfigure}
    \begin{subfigure}{\imgWidth}
        \begin{tikzpicture}[spy using outlines={green,magnification=\ssmag,size=\ssizz},inner sep=0]
            \node [align=center, img] {\includegraphics[width=\textwidth]{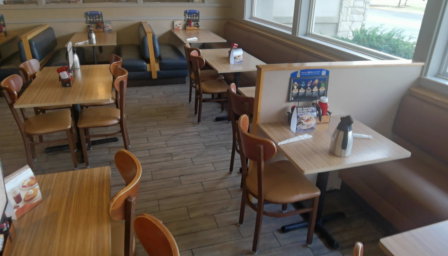}};
            \spy on \reone in node [left] at \rebigone;
    	\end{tikzpicture}
    \end{subfigure}
    \begin{subfigure}{\imgWidth}
		\begin{tikzpicture}[spy using outlines={green,magnification=\ssmag,size=\ssizz},inner sep=0]
            \node [align=center, img] {\includegraphics[width=\textwidth]{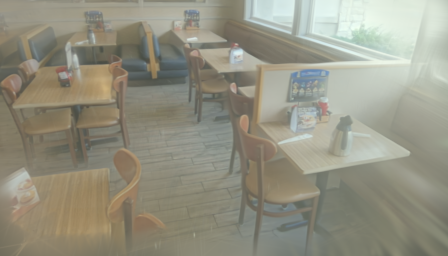}};
            \spy on \reone in node [left] at \rebigone;
    	\end{tikzpicture}
    \end{subfigure}
        \begin{subfigure}{\imgWidth}
        \begin{tikzpicture}[spy using outlines={green,magnification=\ssmag,size=\ssizz},inner sep=0]
            \node [align=center, img] {\includegraphics[width=\textwidth]{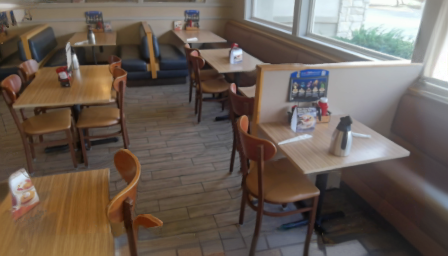}};
            \spy on \reone in node [left] at \rebigone;
    	\end{tikzpicture}
    \end{subfigure}
      \\ %
    \begin{subfigure}{0.1355\textwidth}
        \begin{tikzpicture}[spy using outlines={green,magnification=\ssmag,size=\ssizz},inner sep=0]
            \node [align=center, img] {\includegraphics[width=\textwidth]{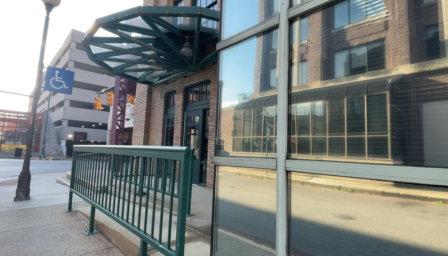}};
    	\end{tikzpicture}
        \\
        \begin{tikzpicture}[spy using outlines={green,magnification=\ssmag,size=\ssizz},inner sep=0]
            \node [align=center, img] {\includegraphics[width=\textwidth]{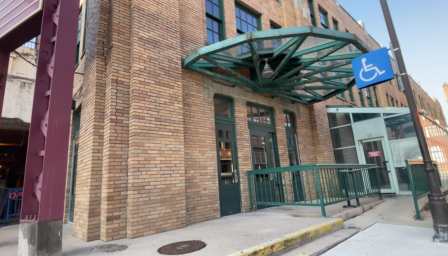}};
    	\end{tikzpicture}
     \caption*{Input Views}
    \end{subfigure}
    \begin{subfigure}{\imgWidth}
        \begin{tikzpicture}[spy using outlines={green,magnification=\ssmag,size=\ssizz},inner sep=0]
            \node [align=center, img] {\includegraphics[width=\textwidth]{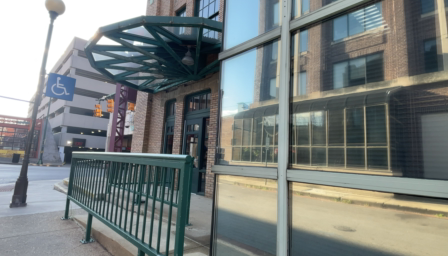}};
            \spy on \retwo in node [left] at \rebigone;
    	\end{tikzpicture}
     \caption*{Target View}
    \end{subfigure}
    \begin{subfigure}{\imgWidth}
		\begin{tikzpicture}[spy using outlines={green,magnification=\ssmag,size=\ssizz},inner sep=0]
            \node [align=center, img] {\includegraphics[width=\textwidth]{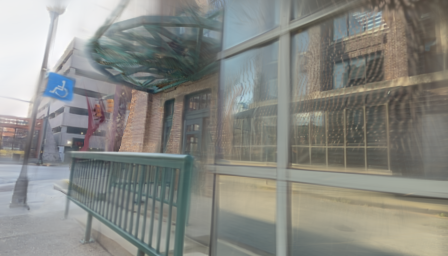}};
            \spy on \retwo in node [left] at \rebigone;
    	\end{tikzpicture}
     \caption*{DepthSplat}
    \end{subfigure}
        \begin{subfigure}{\imgWidth}
        \begin{tikzpicture}[spy using outlines={green,magnification=\ssmag,size=\ssizz},inner sep=0]
            \node [align=center, img] {\includegraphics[width=\textwidth]{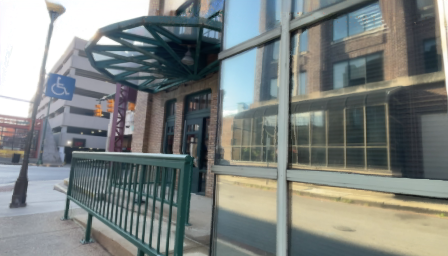}};
            \spy on \retwo in node [left] at \rebigone;
    	\end{tikzpicture}
      \caption*{\textbf{Ours}}
    \end{subfigure}
	\caption{Visual comparisons on the DL3DV-10K dataset}
 \label{fig:main-sparseview-dl3dv}
 \end{subfigure}
\begin{subfigure}{0.355\linewidth}
    \begin{subfigure}{0.1355\textwidth}
        \begin{tikzpicture}[spy using outlines={green,magnification=\ssmag,size=\ssizz},inner sep=0]
            \node [align=center, img] {\includegraphics[width=\textwidth]{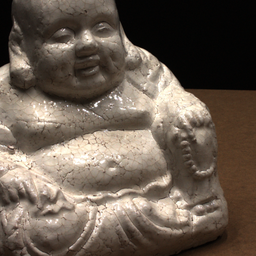}};
    	\end{tikzpicture}
        \\
        \begin{tikzpicture}[spy using outlines={green,magnification=\ssmag,size=\ssizz},inner sep=0]
            \node [align=center, img] {\includegraphics[width=\textwidth]{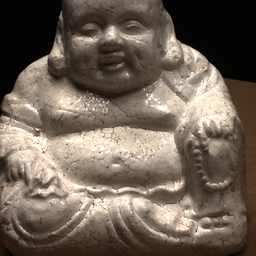}};
    	\end{tikzpicture}
    \end{subfigure}
    \begin{subfigure}{\imgWidth}
        \begin{tikzpicture}[spy using outlines={green,magnification=\ssmag,size=\ssizz},inner sep=0]
            \node [align=center, img] {\includegraphics[width=\textwidth]{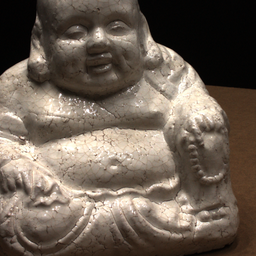}};
            \spy on \dlone in node [left] at \dlbigone;
    	\end{tikzpicture}
    \end{subfigure}
    \begin{subfigure}{\imgWidth}
		\begin{tikzpicture}[spy using outlines={green,magnification=\ssmag,size=\ssizz},inner sep=0]
            \node [align=center, img] {\includegraphics[width=\textwidth]{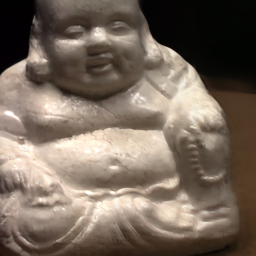}};
            \spy on \dlone in node [left] at \dlbigone;
    	\end{tikzpicture}
    \end{subfigure}
        \begin{subfigure}{\imgWidth}
        \begin{tikzpicture}[spy using outlines={green,magnification=\ssmag,size=\ssizz},inner sep=0]
            \node [align=center, img] {\includegraphics[width=\textwidth]{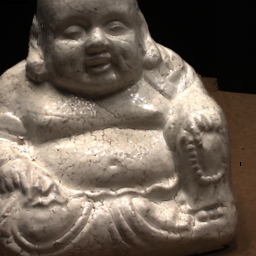}};
            \spy on \dlone in node [left] at \dlbigone;
    	\end{tikzpicture}
    \end{subfigure}
      \\ %
    \begin{subfigure}{0.1355\textwidth}
        \begin{tikzpicture}[spy using outlines={green,magnification=\ssmag,size=\ssizz},inner sep=0]
            \node [align=center, img] {\includegraphics[width=\textwidth]{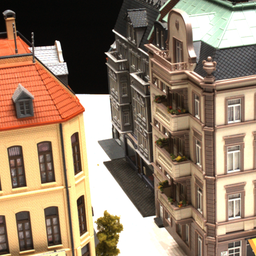}};
    	\end{tikzpicture}
        \\
        \begin{tikzpicture}[spy using outlines={green,magnification=\ssmag,size=\ssizz},inner sep=0]
            \node [align=center, img] {\includegraphics[width=\textwidth]{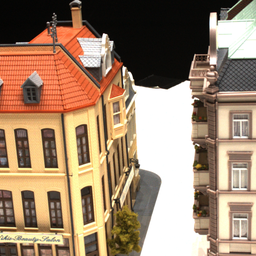}};
    	\end{tikzpicture}
     \caption*{Inputs}
    \end{subfigure}
    \begin{subfigure}{\imgWidth}
        \begin{tikzpicture}[spy using outlines={green,magnification=\ssmag,size=\ssizz},inner sep=0]
            \node [align=center, img] {\includegraphics[width=\textwidth]{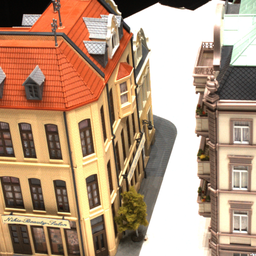}};
            \spy on \dltwo in node [left] at \dlbigone;
    	\end{tikzpicture}
     \caption*{Target View}
    \end{subfigure}
    \begin{subfigure}{\imgWidth}
		\begin{tikzpicture}[spy using outlines={green,magnification=\ssmag,size=\ssizz},inner sep=0]
            \node [align=center, img] {\includegraphics[width=\textwidth]{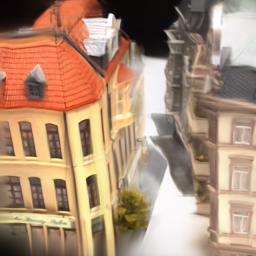}};
            \spy on \dltwo in node [left] at \dlbigone;
    	\end{tikzpicture}
     \caption*{DepthSplat}
    \end{subfigure}
        \begin{subfigure}{\imgWidth}
        \begin{tikzpicture}[spy using outlines={green,magnification=\ssmag,size=\ssizz},inner sep=0]
            \node [align=center, img] {\includegraphics[width=\textwidth]{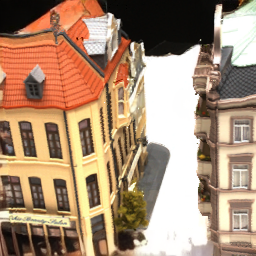}};
            \spy on \dltwo in node [left] at \dlbigone;
    	\end{tikzpicture}
      \caption*{\textbf{Ours}}
    \end{subfigure}
	\caption{Visual comparisons on the DTU dataset}
 \label{fig:main-sparseview-dtu}
 \end{subfigure}
 \caption{\textbf{Sparse-view novel view synthesis results on the DL3DV-10K (in-domain) and DTU (cross-domain) datasets.}}
 \label{fig:main-sparseview}
\end{figure*}

%% file: figs/main/fig-stereconv.tex
\def\imgWidth{0.238\textwidth} %
\def\scc{(-1.9,-1.4)}
\def\rebigone{(-1.1, 0.9)} %
\def\rebigtwo{(-1.1, -0.9)} %
\def\reone{(0.6,-0.05)} %
\def\retwo{(1.3,1.1)} %
\def\rethree{(-1.25,0.7)} %
\def\refour{(1.15,0.6)} %

\def\ssizz{1cm} %
\def\ssmag{3}

\begin{figure*}[!t] 
\centering
\tikzstyle{img} = [rectangle, minimum width=\imgWidth, draw=black]
\begin{subfigure}{\textwidth}
\centering
    \begin{subfigure}{\imgWidth}
        \begin{tikzpicture}[spy using outlines={green,magnification=\ssmag,size=\ssizz},inner sep=0]
            \node [align=center, img] {\includegraphics[width=\textwidth]{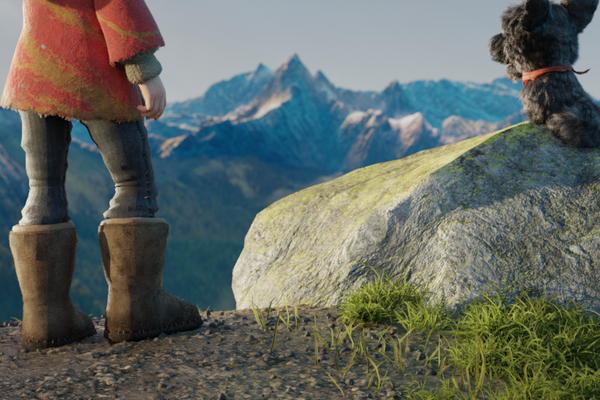}};
    	\end{tikzpicture}
    \end{subfigure}
    \begin{subfigure}{\imgWidth}
        \begin{tikzpicture}[spy using outlines={green,magnification=\ssmag,size=\ssizz},inner sep=0]
            \node [align=center, img] {\includegraphics[width=\textwidth]{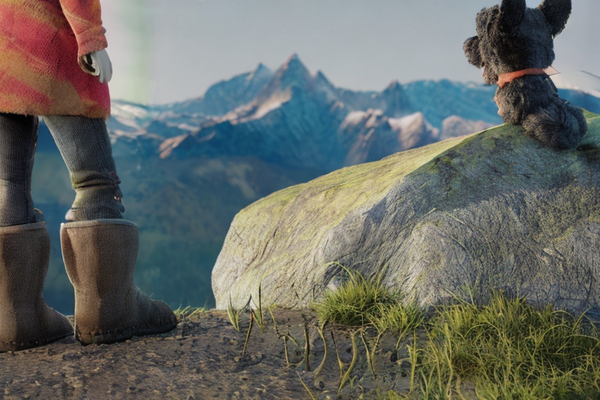}};
            \spy on \rethree in node [left] at \rebigtwo;
    	\end{tikzpicture}
    \end{subfigure}
    \begin{subfigure}{\imgWidth}
		\begin{tikzpicture}[spy using outlines={green,magnification=\ssmag,size=\ssizz},inner sep=0]
            \node [align=center, img] {\includegraphics[width=\textwidth]{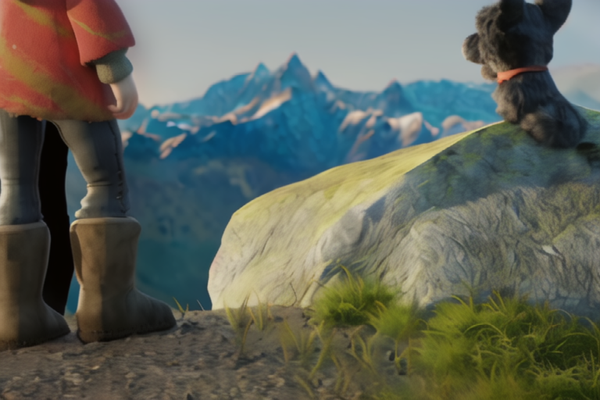}};
            \spy on \rethree in node [left] at \rebigtwo;
    	\end{tikzpicture}
    \end{subfigure}
        \begin{subfigure}{\imgWidth}
        \begin{tikzpicture}[spy using outlines={green,magnification=\ssmag,size=\ssizz},inner sep=0]
            \node [align=center, img] {\includegraphics[width=\textwidth]{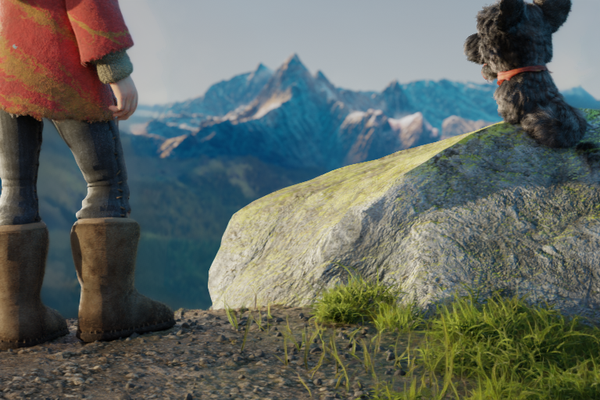}};
            \spy on \rethree in node [left] at \rebigtwo;
    	\end{tikzpicture}
    \end{subfigure}
      \\ %
          \begin{subfigure}{\imgWidth}
        \begin{tikzpicture}[spy using outlines={green,magnification=\ssmag,size=\ssizz},inner sep=0]
            \node [align=center, img] {\includegraphics[width=\textwidth]{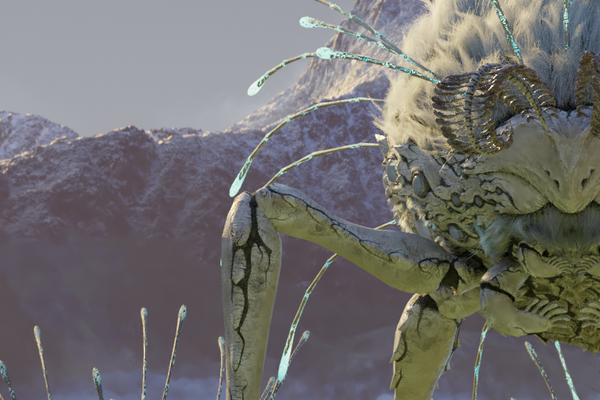}};
    	\end{tikzpicture}
     \caption*{Input View (Left)}
    \end{subfigure}
    \begin{subfigure}{\imgWidth}
        \begin{tikzpicture}[spy using outlines={green,magnification=\ssmag,size=\ssizz},inner sep=0]
            \node [align=center, img] {\includegraphics[width=\textwidth]{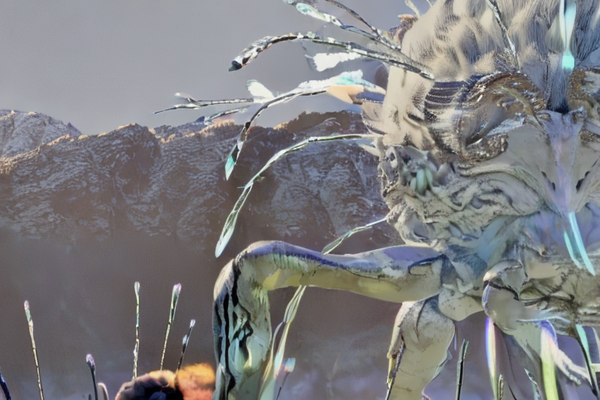}};
            \spy on \refour in node [left] at \rebigtwo;
    	\end{tikzpicture}
     \caption*{ViewCrafter}
    \end{subfigure}
    \begin{subfigure}{\imgWidth}
		\begin{tikzpicture}[spy using outlines={green,magnification=\ssmag,size=\ssizz},inner sep=0]
            \node [align=center, img] {\includegraphics[width=\textwidth]{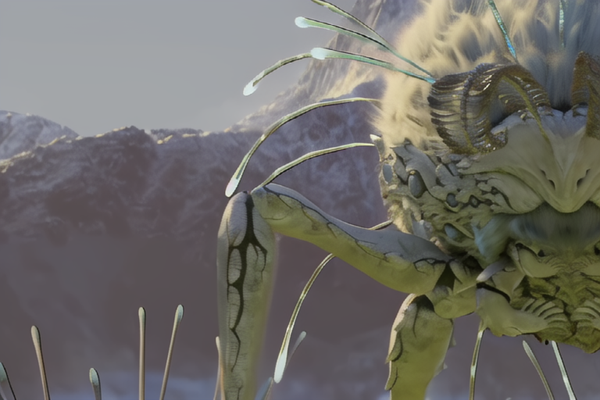}};
            \spy on \refour in node [left] at \rebigtwo;
    	\end{tikzpicture}
     \caption*{StereoCrafter}
    \end{subfigure}
    \begin{subfigure}{\imgWidth}
        \begin{tikzpicture}[spy using outlines={green,magnification=\ssmag,size=\ssizz},inner sep=0]
            \node [align=center, img] {\includegraphics[width=\textwidth]{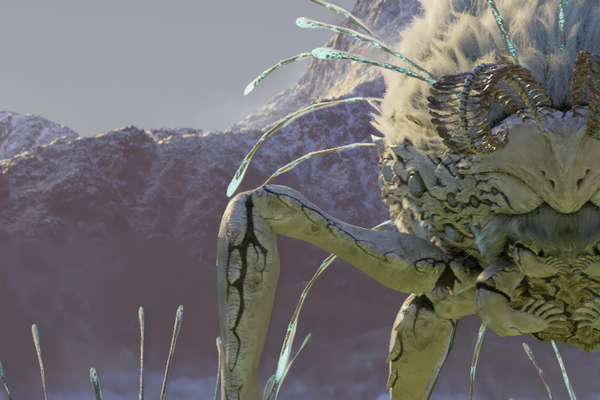}};
            \spy on \refour in node [left] at \rebigtwo;
    	\end{tikzpicture}
      \caption*{\textbf{\method\ (Ours)}}
    \end{subfigure}
\end{subfigure}
 \caption{\textbf{Stereo video conversion results on the Spring dataset.} Right-eye views are synthesized based on the input left-eye views.}
 \label{fig:main-stereoconv}
\end{figure*}

%% file: figs/supp/fig-texture_degradation.tex
\def\imgWidth{0.238\textwidth} %
\def\scc{(-1.9,-1.4)}
\def\rebigone{(-1.28, -0.75)} %
\def\rebigtwo{(-1.28, 0.75)} %
\def\rethree{(2.03,-0.15)} %
\def\refour{(-0.15,-0.7)} 

\def\ssizz{0.8cm} %
\def\ssmag{4}

\begin{figure*}
\centering
\tikzstyle{img} = [rectangle, minimum width=\imgWidth, draw=black]
\centering
\begin{subfigure}{\textwidth}
\centering
          \begin{subfigure}{\imgWidth}
        \begin{tikzpicture}[spy using outlines={green,magnification=\ssmag,size=\ssizz},inner sep=0]
            \node [align=center, img] {\includegraphics[width=\textwidth]{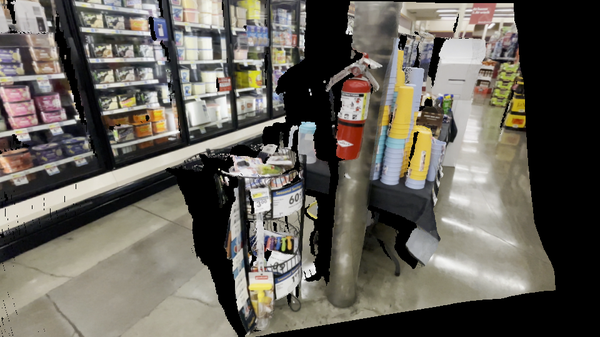}};
            \spy [red] on \rethree in node [left] at \rebigtwo;
            \spy on \refour in node [left] at \rebigone;
    	\end{tikzpicture}
     \caption*{Splatted View}
    \end{subfigure}
    \begin{subfigure}{\imgWidth}
        \begin{tikzpicture}[spy using outlines={green,magnification=\ssmag,size=\ssizz},inner sep=0]
            \node [align=center, img] {\includegraphics[width=\textwidth]{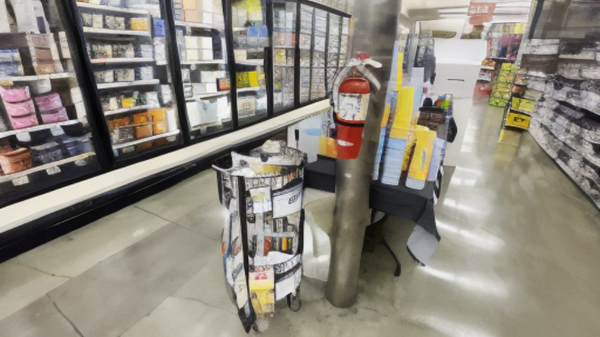}};
            \spy [red] on \rethree in node [left] at \rebigtwo;
            \spy on \refour in node [left] at \rebigone;
    	\end{tikzpicture}
     \caption*{Diffusion Output}
    \end{subfigure}
    \begin{subfigure}{\imgWidth}
		\begin{tikzpicture}[spy using outlines={green,magnification=\ssmag,size=\ssizz},inner sep=0]
            \node [align=center, img] {\includegraphics[width=\textwidth]{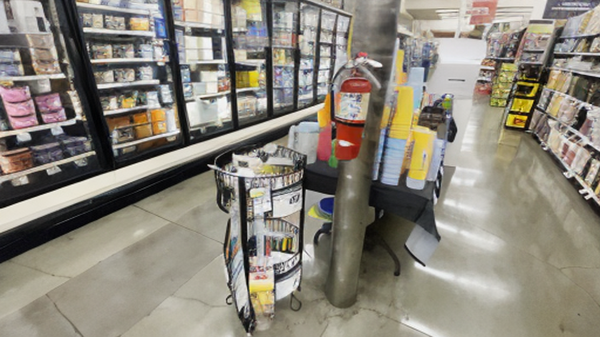}};
            \spy [red] on \rethree in node [left] at \rebigtwo;
            \spy on \refour in node [left] at \rebigone;
    	\end{tikzpicture}
     \caption*{Degraded View}
    \end{subfigure}
    \begin{subfigure}{\imgWidth}
        \begin{tikzpicture}[spy using outlines={green,magnification=\ssmag,size=\ssizz},inner sep=0]
            \node [align=center, img] {\includegraphics[width=\textwidth]{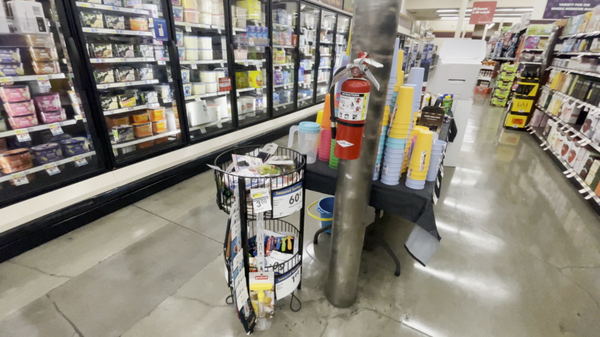}};
            \spy [red] on \rethree in node [left] at \rebigtwo;
            \spy on \refour in node [left] at \rebigone;
    	\end{tikzpicture}
      \caption*{Target View}
    \end{subfigure}
    \end{subfigure}
    \caption{\textbf{Example of texture degradation.} Although the original diffusion output fills the unknown regions in the splatted view, the generated contents usually differ from the target view (red box), resulting in sub-optimal training performance. Thus, we employ the degraded view for training, which shares similar contents with the target view while imitating the texture hallucination effects (green box) in real diffusion outputs. }
 \label{fig:supp-texture-degradation}
\end{figure*}

%% file: figs/supp/fig-ablation-vis.tex
\def\imgWidth{0.238\textwidth} %
\def\scc{(-1.9,-1.4)}
\def\rebigone{(2.1, -0.68)} %

\def\reone{(0.35,-1)} %
\def\retwo{(-0.1,0.05)} %

\def\ssizz{1cm} %
\def\ssmag{3}

\begin{figure*}[!t] 
\centering
\tikzstyle{img} = [rectangle, minimum width=\imgWidth, draw=black]
\centering
\begin{subfigure}{\textwidth}
    \centering
          \begin{subfigure}{\imgWidth}
        \begin{tikzpicture}[spy using outlines={green,magnification=\ssmag,size=\ssizz},inner sep=0]
            \node [align=center, img] {\includegraphics[width=\textwidth]{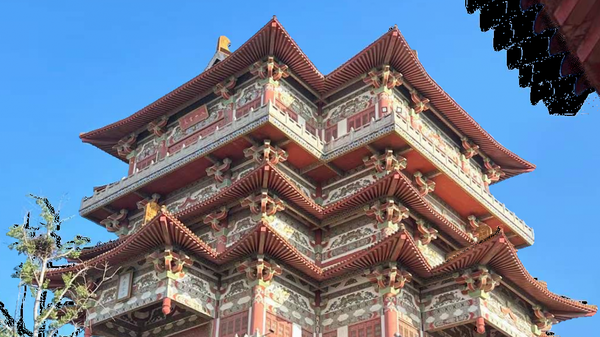}};
            \spy on \reone in node [left] at \rebigone;
    	\end{tikzpicture}
     \caption*{Splatted View}
    \end{subfigure}
    \begin{subfigure}{\imgWidth}
        \begin{tikzpicture}[spy using outlines={green,magnification=\ssmag,size=\ssizz},inner sep=0]
            \node [align=center, img] {\includegraphics[width=\textwidth]{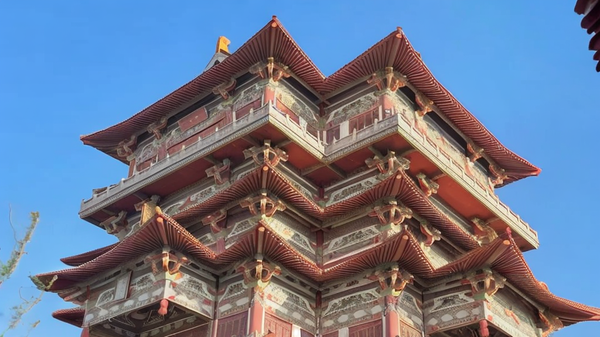}};
            \spy on \reone in node [left] at \rebigone;
    	\end{tikzpicture}
     \caption*{Baseline (\#1)}
    \end{subfigure}
    \begin{subfigure}{\imgWidth}
		\begin{tikzpicture}[spy using outlines={green,magnification=\ssmag,size=\ssizz},inner sep=0]
            \node [align=center, img] {\includegraphics[width=\textwidth]{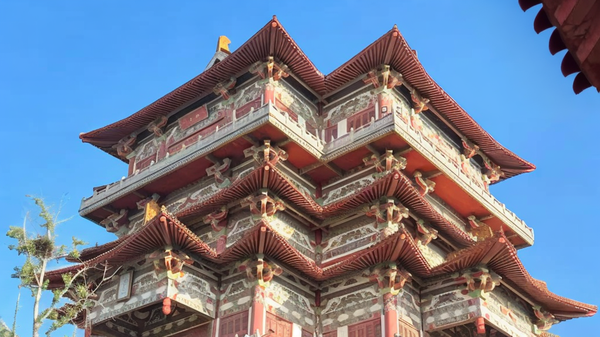}};
            \spy on \reone in node [left] at \rebigone;
    	\end{tikzpicture}
     \caption*{w/ Aligned Synthesis (\#3)}
    \end{subfigure}
    \begin{subfigure}{\imgWidth}
        \begin{tikzpicture}[spy using outlines={green,magnification=\ssmag,size=\ssizz},inner sep=0]
            \node [align=center, img] {\includegraphics[width=\textwidth]{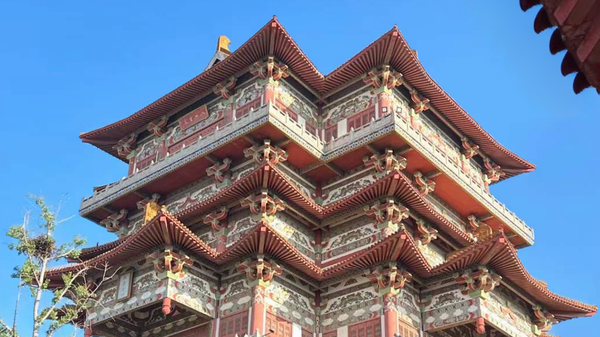}};
            \spy on \reone in node [left] at \rebigone;
    	\end{tikzpicture}
      \caption*{w/ All (\#5)}
\end{subfigure}
\begin{subfigure}{\textwidth}
    \centering
    \begin{subfigure}{\imgWidth}
        \begin{tikzpicture}[spy using outlines={green,magnification=\ssmag,size=\ssizz},inner sep=0]
            \node [align=center, img] {\includegraphics[width=\textwidth]{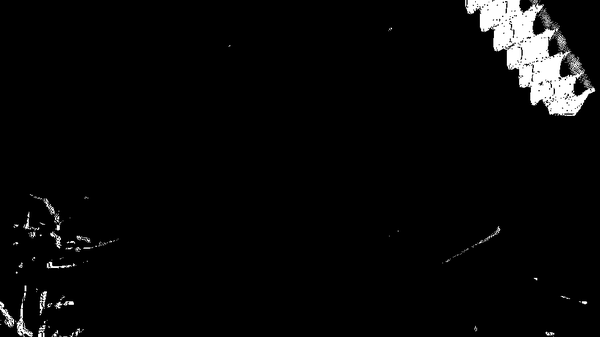}};
    	\end{tikzpicture}
     \caption*{Unkown Region}
    \end{subfigure}
    \begin{subfigure}{\imgWidth}
		\begin{tikzpicture}[spy using outlines={green,magnification=\ssmag,size=\ssizz},inner sep=0]
            \node [align=center, img] {\includegraphics[width=\textwidth]{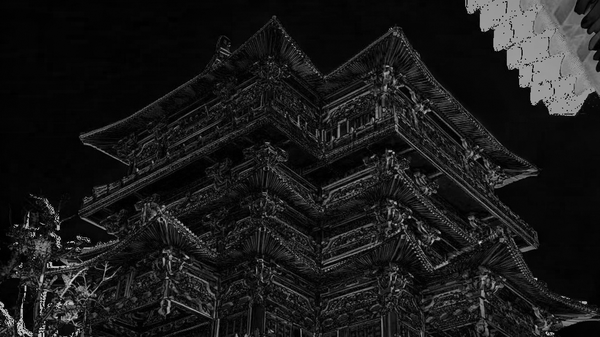}};
    	\end{tikzpicture}
     \caption*{Diff. Map (Baseline)}
    \end{subfigure}
    \begin{subfigure}{\imgWidth}
        \begin{tikzpicture}[spy using outlines={green,magnification=\ssmag,size=\ssizz},inner sep=0]
            \node [align=center, img] {\includegraphics[width=\textwidth]{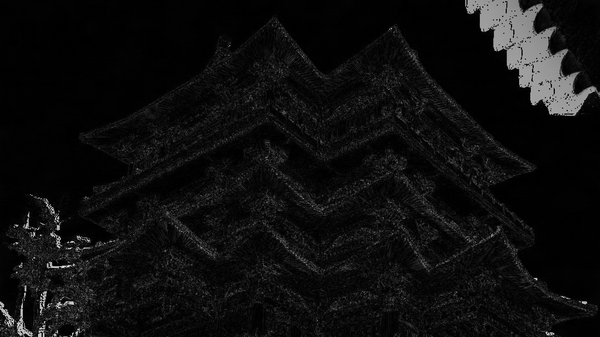}};
    	\end{tikzpicture}
     \caption*{Diff. Map (w/ Aligned Synthesis)}
    \end{subfigure}
    \begin{subfigure}{\imgWidth}
        \begin{tikzpicture}[spy using outlines={green,magnification=\ssmag,size=\ssizz},inner sep=0]
            \node [align=center, img] {\includegraphics[width=\textwidth]{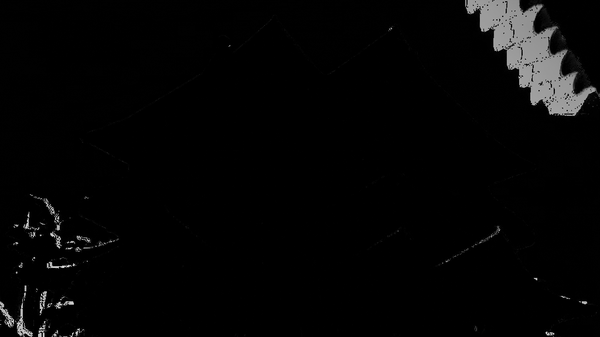}};
    	\end{tikzpicture}
      \caption*{Diff. Map (w/ All)}
    \end{subfigure}
\end{subfigure}
\end{subfigure}
    \caption{\textbf{Visual comparisons of ablation study.} The difference map shows the absolute difference between the splatted view and the corresponding novel view. The model ID is consistent with Tab. 2 in the main paper. The baseline model generates a misaligned novel view with the conditioned splatted view, showing significant differences across the image. With the proposed aligned synthesis strategy, model \#3 better follows the conditioning but still suffers from texture hallucination (green box). Combining the aligned synthesis and the texture bridge, our model (\#5) synthesizes geometry-aligned novel views while recovering high-fidelity texture. }
 \label{fig:supp-ablation-vis}
\end{figure*}

%% file: figs/supp/fig-singleview-wild.tex
\def\imgWidth{0.238\textwidth} %
\def\scc{(-1.9,-1.4)}
\def\rebigone{(2.1, -0.65)} %

\def\reone{(1.1,0.5)} %
\def\reseight{(-1.05,-0.3)} %
\def\retwo{(-0.1,0.05)} %
\def\rethree{(-0.35,0.35)} %
\def\refour{(0,0.73)} %
\def\refive{(0.8,0.5)} %
\def\resix{(1.4,0.63)} %
\def\reseven{(-1.4,-0.75)} %

\def\ssizz{1cm} %
\def\ssmag{3}

\begin{figure*}[!t] 
\centering
\tikzstyle{img} = [rectangle, minimum width=\imgWidth, draw=black]
\centering
\begin{subfigure}{\textwidth}
\centering
    \begin{subfigure}{\imgWidth}
        \begin{tikzpicture}[spy using outlines={green,magnification=\ssmag,size=\ssizz},inner sep=0]
            \node [align=center, img] {\includegraphics[width=\textwidth]{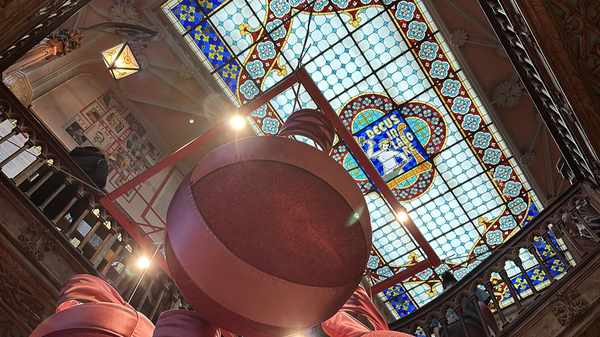}};
    	\end{tikzpicture}
    \end{subfigure}
    \begin{subfigure}{\imgWidth}
        \begin{tikzpicture}[spy using outlines={green,magnification=\ssmag,size=\ssizz},inner sep=0]
            \node [align=center, img] {\includegraphics[width=\textwidth]{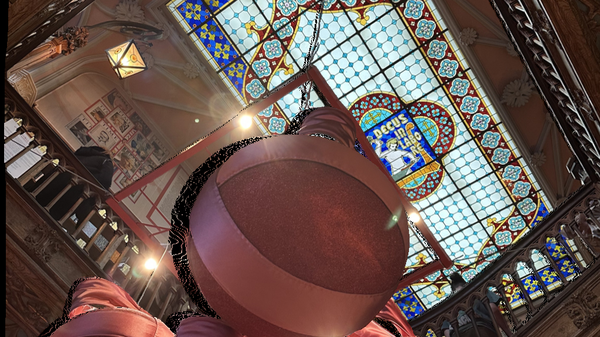}};
            \spy on \reone in node [left] at \rebigone;
    	\end{tikzpicture}
    \end{subfigure}
    \begin{subfigure}{\imgWidth}
		\begin{tikzpicture}[spy using outlines={green,magnification=\ssmag,size=\ssizz},inner sep=0]
            \node [align=center, img] {\includegraphics[width=\textwidth]{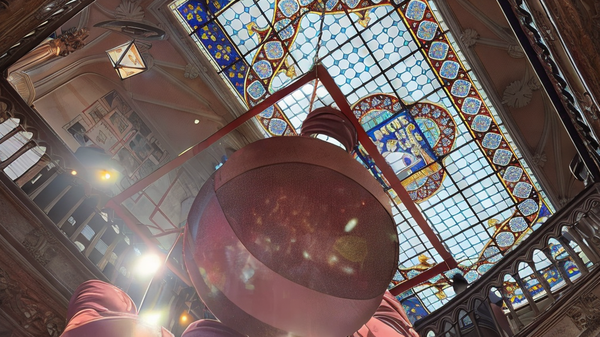}};
            \spy on \reone in node [left] at \rebigone;
    	\end{tikzpicture}
    \end{subfigure}
        \begin{subfigure}{\imgWidth}
        \begin{tikzpicture}[spy using outlines={green,magnification=\ssmag,size=\ssizz},inner sep=0]
            \node [align=center, img] {\includegraphics[width=\textwidth]{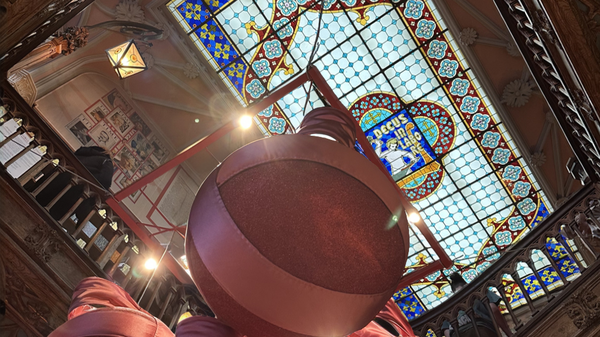}};
            \spy on \reone in node [left] at \rebigone;
    	\end{tikzpicture}
    \end{subfigure}
    \begin{subfigure}{\imgWidth}
        \begin{tikzpicture}[spy using outlines={green,magnification=\ssmag,size=\ssizz},inner sep=0]
            \node [align=center, img] {\includegraphics[width=\textwidth]{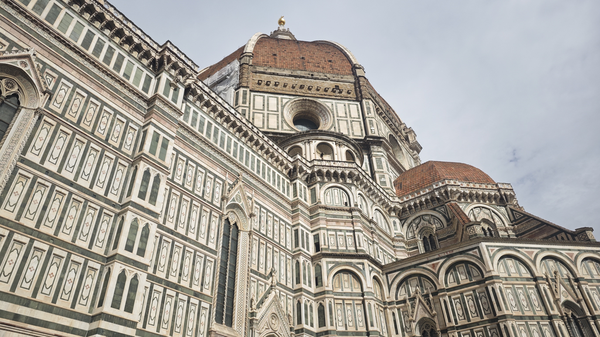}};
    	\end{tikzpicture}
    \end{subfigure}
    \begin{subfigure}{\imgWidth}
        \begin{tikzpicture}[spy using outlines={green,magnification=\ssmag,size=\ssizz},inner sep=0]
            \node [align=center, img] {\includegraphics[width=\textwidth]{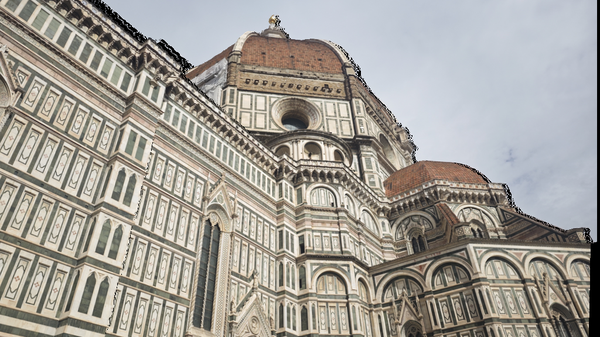}};
            \spy on \reseight in node [left] at \rebigone;
    	\end{tikzpicture}
    \end{subfigure}
    \begin{subfigure}{\imgWidth}
		\begin{tikzpicture}[spy using outlines={green,magnification=\ssmag,size=\ssizz},inner sep=0]
            \node [align=center, img] {\includegraphics[width=\textwidth]{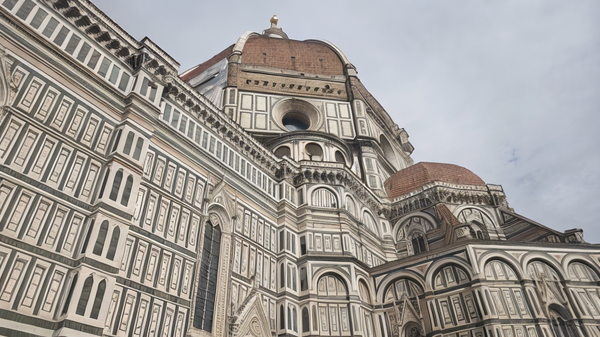}};
            \spy on \reseight in node [left] at \rebigone;
    	\end{tikzpicture}
    \end{subfigure}
        \begin{subfigure}{\imgWidth}
        \begin{tikzpicture}[spy using outlines={green,magnification=\ssmag,size=\ssizz},inner sep=0]
            \node [align=center, img] {\includegraphics[width=\textwidth]{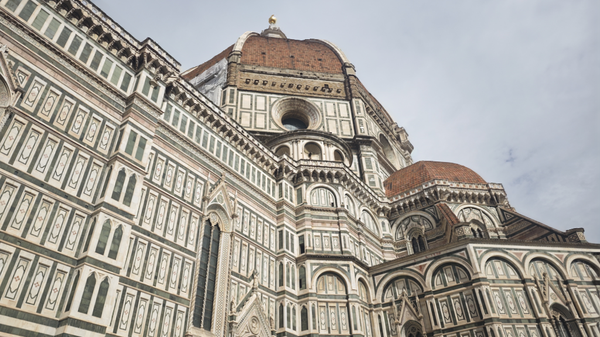}};
            \spy on \reseight in node [left] at \rebigone;
    	\end{tikzpicture}
    \end{subfigure}
        \\ %
    \begin{subfigure}{\imgWidth}
        \begin{tikzpicture}[spy using outlines={green,magnification=\ssmag,size=\ssizz},inner sep=0]
            \node [align=center, img] {\includegraphics[width=\textwidth]{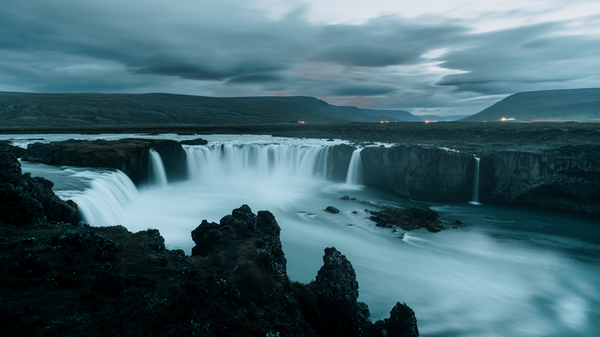}};
    	\end{tikzpicture}
    \end{subfigure}
    \begin{subfigure}{\imgWidth}
        \begin{tikzpicture}[spy using outlines={green,magnification=\ssmag,size=\ssizz},inner sep=0]
            \node [align=center, img] {\includegraphics[width=\textwidth]{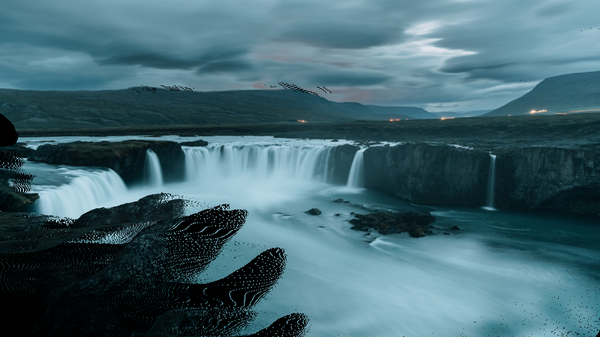}};
            \spy on \retwo in node [left] at \rebigone;
    	\end{tikzpicture}
    \end{subfigure}
    \begin{subfigure}{\imgWidth}
		\begin{tikzpicture}[spy using outlines={green,magnification=\ssmag,size=\ssizz},inner sep=0]
            \node [align=center, img] {\includegraphics[width=\textwidth]{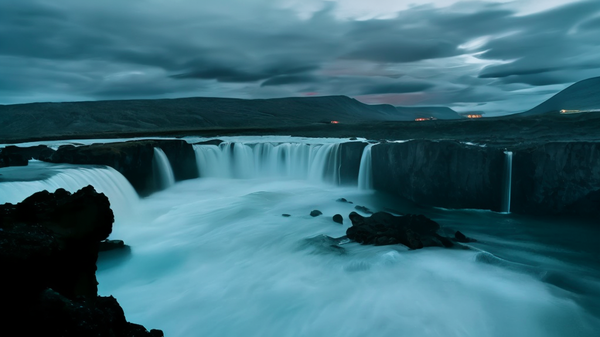}};
            \spy on \retwo in node [left] at \rebigone;
    	\end{tikzpicture}
    \end{subfigure}
        \begin{subfigure}{\imgWidth}
        \begin{tikzpicture}[spy using outlines={green,magnification=\ssmag,size=\ssizz},inner sep=0]
            \node [align=center, img] {\includegraphics[width=\textwidth]{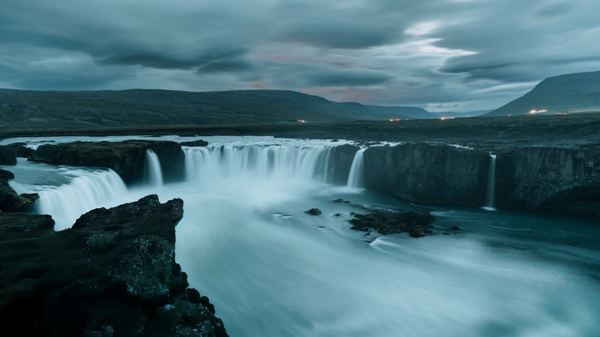}};
            \spy on \retwo in node [left] at \rebigone;
    	\end{tikzpicture}
    \end{subfigure}
      \\ %
    \begin{subfigure}{\imgWidth}
        \begin{tikzpicture}[spy using outlines={green,magnification=\ssmag,size=\ssizz},inner sep=0]
            \node [align=center, img] {\includegraphics[width=\textwidth]{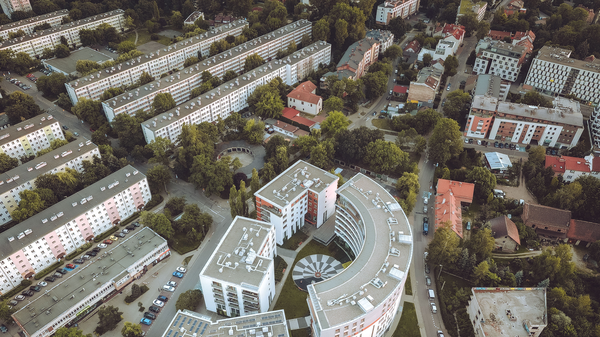}};
    	\end{tikzpicture}
    \end{subfigure}
    \begin{subfigure}{\imgWidth}
        \begin{tikzpicture}[spy using outlines={green,magnification=\ssmag,size=\ssizz},inner sep=0]
            \node [align=center, img] {\includegraphics[width=\textwidth]{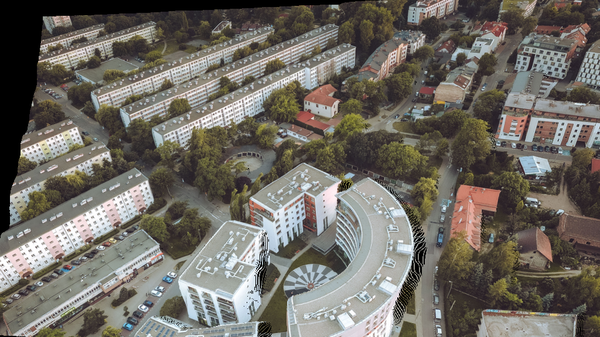}};
            \spy on \rethree in node [left] at \rebigone;
    	\end{tikzpicture}
    \end{subfigure}
    \begin{subfigure}{\imgWidth}
		\begin{tikzpicture}[spy using outlines={green,magnification=\ssmag,size=\ssizz},inner sep=0]
            \node [align=center, img] {\includegraphics[width=\textwidth]{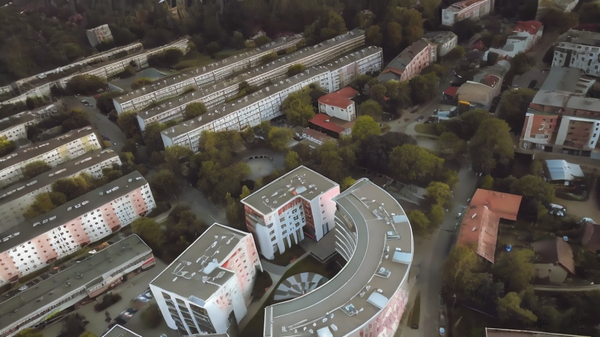}};
            \spy on \rethree in node [left] at \rebigone;
    	\end{tikzpicture}
    \end{subfigure}
        \begin{subfigure}{\imgWidth}
        \begin{tikzpicture}[spy using outlines={green,magnification=\ssmag,size=\ssizz},inner sep=0]
            \node [align=center, img] {\includegraphics[width=\textwidth]{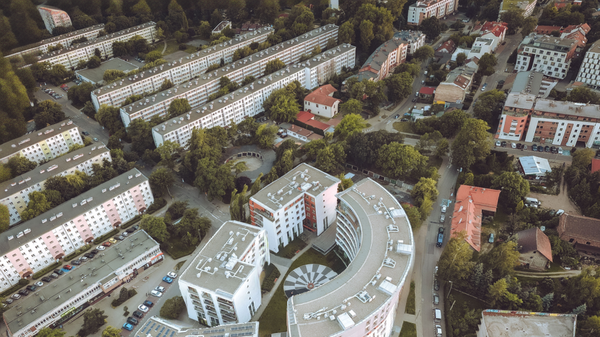}};
            \spy on \rethree in node [left] at \rebigone;
    	\end{tikzpicture}
    \end{subfigure}
      \\ %
          \begin{subfigure}{\imgWidth}
        \begin{tikzpicture}[spy using outlines={green,magnification=\ssmag,size=\ssizz},inner sep=0]
            \node [align=center, img] {\includegraphics[width=\textwidth]{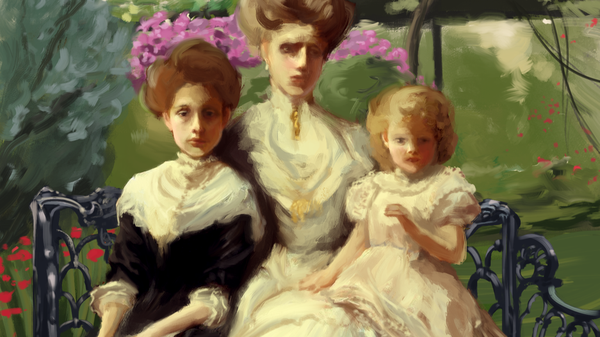}};
    	\end{tikzpicture}
    \end{subfigure}
    \begin{subfigure}{\imgWidth}
        \begin{tikzpicture}[spy using outlines={green,magnification=\ssmag,size=\ssizz},inner sep=0]
            \node [align=center, img] {\includegraphics[width=\textwidth]{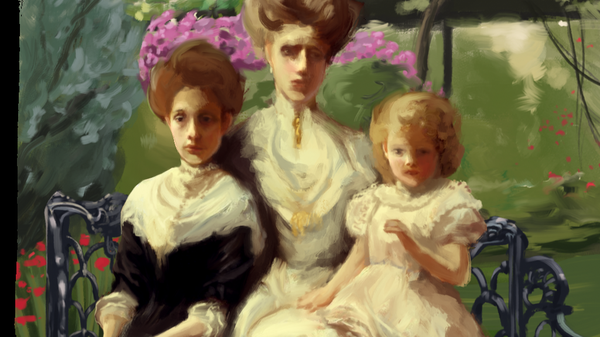}};
            \spy on \refour in node [left] at \rebigone;
    	\end{tikzpicture}
    \end{subfigure}
    \begin{subfigure}{\imgWidth}
		\begin{tikzpicture}[spy using outlines={green,magnification=\ssmag,size=\ssizz},inner sep=0]
            \node [align=center, img] {\includegraphics[width=\textwidth]{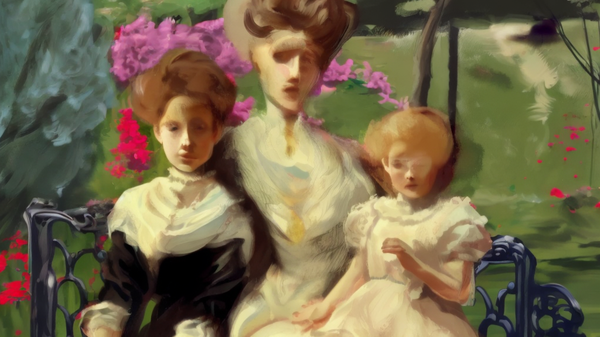}};
            \spy on \refour in node [left] at \rebigone;
    	\end{tikzpicture}
    \end{subfigure}
        \begin{subfigure}{\imgWidth}
        \begin{tikzpicture}[spy using outlines={green,magnification=\ssmag,size=\ssizz},inner sep=0]
            \node [align=center, img] {\includegraphics[width=\textwidth]{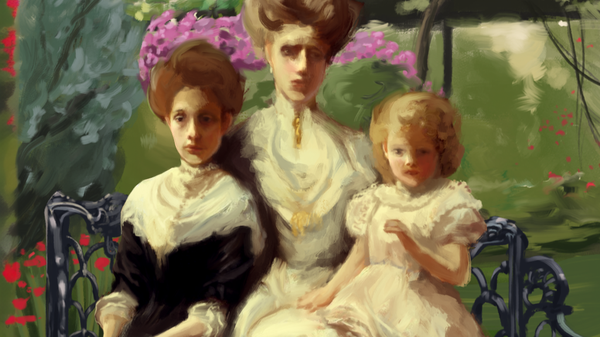}};
            \spy on \refour in node [left] at \rebigone;
    	\end{tikzpicture}
    \end{subfigure}
      \\ %
      \begin{subfigure}{\imgWidth}
        \begin{tikzpicture}[spy using outlines={green,magnification=\ssmag,size=\ssizz},inner sep=0]
            \node [align=center, img] {\includegraphics[width=\textwidth]{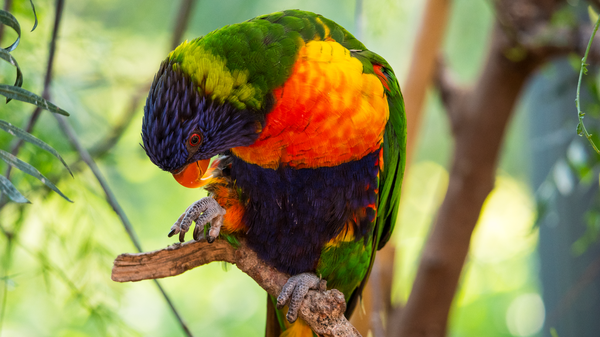}};
    	\end{tikzpicture}
    \end{subfigure}
    \begin{subfigure}{\imgWidth}
        \begin{tikzpicture}[spy using outlines={green,magnification=\ssmag,size=\ssizz},inner sep=0]
            \node [align=center, img] {\includegraphics[width=\textwidth]{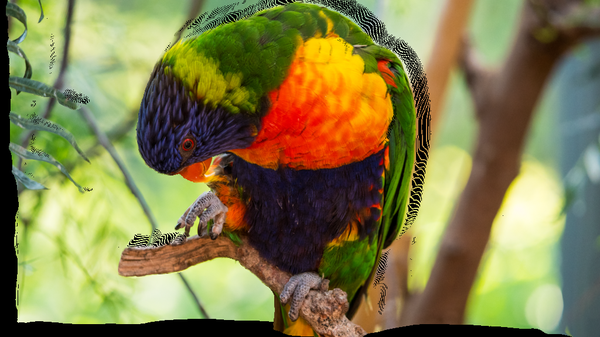}};
            \spy on \refive in node [left] at \rebigone;
    	\end{tikzpicture}
    \end{subfigure}
    \begin{subfigure}{\imgWidth}
		\begin{tikzpicture}[spy using outlines={green,magnification=\ssmag,size=\ssizz},inner sep=0]
            \node [align=center, img] {\includegraphics[width=\textwidth]{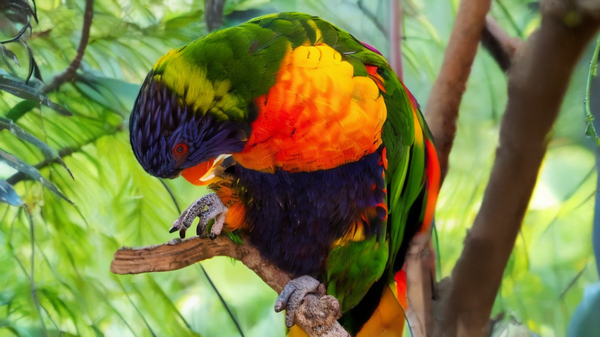}};
            \spy on \refive in node [left] at \rebigone;
    	\end{tikzpicture}
    \end{subfigure}
        \begin{subfigure}{\imgWidth}
        \begin{tikzpicture}[spy using outlines={green,magnification=\ssmag,size=\ssizz},inner sep=0]
            \node [align=center, img] {\includegraphics[width=\textwidth]{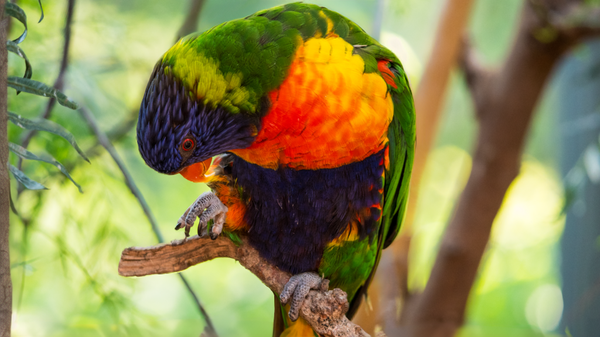}};
            \spy on \refive in node [left] at \rebigone;
    	\end{tikzpicture}
    \end{subfigure}
      \\ %
      \begin{subfigure}{\imgWidth}
        \begin{tikzpicture}[spy using outlines={green,magnification=\ssmag,size=\ssizz},inner sep=0]
            \node [align=center, img] {\includegraphics[width=\textwidth]{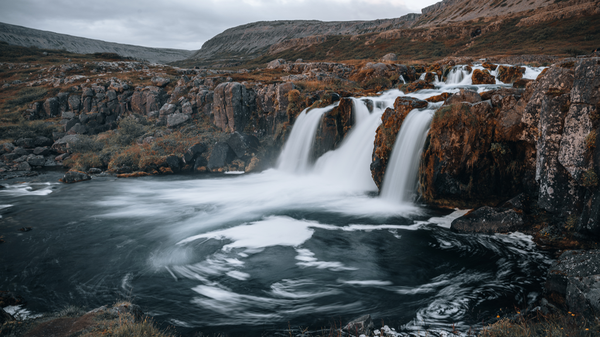}};
    	\end{tikzpicture}
    \end{subfigure}
    \begin{subfigure}{\imgWidth}
        \begin{tikzpicture}[spy using outlines={green,magnification=\ssmag,size=\ssizz},inner sep=0]
            \node [align=center, img] {\includegraphics[width=\textwidth]{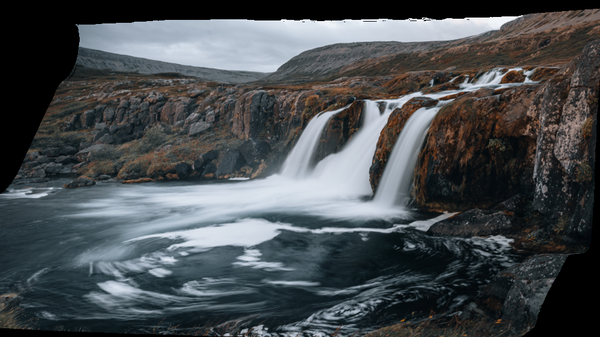}};
            \spy on \resix in node [left] at \rebigone;
    	\end{tikzpicture}
    \end{subfigure}
    \begin{subfigure}{\imgWidth}
		\begin{tikzpicture}[spy using outlines={green,magnification=\ssmag,size=\ssizz},inner sep=0]
            \node [align=center, img] {\includegraphics[width=\textwidth]{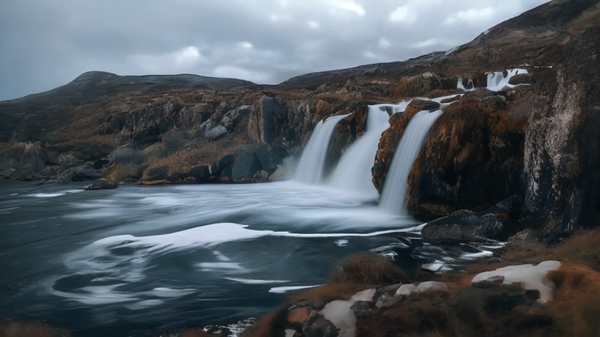}};
            \spy on \resix in node [left] at \rebigone;
    	\end{tikzpicture}
    \end{subfigure}
        \begin{subfigure}{\imgWidth}
        \begin{tikzpicture}[spy using outlines={green,magnification=\ssmag,size=\ssizz},inner sep=0]
            \node [align=center, img] {\includegraphics[width=\textwidth]{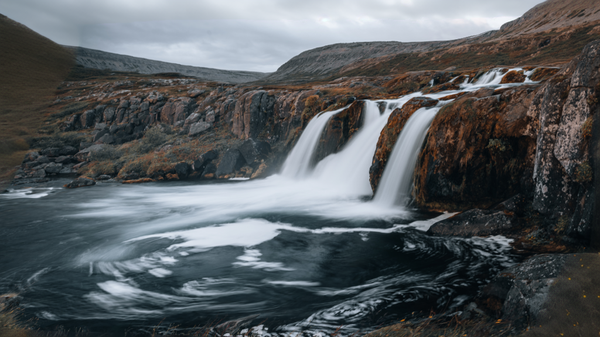}};
            \spy on \resix in node [left] at \rebigone;
    	\end{tikzpicture}
    \end{subfigure}
      \\ %
          \begin{subfigure}{\imgWidth}
        \begin{tikzpicture}[spy using outlines={green,magnification=\ssmag,size=\ssizz},inner sep=0]
            \node [align=center, img] {\includegraphics[width=\textwidth]{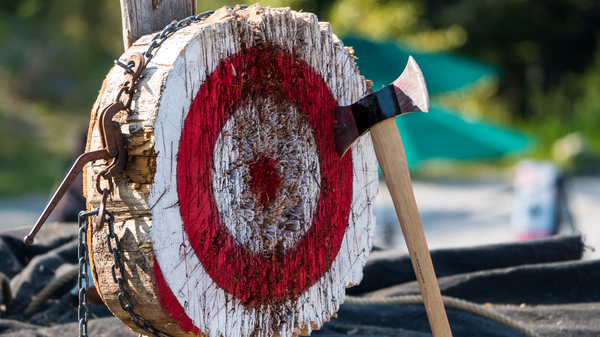}};
    	\end{tikzpicture}
     \caption*{Input View}
    \end{subfigure}
    \begin{subfigure}{\imgWidth}
        \begin{tikzpicture}[spy using outlines={green,magnification=\ssmag,size=\ssizz},inner sep=0]
            \node [align=center, img] {\includegraphics[width=\textwidth]{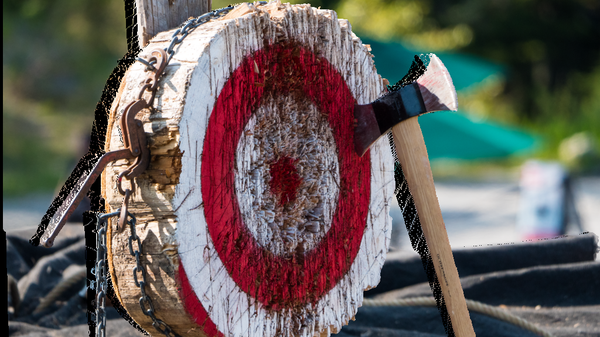}};
            \spy on \reseven in node [left] at \rebigone;
    	\end{tikzpicture}
     \caption*{Splatted View}
    \end{subfigure}
    \begin{subfigure}{\imgWidth}
		\begin{tikzpicture}[spy using outlines={green,magnification=\ssmag,size=\ssizz},inner sep=0]
            \node [align=center, img] {\includegraphics[width=\textwidth]{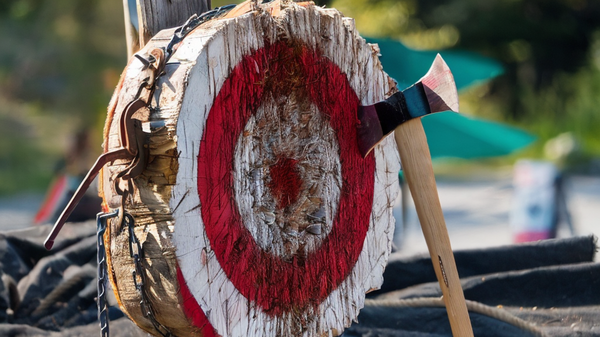}};
            \spy on \reseven in node [left] at \rebigone;
    	\end{tikzpicture}
     \caption*{ViewCrafter}
    \end{subfigure}
    \begin{subfigure}{\imgWidth}
        \begin{tikzpicture}[spy using outlines={green,magnification=\ssmag,size=\ssizz},inner sep=0]
            \node [align=center, img] {\includegraphics[width=\textwidth]{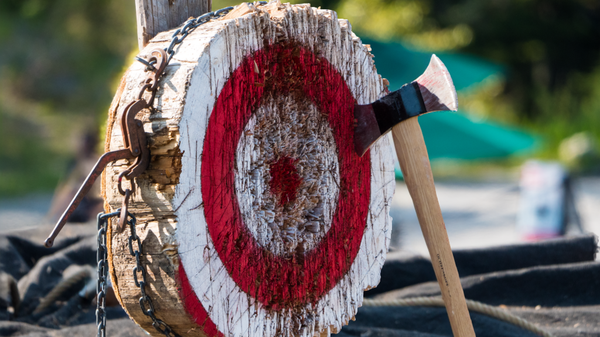}};
            \spy on \reseven in node [left] at \rebigone;
    	\end{tikzpicture}
      \caption*{\textbf{\method\ (Ours)}}
    \end{subfigure}
    \end{subfigure}
    \caption{\textbf{Qualitative comparisons on in-the-wild high-resolution (576$\times$1024) samples.} }
 \label{fig:supp-singleview-wild}
\end{figure*}

%% file: figs/supp/fig-stereoconv.tex
\def\imgWidth{0.238\textwidth} %
\def\scc{(-1.9,-1.4)}
\def\rebigone{(2.1, -0.53)} %

\def\reone{(-0.3,-0.5)} %
\def\reseight{(0.6,0.25)} %
\def\retwo{(0.8,0.3)} %
\def\rethree{(-1.8,0.-0.7)} %

\def\refour{(-0.1,0.15)} %
\def\refive{(-0.1,0.15)} %
\def\resix{(-0.175,0.16)} %
\def\reseven{(-0.2,0.15)} %

\def\ssizz{1cm} %
\def\ssmag{3}

\begin{figure*}[!t] 
\centering
\tikzstyle{img} = [rectangle, minimum width=\imgWidth, draw=black]
\centering
\begin{subfigure}{\textwidth}
\centering
    \begin{subfigure}{\imgWidth}
        \begin{tikzpicture}[spy using outlines={green,magnification=\ssmag,size=\ssizz},inner sep=0]
            \node [align=center, img] {\includegraphics[width=\textwidth]{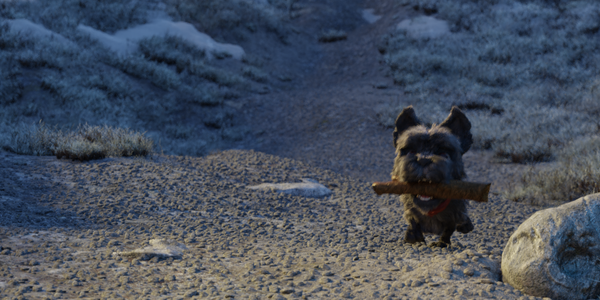}};
    	\end{tikzpicture}
    \end{subfigure}
    \begin{subfigure}{\imgWidth}
        \begin{tikzpicture}[spy using outlines={green,magnification=\ssmag,size=\ssizz},inner sep=0]
            \node [align=center, img] {\includegraphics[width=\textwidth]{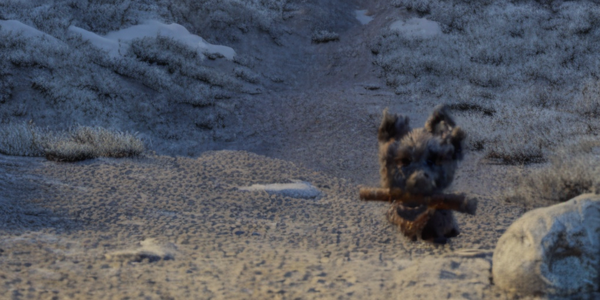}};
            \spy on \reone in node [left] at \rebigone;
    	\end{tikzpicture}
    \end{subfigure}
    \begin{subfigure}{\imgWidth}
		\begin{tikzpicture}[spy using outlines={green,magnification=\ssmag,size=\ssizz},inner sep=0]
            \node [align=center, img] {\includegraphics[width=\textwidth]{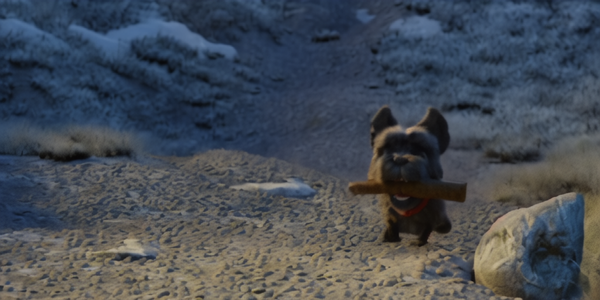}};
            \spy on \reone in node [left] at \rebigone;
    	\end{tikzpicture}
    \end{subfigure}
        \begin{subfigure}{\imgWidth}
        \begin{tikzpicture}[spy using outlines={green,magnification=\ssmag,size=\ssizz},inner sep=0]
            \node [align=center, img] {\includegraphics[width=\textwidth]{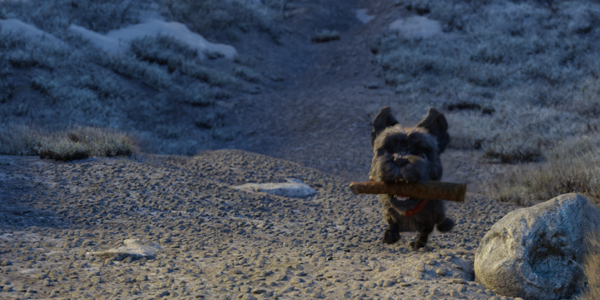}};
            \spy on \reone in node [left] at \rebigone;
    	\end{tikzpicture}
    \end{subfigure}
    \begin{subfigure}{\imgWidth}
        \begin{tikzpicture}[spy using outlines={green,magnification=\ssmag,size=\ssizz},inner sep=0]
            \node [align=center, img] {\includegraphics[width=\textwidth]{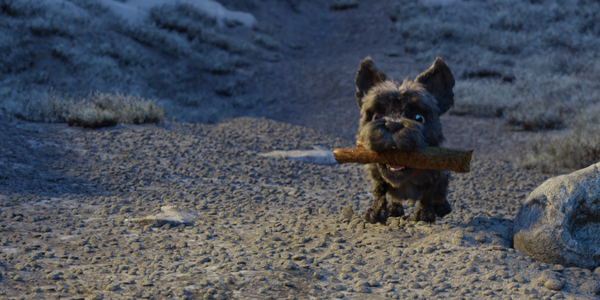}};
    	\end{tikzpicture}
    \end{subfigure}
    \begin{subfigure}{\imgWidth}
        \begin{tikzpicture}[spy using outlines={green,magnification=\ssmag,size=\ssizz},inner sep=0]
            \node [align=center, img] {\includegraphics[width=\textwidth]{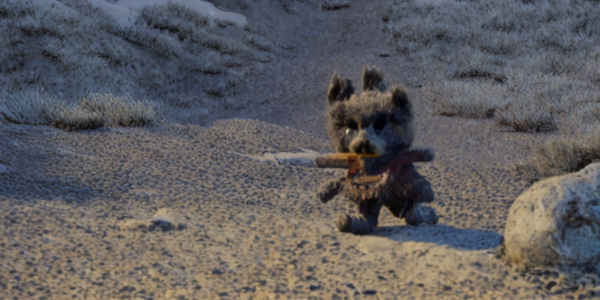}};
            \spy on \reseight in node [left] at \rebigone;
    	\end{tikzpicture}
    \end{subfigure}
    \begin{subfigure}{\imgWidth}
		\begin{tikzpicture}[spy using outlines={green,magnification=\ssmag,size=\ssizz},inner sep=0]
            \node [align=center, img] {\includegraphics[width=\textwidth]{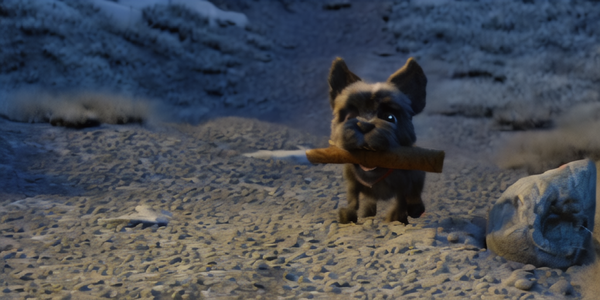}};
            \spy on \reseight in node [left] at \rebigone;
    	\end{tikzpicture}
    \end{subfigure}
        \begin{subfigure}{\imgWidth}
        \begin{tikzpicture}[spy using outlines={green,magnification=\ssmag,size=\ssizz},inner sep=0]
            \node [align=center, img] {\includegraphics[width=\textwidth]{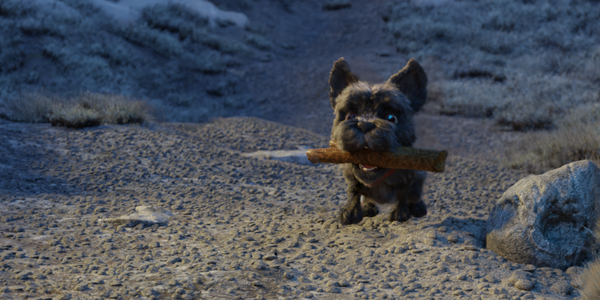}};
            \spy on \reseight in node [left] at \rebigone;
    	\end{tikzpicture}
    \end{subfigure}
        \\ %
    \begin{subfigure}{\imgWidth}
        \begin{tikzpicture}[spy using outlines={green,magnification=\ssmag,size=\ssizz},inner sep=0]
            \node [align=center, img] {\includegraphics[width=\textwidth]{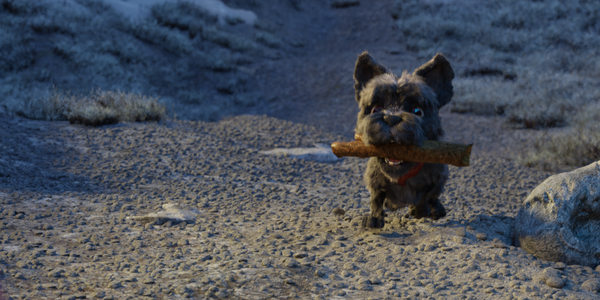}};
    	\end{tikzpicture}
    \end{subfigure}
    \begin{subfigure}{\imgWidth}
        \begin{tikzpicture}[spy using outlines={green,magnification=\ssmag,size=\ssizz},inner sep=0]
            \node [align=center, img] {\includegraphics[width=\textwidth]{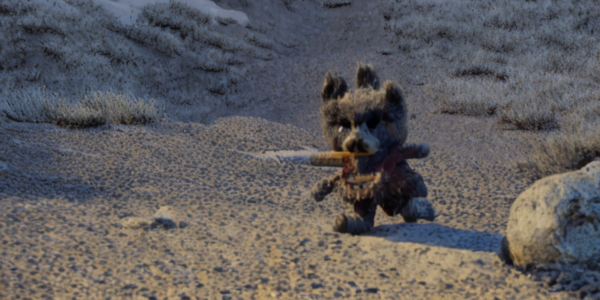}};
            \spy on \retwo in node [left] at \rebigone;
    	\end{tikzpicture}
    \end{subfigure}
    \begin{subfigure}{\imgWidth}
		\begin{tikzpicture}[spy using outlines={green,magnification=\ssmag,size=\ssizz},inner sep=0]
            \node [align=center, img] {\includegraphics[width=\textwidth]{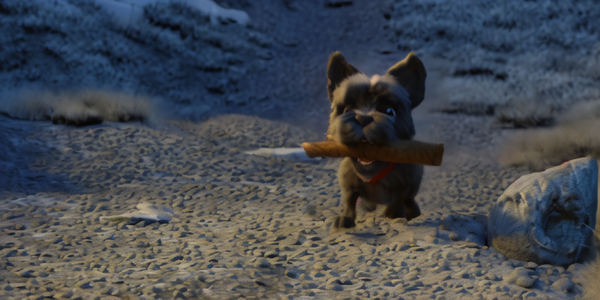}};
            \spy on \retwo in node [left] at \rebigone;
    	\end{tikzpicture}
    \end{subfigure}
        \begin{subfigure}{\imgWidth}
        \begin{tikzpicture}[spy using outlines={green,magnification=\ssmag,size=\ssizz},inner sep=0]
            \node [align=center, img] {\includegraphics[width=\textwidth]{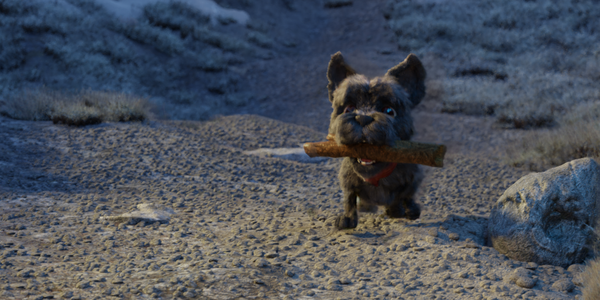}};
            \spy on \retwo in node [left] at \rebigone;
    	\end{tikzpicture}
    \end{subfigure}
      \\ %
    \begin{subfigure}{\imgWidth}
        \begin{tikzpicture}[spy using outlines={green,magnification=\ssmag,size=\ssizz},inner sep=0]
            \node [align=center, img] {\includegraphics[width=\textwidth]{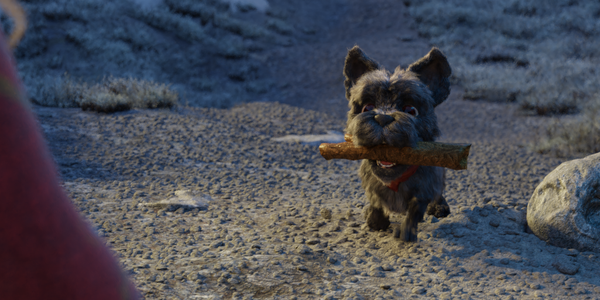}};
    	\end{tikzpicture}
     \caption*{Input Views (Left)}
    \end{subfigure}
    \begin{subfigure}{\imgWidth}
        \begin{tikzpicture}[spy using outlines={green,magnification=\ssmag,size=\ssizz},inner sep=0]
            \node [align=center, img] {\includegraphics[width=\textwidth]{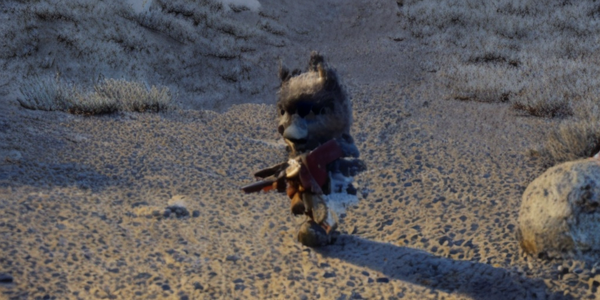}};
            \spy on \rethree in node [left] at \rebigone;
    	\end{tikzpicture}
     \caption*{ViewCrafter}
    \end{subfigure}
    \begin{subfigure}{\imgWidth}
		\begin{tikzpicture}[spy using outlines={green,magnification=\ssmag,size=\ssizz},inner sep=0]
            \node [align=center, img] {\includegraphics[width=\textwidth]{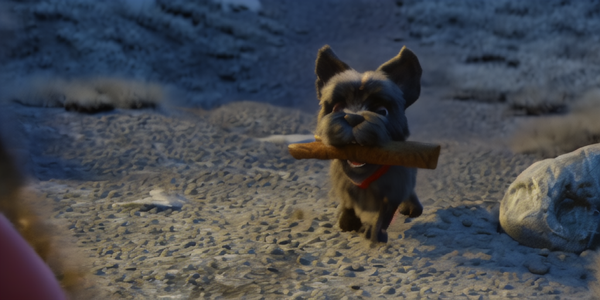}};
            \spy on \rethree in node [left] at \rebigone;
    	\end{tikzpicture}
     \caption*{StereoCrafter}
    \end{subfigure}
        \begin{subfigure}{\imgWidth}
        \begin{tikzpicture}[spy using outlines={green,magnification=\ssmag,size=\ssizz},inner sep=0]
            \node [align=center, img] {\includegraphics[width=\textwidth]{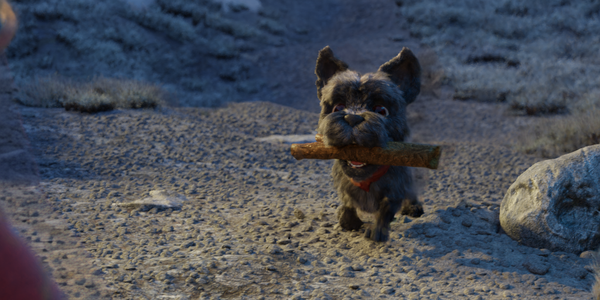}};
            \spy on \rethree in node [left] at \rebigone;
    	\end{tikzpicture}
      \caption*{\textbf{\method\ (Ours)}}
    \end{subfigure}
    \caption{Visual comparisons on details}
    \label{fig:supp-stereoconv-detail}
      \end{subfigure}
       \begin{subfigure}{\textwidth}
\centering
    \begin{subfigure}{\imgWidth}
        \begin{tikzpicture}[spy using outlines={green,magnification=\ssmag,size=\ssizz},inner sep=0]
            \node [align=center, img] {\includegraphics[width=\textwidth]{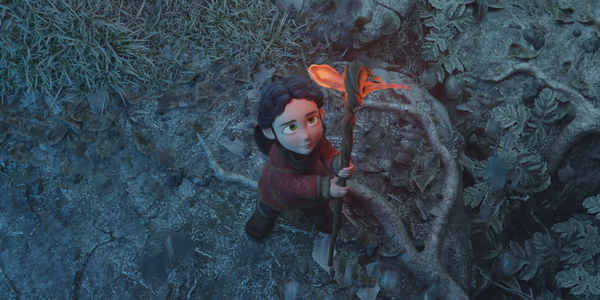}};
    	\end{tikzpicture}
    \end{subfigure}
    \begin{subfigure}{\imgWidth}
        \begin{tikzpicture}[spy using outlines={green,magnification=\ssmag,size=\ssizz},inner sep=0]
            \node [align=center, img] {\includegraphics[width=\textwidth]{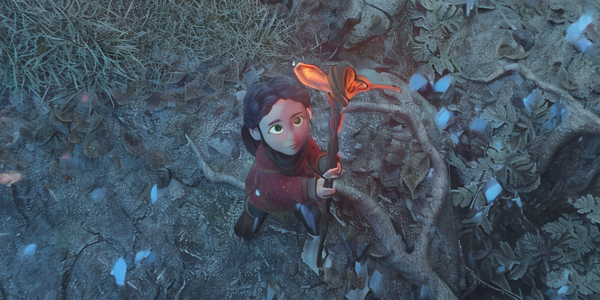}};
            \spy on \refour in node [left] at \rebigone;
    	\end{tikzpicture}
    \end{subfigure}
    \begin{subfigure}{\imgWidth}
		\begin{tikzpicture}[spy using outlines={green,magnification=\ssmag,size=\ssizz},inner sep=0]
            \node [align=center, img] {\includegraphics[width=\textwidth]{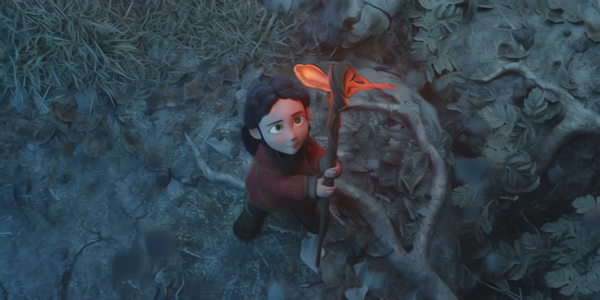}};
            \spy on \refour in node [left] at \rebigone;
    	\end{tikzpicture}
    \end{subfigure}
        \begin{subfigure}{\imgWidth}
        \begin{tikzpicture}[spy using outlines={green,magnification=\ssmag,size=\ssizz},inner sep=0]
            \node [align=center, img] {\includegraphics[width=\textwidth]{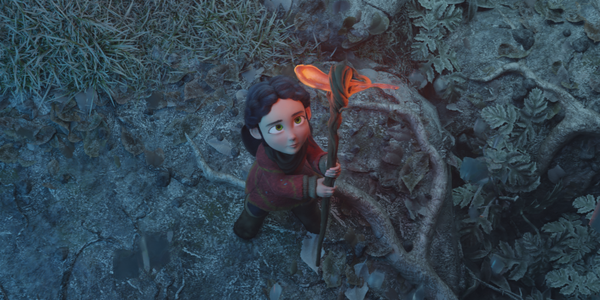}};
            \spy on \refour in node [left] at \rebigone;
    	\end{tikzpicture}
    \end{subfigure}
    \begin{subfigure}{\imgWidth}
        \begin{tikzpicture}[spy using outlines={green,magnification=\ssmag,size=\ssizz},inner sep=0]
            \node [align=center, img] {\includegraphics[width=\textwidth]{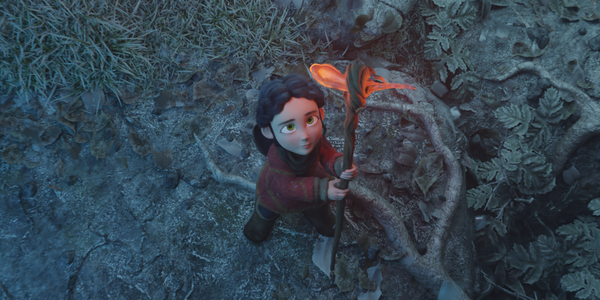}};
    	\end{tikzpicture}
    \end{subfigure}
    \begin{subfigure}{\imgWidth}
        \begin{tikzpicture}[spy using outlines={green,magnification=\ssmag,size=\ssizz},inner sep=0]
            \node [align=center, img] {\includegraphics[width=\textwidth]{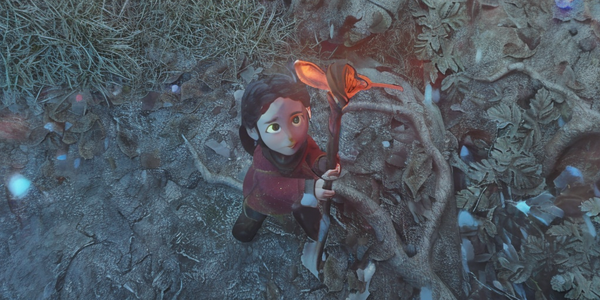}};
            \spy on \refive in node [left] at \rebigone;
    	\end{tikzpicture}
    \end{subfigure}
    \begin{subfigure}{\imgWidth}
		\begin{tikzpicture}[spy using outlines={green,magnification=\ssmag,size=\ssizz},inner sep=0]
            \node [align=center, img] {\includegraphics[width=\textwidth]{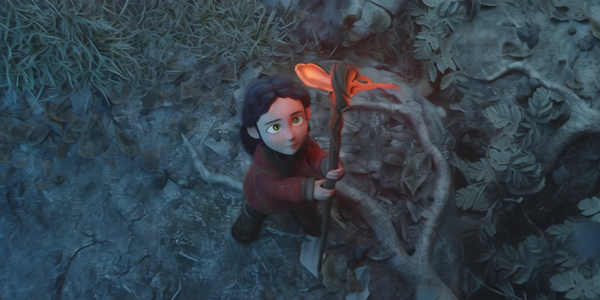}};
            \spy on \refive in node [left] at \rebigone;
    	\end{tikzpicture}
    \end{subfigure}
        \begin{subfigure}{\imgWidth}
        \begin{tikzpicture}[spy using outlines={green,magnification=\ssmag,size=\ssizz},inner sep=0]
            \node [align=center, img] {\includegraphics[width=\textwidth]{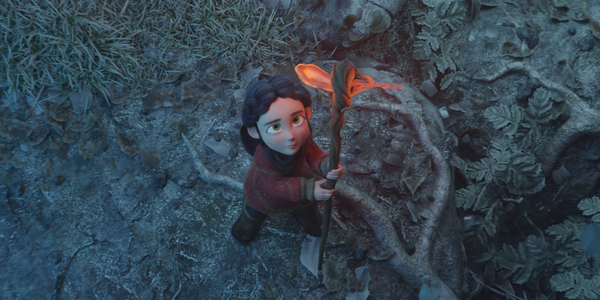}};
            \spy on \refive in node [left] at \rebigone;
    	\end{tikzpicture}
    \end{subfigure}
        \\ %
    \begin{subfigure}{\imgWidth}
        \begin{tikzpicture}[spy using outlines={green,magnification=\ssmag,size=\ssizz},inner sep=0]
            \node [align=center, img] {\includegraphics[width=\textwidth]{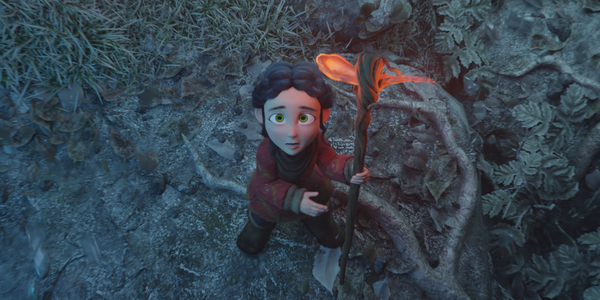}};
    	\end{tikzpicture}
    \end{subfigure}
    \begin{subfigure}{\imgWidth}
        \begin{tikzpicture}[spy using outlines={green,magnification=\ssmag,size=\ssizz},inner sep=0]
            \node [align=center, img] {\includegraphics[width=\textwidth]{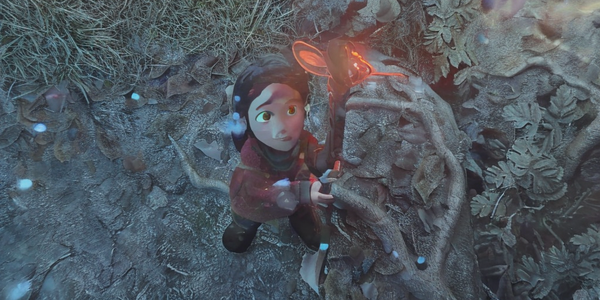}};
            \spy on \resix in node [left] at \rebigone;
    	\end{tikzpicture}
    \end{subfigure}
    \begin{subfigure}{\imgWidth}
		\begin{tikzpicture}[spy using outlines={green,magnification=\ssmag,size=\ssizz},inner sep=0]
            \node [align=center, img] {\includegraphics[width=\textwidth]{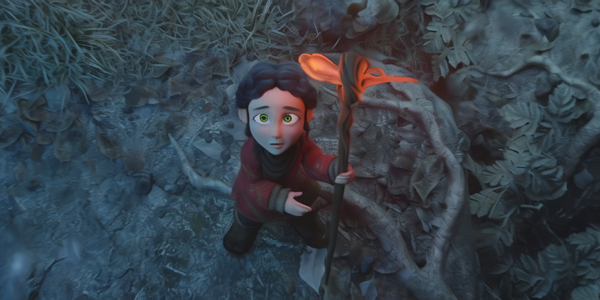}};
            \spy on \resix in node [left] at \rebigone;
    	\end{tikzpicture}
    \end{subfigure}
        \begin{subfigure}{\imgWidth}
        \begin{tikzpicture}[spy using outlines={green,magnification=\ssmag,size=\ssizz},inner sep=0]
            \node [align=center, img] {\includegraphics[width=\textwidth]{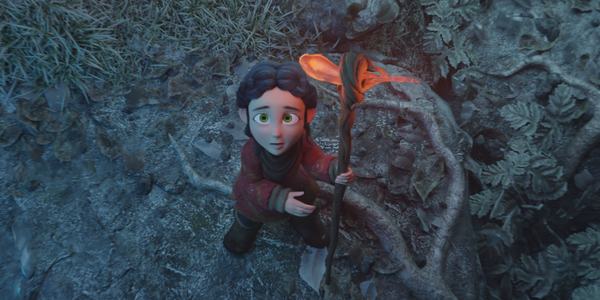}};
            \spy on \resix in node [left] at \rebigone;
    	\end{tikzpicture}
    \end{subfigure}
      \\ %
    \begin{subfigure}{\imgWidth}
        \begin{tikzpicture}[spy using outlines={green,magnification=\ssmag,size=\ssizz},inner sep=0]
            \node [align=center, img] {\includegraphics[width=\textwidth]{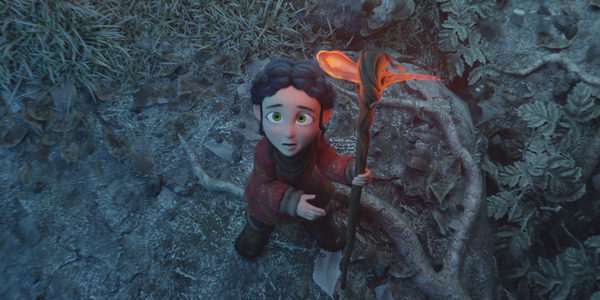}};
    	\end{tikzpicture}
     \caption*{Input Views (Left)}
    \end{subfigure}
    \begin{subfigure}{\imgWidth}
        \begin{tikzpicture}[spy using outlines={green,magnification=\ssmag,size=\ssizz},inner sep=0]
            \node [align=center, img] {\includegraphics[width=\textwidth]{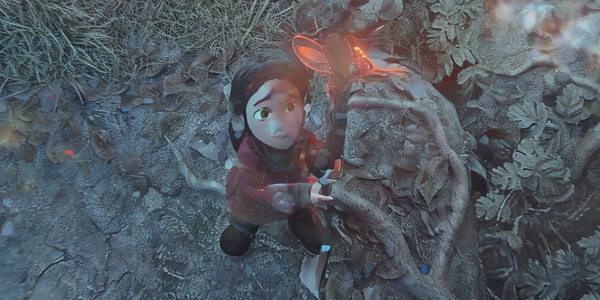}};
            \spy on \reseven in node [left] at \rebigone;
    	\end{tikzpicture}
     \caption*{ViewCrafter}
    \end{subfigure}
    \begin{subfigure}{\imgWidth}
		\begin{tikzpicture}[spy using outlines={green,magnification=\ssmag,size=\ssizz},inner sep=0]
            \node [align=center, img] {\includegraphics[width=\textwidth]{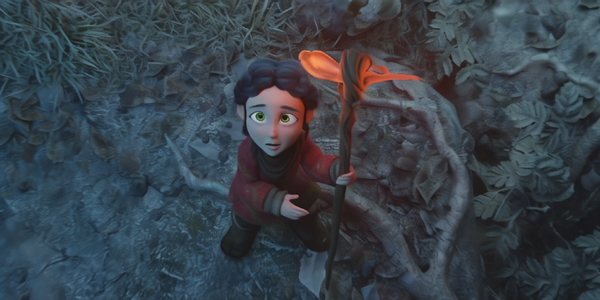}};
            \spy on \reseven in node [left] at \rebigone;
    	\end{tikzpicture}
     \caption*{StereoCrafter}
    \end{subfigure}
        \begin{subfigure}{\imgWidth}
        \begin{tikzpicture}[spy using outlines={green,magnification=\ssmag,size=\ssizz},inner sep=0]
            \node [align=center, img] {\includegraphics[width=\textwidth]{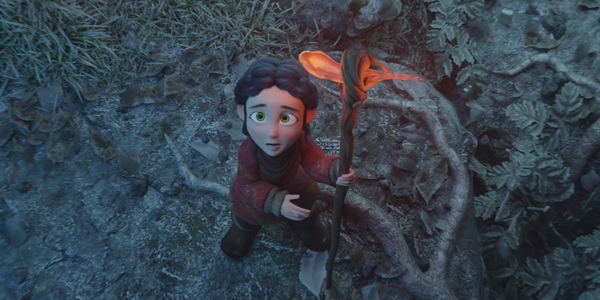}};
            \spy on \reseven in node [left] at \rebigone;
    	\end{tikzpicture}
      \caption*{\textbf{\method\ (Ours)}}
    \end{subfigure}
    \caption{Visual comparisons on consistency}
    \label{fig:supp-stereoconv-consistency}
    \end{subfigure}
    \caption{\textbf{Stereo video conversion results on the Spring dataset.} Right-eye views are synthesized based on the input left-eye views.}
 \label{fig:supp-stereoconv}
\end{figure*}